\documentclass[letterpaper]{article}
\usepackage[preprint]{aaai2027}
\usepackage[hyphens]{url}
\usepackage{graphicx}
\usepackage{natbib}
\usepackage{caption}
\usepackage{algorithm}
\usepackage{algorithmic}
\usepackage{booktabs}
\usepackage{xcolor}
\usepackage{cleveref}
\usepackage{xspace}
\usepackage{tikz}

\usetikzlibrary{positioning, arrows.meta}
\definecolor{turnbg}{HTML}{F2F4F7}
\definecolor{turnrule}{HTML}{6B7280}
\definecolor{movebg}{HTML}{FDF1DC}
\newcommand{\turn}[4]{%
  \noindent\colorbox{#1}{%
    \begin{minipage}{0.95\columnwidth}%
      \vspace{2pt}%
      {\footnotesize\textcolor{turnrule}{\textsc{\textbf{#2}}}%
       \ifx\relax#3\relax\else\ \textcolor{turnrule}{$\rightarrow$ \textsc{#3}}\fi}\\[1pt]
      {\small #4}\vspace{3pt}%
    \end{minipage}}\\[2.5pt]}

\newcommand{\benchname}{\textsc{GPS-Bench}\xspace}

\title{\benchname: A Governance Policy Benchmark for Automating Policy Analysis}
\author{
  Linh Le\textsuperscript{1},
  Melanie Bui\textsuperscript{1},
  My Chiffon Nguyen\textsuperscript{1},
  Zachary Schlosser\textsuperscript{1},
  David Williams-King\textsuperscript{1,2}
}
\affiliations{
  \textsuperscript{1}Lida Safety\\
  \textsuperscript{2}ERA
}

\begin{document}
  \maketitle

  \begin{abstract}
Policy analysis requires more than predicting whether a proposal will pass: it requires
identifying who will be affected, how those actors respond, and what follows. LLM-based policy
simulations model these processes at scale, but their validity is hard to establish when
plausible behaviour is never compared with observed outcomes. We introduce \benchname, an
evidence-grounded benchmark for governance policy simulation that links policies to relevant
actors, actor actions and downstream impacts using legislative records, lobbying disclosures,
regulatory documents, corporate filings, economic data and other public evidence. Actors are
reconstructed from the dated record rather than prompted as archetypes, so a persona is an
evidence object with provenance; a human-annotated pool forms the Gold evaluation set, while
cases labelled by a separate LLM from retrieved evidence are treated as Silver supervision and
never as test labels. Because every inference mode reads the same grounded state and emits the
same schema, \benchname turns ``does multi-agent simulation help?'' into a controlled
comparison: we contrast joint reasoning, independent and communicating actor agents,
graph-based methods and weight-level fine-tuning over one policy state. Fine-tuning on the
grounded record gives the strongest actor-level impact prediction, and decomposition does not
beat it; what decomposition adds is mechanism. Agents hold private, non-identical evidence,
each seeing its own exposure clause, and address named partners with concrete joint
proposals, what they offer, what they need in return, and why acting together beats acting
alone, so the coalitions that form can be checked against the commitments the record holds.
\benchname therefore gives a common empirical setting for studying when evidence, actor
modelling and multi-agent interaction improve the prediction and interpretation of policy
outcomes.
\end{abstract}

  \section{Introduction}
  Large language models (LLMs) are increasingly used to simulate political and institutional
  decision-making, cooperation and resource management \citep{piatti2024cooperate}, escalation
  \citep{hua2023waragent,rivera2024escalation}, collaborative law-making \citep{hota2025nomiclaw},
  roll-call prediction \citep{li2025political,grossmann2023ai}. Unlike rule-based models,
  LLM agents reason in natural language and simulate open-ended behaviour, but their validity
  is hard to establish: they can produce plausible interactions without reproducing the
  actors, actions, or outcomes of real policy processes. In practice they are judged by qualitative
  plausibility rather than realized outcomes \citep{larooij2025validation,barnett2024simulating},
  and role-played agents over-converge \citep{chuang2024opinion}, mistrack real groups \citep{santurkar2023whose},
  and carry political biases \citep{rozado2024political}.

  This validity gap is consequential in AI governance, where policy must anticipate fast-moving
  technology: frontier training compute has grown by roughly an order of magnitude per year
  \citep{sevilla2022compute,maslej2025aiindex} with large but uncertain stakes
  \citep{briggs2023growth,brynjolfsson2025genai,grace2024thousands,hendrycks2025superintelligence},
  and rules written against it are often obsolete on arrival
  \citep{collingridge1980,heim2024thresholds,gstrein2024gpai,oecd2025governing}. Forward-looking
  simulation is therefore attractive \citep{anthis2025llmsocial}, but equilibrium- and
  agent-based frameworks \citep{fagiolo2017macro} and their LLM successors are rarely scored
  against what actually happened.

  These settings share a design, LLM agents standing in for institutions, but differ in what
  they evaluate: one line studies \emph{emergent interaction} judged by plausibility
  \citep{hua2023waragent,rivera2024escalation,piatti2024cooperate,hota2025nomiclaw}; another
  predicts a single downstream label \citep{nay2017predicting,li2025political}, treating actors
  as features. We keep the agents-as-institutions setting and change the object of evaluation:
  governance simulation as \emph{evidence-grounded structured prediction}: actors
  reconstructed from the record rather than prompted as archetypes, conditioned on the dated
  situation, and scored along the whole actor\,$\to$\,action\,$\to$\,impact chain against
  predictor-independent targets.

  We introduce \benchname, an evidence-grounded governance policy benchmark built on
  the \emph{Governance Policy Simulator} (GPS). GPS separates policy simulation into two components:
  \textbf{state construction} and \textbf{inference}. For a policy $b$, public records
  reconstruct the information available before the decision point as an observed graph
  $G_{\le t}(b)$, containing the policy, antecedent events, and policy-specific actor states.
  An inference mechanism then predicts a continuation $\hat G^{\ast}(b)$, containing legislative
  outcome, affected actors, actor actions, intermediate state changes, and actor-level impacts.
  Historical impact and action records are used only as evidence or as evaluation
  targets for past policies; they are not inserted into the predicted portion of a held-out
  policy.

  \textbf{The actor layer is the core of the benchmark.} A forecast that stops at ``will
  it pass'' does not tell an analyst \emph{which} stakeholders a measure moves, \emph{how}
  each responds, or \emph{whether} it is helped or harmed. \textbf{\benchname therefore
  evaluates three linked targets over one evidence-grounded graph: legislative passage,
  affected-actor identification, and actor-level impact direction}, the two actor-level
  targets being its spine. Other trajectory components
  (mechanisms, world state changes, multi-step edges) are emitted and structured but not scored,
  so we test only the parts with retrospective labels (appendix).

  Because every inference mode operates over the same $G_{\le t}$ and emits the same
  output schema, GPS provides a controlled test of whether decomposing policy reasoning into
  interacting actor agents improves prediction. We compare joint inference over the
  complete policy state with independent actor inference, single-revision collaboration,
  and multi-round communication.

  \noindent
  \textbf{Contributions.}
  \begin{itemize}
    \itemsep2pt
    \setlength{\leftmargini}{1.2em}

    \item \textbf{Benchmark.} An evidence-grounded temporal benchmark linking AI policies to
      dated antecedents, policy-specific actors, recorded behaviour and observed stakeholder
      effects, with node- and edge-level provenance to public records.

    \item \textbf{Representation.} A typed temporal graph that cleanly separates the
      observed state $G_{\le t}$ from the predicted continuation $\hat G^{\ast}$ (Figure~\ref{fig:simframework}),
      decoupling knowledge grounding from inference.

    \item \textbf{Tasks.} Three linked, retrospectively scored tasks, legislative passage,
      affected-actor identification and actor-level impact direction, with contamination
      audits, a held-out behavioural-fidelity probe, and a label taxonomy keeping observed,
      human-verified and Silver targets distinct.

    \item \textbf{Inference study.} A controlled comparison of joint, independent,
      collaborative, and communicating inference, and of training-free vs.\ weight-level
      adaptation, over identical evidence, showing that the effective architecture is target-dependent
      and that a weight-level fine-tune on the grounded record surpasses in-context inference
      on every actor target.

    \item \textbf{Coalitions the record can check.} Agents holding private evidence address
      named partners with concrete joint proposals, what they offer, what they need back, why
      together beats alone, and the resulting coalitions are scored against the dated
      commitments on record (Tables~\ref{tab:pitch} and~\ref{tab:coalition}).
  \end{itemize}

  Our experiments address four questions. \emph{(RQ1, Grounding and behavioural fidelity)}
  Can public records reconstruct policy-specific actors and context, and do grounded agents
  reproduce held-out actor behaviour? \emph{(RQ2, Prediction)} How accurately can GPS predict
  legislative passage, affected actors, and actor-level impact direction? \emph{(RQ3,
  Inference)} How does joint inference compare with actor-decomposed inference when both
  receive the same grounded policy state? \emph{(RQ4, Interaction)} When actor
  decomposition is used, what does communication change in predictive performance and interaction
  structure?

  As a running example we use the \textbf{2022 U.S.\ advanced-chip export controls} (BIS,
  7 October 2022), which barred exports to China of advanced AI chips, the tools to make them,
  and U.S.-person support. The measure reached firms, foundries, allied toolmakers and states
  that its text never names, and each responded from a different position: chipmakers lobbied
  to recalibrate thresholds, foundries and allies re-planned supply, and the target state
  answered with its own controls. Which parties are reached, what each does, and who is helped
  or harmed are exactly the questions \benchname scores, and this instrument carries them
  through the paper, into the actor exchange of \Cref{tab:pitch} and the case study in the
  appendix.

  \section{Related Work and Dataset Gap}
  \textbf{LLM-based social and multi-agent simulation.} LLM agents stand in for institutions in
  political-military wargaming and escalation \citep{hua2023waragent,rivera2024escalation},
  cooperation and common-resource games \citep{piatti2024cooperate}, collaborative law-making
  \citep{hota2025nomiclaw}, and scenario generation for policy analysis
  \citep{grossmann2023ai}. These study \emph{emergent interaction} but do not ask whether the
  simulated actors and consequences correspond to a historical policy record; \benchname makes
  that correspondence the object of evaluation.

  \textbf{Grounding and validity of simulated agents.} The validity of LLM agents as proxies
  for real groups is contested: role-played opinions over-converge \citep{chuang2024opinion},
  mistrack the populations they represent \citep{santurkar2023whose}, and carry documented
  political bias \citep{rozado2024political}; a review of $35$ generative agent-based studies
  found $15$ rested on plausibility alone \citep{larooij2025validation}. We treat grounding not
  as preprocessing but as a measured contribution: actors are reconstructed from the public
  record and scored against behaviour they are documented to have taken.

  \textbf{Outcome prediction and policy impact.} A second line predicts legislative outcomes
  from bill text and sponsorship \citep{nay2017predicting} or roll-call votes from legislator
  personas \citep{li2025political}, and a third generates policy scenarios judged by perceived
  impact \citep{barnett2024simulating}. The first two are grounded in outcomes but forecast a
  single label, treat actors as features rather than reconstructing them, and stop at
  enactment; the third reconstructs consequences but scores them by plausibility.

  The gap is a benchmark that (i) reconstructs the \emph{bill-specific} cast from the public
  record rather than prompting archetypes, (ii) conditions on the dated situation a bill
  entered, and (iii) scores \emph{both} whether it passes and what follows against targets
  built independently of the predictor. \benchname is built to fill it, with provenance-graded
  actors and antecedents on the input side, official status and implementation records on the
  scoring side, every number reported against a trivial baseline, and every leakage limit
  quantified rather than hidden.

  \section{Problem Formulation}

  \begin{figure*}[t]
    \centering
    \includegraphics[width=\textwidth]{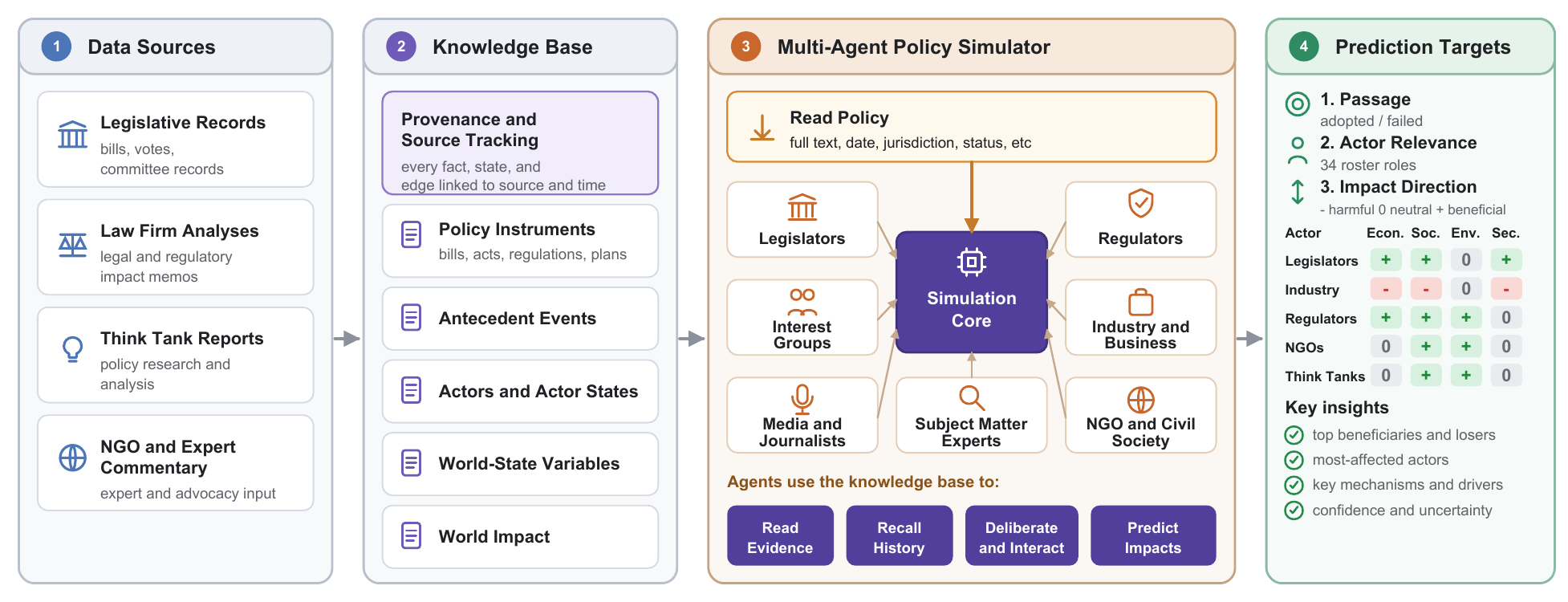}
    \caption{\textbf{The \benchname governance-policy simulation framework} (inputs
    $\to$ simulator $\to$ impact). Trusted, provenance-tracked data sources populate an evidence-grounded
    knowledge base in which every fact, state, and edge is traced to its source and time.
    For a policy, LLM-driven actor agents reason over the knowledge base associated with that
    policy: reading evidence and states, recalling history and positions, deliberating,
    and predicting actions and impacts, to produce per-bill impacts on each actor group
    by direction and magnitude across world dimensions (economic, social, environmental,
    innovation, security).}
    \label{fig:simframework}
  \end{figure*}

  GPS represents policy simulation as the continuation of a partially observed typed
  temporal graph (Figure~\ref{fig:simframework}; node signatures and the per-edge tuple
  in the appendix). For policy $b$, the \emph{observed state} at prediction
  time $t$ is $G_{\le t}(b)=\{P,A,X\}$, provisions $P$, dated antecedents $A$, and
  policy-specific actor states $X$ reconstructed from records available by $t$. Given $\mathrm{do}
  (P)$, an inference mechanism predicts a continuation
  $\hat G^{\ast}(b)=\{\hat Y,\hat R,\hat A,\hat M,\hat W,\hat I\}$ (passage outcome,
  relevant actors, actions, mechanisms, world-state changes, impacts); edges carry
  direction, probability, lag, controlling actor, and provenance. The boundary is temporal:
  information before $t$ is input, while post-$t$ events and consequences are prediction targets
  or evaluation evidence, \emph{not} input for the held-out policy, so a historical record
  can label an earlier policy without leaking into the one being predicted.

  The benchmark scores three linked targets along the causal spine of the graph, \emph{policy
  $\to$ affected actor $\to$ impact}, with passage as a policy-level target:
  \begin{itemize}
    \itemsep1pt
    \setlength{\leftmargini}{1.1em}

    \item \textbf{Passage} ($\hat Y$): adopted / failed, scored by balanced accuracy
      stratified by jurisdiction; predicted first, so an impact is not scored against a
      bill the model expected to die.

    \item \textbf{Actor relevance} ($\hat R$): for each of the $34$ roster roles,
      whether the bill materially affects it (macro-F1).

    \item \textbf{Impact direction} ($\hat I$): the welfare sign $\pm$ on each affected actor
      (macro-F1), scored either by self-report or, alternatively, inferred from the
      actor's predicted response by an independent judge.
  \end{itemize}
  The two actor-level targets ($\hat R,\hat I$) make the actor layer the core of the benchmark:
  they test not only \emph{which} stakeholders a policy moves, but \emph{whether} it helps or
  harms them. The predicted actions $\hat A$, intermediate mechanisms $\hat M$ and world-state
  changes $\hat W$ remain inspectable components of $\hat G^{\ast}$, and are emitted and
  auditable but not scored here, giving the contract $G_{\le
  t}+\mathrm{mode}\rightarrow\hat G^{\ast}$. The Appendix gives each field's
  label source and validation status, and traverses the formalism on a single provision.

  \section{The \benchname Benchmark}
  GPS constructs a temporal graph from the public record for one governance instrument,
  proposed or in force, and predicts its continuation under a stated intervention. Construction
  has two phases. \emph{Data collection} extracts each field from a named public source;
  \emph{data customization} transforms them into the graph the simulation is scored on. Both
  obey one constraint: every input traces to an inspectable source, and every evaluated output
  has a target built \emph{independently of the predictor}. Targets differ in evidential
  strength, from observed outcomes to a human-verified impact subset and Silver
  relevance and impact labels from a separate annotator (appendix).
  Figure~\ref{fig:simframework} gives the typed schema and the per-actor to world-level
  aggregation.

  \subsection{Data collection}
  Automated discovery does not determine inclusion in \benchname. Every record begins with the
  URL of an inspectable official publication, retrieved and staged; a curator then verifies the
  source and declares the instrument's title, issuing authority, jurisdiction, type, legal
  status, normative character, and date before it is admitted. Automation supports retrieval and
  validation but never determines inclusion or authoritative metadata. The corpus holds
  $1{,}233$ instruments, each with its full text. Non-English instruments retain their
  authoritative source text, and an attached English rendering never defines a label.
  the appendix details the procedure, provenance fields, and language coverage.

  \subsection{Policy episodes and Gold/Silver tiers}
  Each policy episode carries re-derived outcome labels and a dated pre-decision context, on a
  strict temporal split (train before 2024, test from 2024 on). The appendix maps the
  purpose-built sets; none is a superset of another.

The benchmark separates evaluation quality from supervision scale. The primary \emph{Gold}
pool contains $679$ human-annotated policy cases. Annotators code the
policy-specific actors, observable actor actions, and actor-level impacts from cited evidence;
official outcomes and recorded actions remain observed Gold. A further $4{,}235$
evidence-grounded records form an \emph{Evidence-Silver} expansion: a separate annotation LLM
receives retrieved evidence from the trusted source collection and must return both the label
and its supporting source span. Silver labels may be used for training, retrieval, and
ablation studies, but headline evaluation is performed on held-out Gold cases only.

Every instrument stores official title, issuing authority, jurisdiction, status, dates and
verified text, and every derived field stores an \texttt{evidence\_grade} and
\texttt{attribution}. Because adoption and failure are jurisdictionally confounded, passage is
always reported with jurisdiction-aware baselines.

\subsection{Actor, action and impact ontology}
  Actors are drawn from a fixed roster of $34$ stakeholder roles covering every actor named in
  the bills' causal chains; per bill a cast is selected by scoring each roster actor for
  relevance against the actors the record names, where recall is the meaningful side because
  the label counts only \emph{named} actors. Responses are coded on a controlled $15$-way
  action vocabulary (comply, disclose, adopt, build, restrict, enforce, litigate, lobby,
  support, oppose, fund, audit, coordinate, adapt, monitor).

  \textbf{Personas from the record.} Each actor's collected behavioural facts (roll-call and
  sponsorship history, filings, lobbying, trade series) become per-actor dated memories: a
  persona built from documented behaviour rather than a stipulated objective. A model-written
  persona is not gold; it is admissible only as scored against behaviour the actor is recorded
  as having taken, and sparse-record classes are reported rather than averaged away.
  Per-bill antecedent events are assembled under two rules, each dated strictly before the
  bill, and the bill's own downstream events withheld, which truncates what we inject but not
  the simulator's pretraining, so a clean record is necessary but not sufficient for leakage
  control (appendix).

  \textbf{Impact and world state.} Every actor--impact record is projected onto a second
  ontology, actor\,$\to$\,action\,$\to$\,impact, classifying each action into one of $12$
  mechanisms and each impact by world \emph{perspective} (ten), causal \emph{depth} (D1--D4)
  and \emph{functional role} (twelve), all deterministic from the collected fields. Over the
  corpus this yields $402$ typed impact nodes and $214$ graded causal maps across
  $1{,}233$ instruments. The functional-role grouping is what makes actor-level supervision
  viable, raising mean pre-2024 records per actor class from $3.2$ to $63.9$; the causal graphs
  are in the appendix.

\section{Governance Simulation Methods}

  \subsection{Inference over the graph}
  \label{sec:methods-inference} GPS separates the construction of the observed policy
  state from the mechanism used to predict its continuation. For bill $b$, all inference
  modes receive the same evidence-grounded graph $G_{\le t}(b)$ and produce predictions
  in the same output schema $G^{\ast}(b)$, so inference strategies can be compared without
  changing the underlying representation or the evidence available to the system. We
  evaluate three modes that differ only in what each LLM call sees, not in the evidence available:
  \begin{itemize}
    \itemsep1pt
    \setlength{\leftmargini}{1.1em}

    \item \textbf{Joint} $f_{\mathrm{joint}}$: full bill $+$ full cast $+$ all actor states
      $\to$ all actor outputs in one pass.

    \item \textbf{Independent} $f_{\mathrm{ind}}$: the same graph partitioned to one actor
      at a time $\to$ that actor's output, no cross-actor context.

    \item \textbf{Collaborative} $f_{\mathrm{collab}}$: the same partition $+$ exchanged messages
      $\to$ revised per-actor outputs.
  \end{itemize}
  All three emit the same output schema $\hat G^{\ast}$, so the comparison is diagnostic rather
  than architectural: if joint inference outperforms independent agents, actor decomposition
  has removed useful cross-actor context; if communication recovers performance, interaction
  has restored some of it. Within the graph, an actor's action drives an impact through an
  explicit two-layer typing: a \emph{mechanism}, one of twelve (cost, access, incentive,
  capacity, information, risk, competition, demand, supply, coordination, substitution, retaliation): changes
  a \emph{world-state variable} on one of ten world dimensions (economy, labour,
  technology, competition, safety, rights, governance, trade, geopolitics, environment). Mechanisms
  and world dimensions are thus separate ontologies, how a bill acts versus what it changes, and
  both populate $G^{\ast}$ without being scored; the scored consequence is the welfare
  \emph{direction} of the impact on each affected actor (\Cref{fig:simframework}).

  \subsection{Actor-decomposed and interactive inference}
  Multi-agent inference provides an actor-decomposed alternative to joint reasoning
  within GPS. Its purpose is twofold: to test whether actor-local reasoning improves
  prediction, and to expose interaction structure: coalition formation, persuasion, and coordinated
  action: that a single joint prediction does not represent explicitly. We specify how
  each actor agent is constructed from the record, conditioned on prior measures, and
  made to interact.

  \textbf{Evidence-grounded actor states.} Each agent is initialised from a dated actor state
  derived from public records: institutional role, observed positions, prior actions,
  resources, relationships and relevant indicators, each carrying source provenance and
  restricted to information available before the simulated decision point. A low-weight
  scenario-prior layer from published forward-looking analyses \citep{ai2027,ai2040} is
  evaluated separately and never used as a ground-truth label. The state is realised in the
  prompt as a \emph{persona}, whose reconstruction accuracy against held-out records is
  reported in the Results.

  \textbf{Conditioning on prior measures.} Each agent is additionally \emph{conditioned in
  context} on labelled pre-2024 examples: few-shot, our surrogate for fine-tuning the hosted
  $72$B models, alongside a genuine weight-level LoRA of an open $7$B model
  (Table~\ref{tab:impact-main}), so predictions are calibrated to how comparable actors
  behaved on comparable measures without seeing the $\ge$2024 test period.

  \textbf{Interaction regimes.} Over identical inputs the agents run in three regimes: \emph{independent}
  (each answers from its own persona in isolation), \emph{collaborative} (each sees a
  shared board of the others' positions and revises once), and \emph{communication} (over
  several rounds each agent chooses which others to convince or ally with, sends a
  message, and updates): followed by a synthesis pass that reads a single world state
  off the transcript. Comparing these against the zero-shot single call lets us measure the
  contribution of the agent machinery rather than assume it (Results).

  \section{Experimental Setup}
  Every forecasting claim uses a strict temporal split: train $<$2024, test $\ge$2024, never
  random, so no test instrument is seen in training; the primary predictor is separated from
  the test window by its model cutoff and is not fitted. Predictors see outcome-scrubbed
  context only, headline evaluation uses held-out Human Gold, and Evidence-Silver comes from a
  separate annotator that is never a predictor, so no model is scored against its own labels.
  Datasets and per-split sizes, the model list, input construction, decision modes, metrics,
  leakage controls and the fine-tuning configuration are given in full in
  the appendix, together with the verbatim prompts.

\section{Results and Analysis}
\label{sec:results}

We organize the evaluation by prediction target rather than by model family. Each result varies
the \emph{dataset/supervision setting} (No training, Gold, Silver, or Gold+Silver) and the
\emph{inference environment} (joint, independent actor agents, collaborative agents,
communicating agents, graph-based baselines, or weight-level fine-tuning).
\textbf{Every test sample reported below is Gold}, held out and human-annotated or observed,
with a cited source per field; Silver enters training only, so a supervision arm is measured by
what it transfers rather than by how closely it reproduces its own annotator. The one exception
is flagged in Table~\ref{tab:impact-main}. Extended results, contamination audits and protocol
sweeps are in the appendix.

\begin{figure}[t]
  \centering
  \turn{turnbg}{U.S.\ executive}{\relax}{%
    \emph{Acting alone:} ``Lower AI-chip export thresholds to include the A800/H800 class.''
    \textcolor{turnrule}{\footnotesize Why: prevent circumvention with downgraded chips.}}
  \turn{turnbg}{U.S.\ chipmakers}{U.S.\ executive}{%
    ``Lobby BIS for calibrated thresholds with tiered performance limits (TOPS, bandwidth) to
    preserve mid-market exports.''\\[2pt]
    \textbf{\footnotesize I offer}\ {\footnotesize Data on China's substitution gaps, plus
      lobbying leverage with allies.}\\
    \textbf{\footnotesize I need}\ {\footnotesize Legal safe harbour for compliant chips, and
      faster licence approvals.}\\
    \textbf{\footnotesize Beats alone}\ {\footnotesize Aligns industry and policymakers on
      enforceable thresholds, avoiding total market loss to SMIC and Huawei.}}
  \turn{movebg}{U.S.\ executive}{\relax}{%
    \emph{Commits, alongside Congress:} calibration, \textbf{not} the unilateral tightening it
    chose alone.}
  \caption{\textbf{One agent convincing another to act jointly}, on the 2022 U.S.\
  advanced-chip export controls. Each agent reasons from its own private evidence and neither
  sees the other's proposal before making its own; the recipient's committed action differs
  from the position it held alone.}
  \label{tab:pitch}
\end{figure}

\subsection{Grounding and annotation validity (RQ1)}

Before evaluating forecasts, we test whether the benchmark reconstructs the policy state every
inference mode depends on. Cast selection recovers most actors named in the legislative record,
antecedent extraction passes verbatim-quote verification at a comparable rate, and
reconstruction is strongest for classes with dense public records. On a separate overlap audit
the Silver annotator agrees with human coding more strongly on extractable structural fields
than on interpretive ones, so Silver is scalable supervision rather than a substitute for Gold
evaluation.

Historical behaviour also carries predictive information: conditioning an actor on its dated
prior actions helps where history exists, while rich narrative personas alone are less
reliable. The useful signal is documented precedent, not descriptive text.

\subsection{Legislative passage (RQ2)}

Table~\ref{tab:passage-main} separates evidence conditions from inference environments.
Passage is unusually sensitive to political timing: a calibrated joint forecast reaches
balanced accuracy $0.80$, and dated antecedent events raise it to $0.89$. A jurisdiction-only
baseline reaches $0.86$, so the passage corpus carries a substantial institutional confound.
The multi-agent score ($0.96$) is retained only as a contamination-bounded upper bound.

\begin{table}[t]
  \centering
  \small
  \setlength{\tabcolsep}{4pt}
  \caption{\textbf{Legislative passage under different evidence and inference settings},
  scored by balanced accuracy (BA), which is the metric passage is reported on throughout
  because the pool is jurisdictionally skewed. ``Gold'' denotes the held-out human/observed
  evaluation pool. The multi-agent synthesis row is contaminated and shown only as an upper
  bound.}
  \label{tab:passage-main}
  \begin{tabular}{@{}llcc@{}}
    \toprule
    \textbf{Dataset / evidence} & \textbf{Environment} & \textbf{BA} & \textbf{Status} \\
    \midrule
    Gold, jurisdiction only      & Statistical prior & 0.86 & baseline \\
    Gold, bill only              & Joint calibrated  & 0.80 & primary \\
    Gold, bill + antecedents     & Joint calibrated  & \textbf{0.89} & primary \\
    Gold, bill + actor synthesis & Multi-agent       & 0.96 & contaminated UB \\
    \bottomrule
  \end{tabular}
\end{table}

Passage and impact should therefore not share a single context policy: antecedents help
passage because they encode timing and institutional pressure, whereas actor-level impact
depends more on the mechanism of the policy itself.

\begin{table}[t]
  \centering
  \caption{\textbf{A reciprocated coalition} on the NAIRR pilot (dated $2024$, held out of
  training). Each agent proposes independently, without seeing the other's proposal; the
  record rows are the dated commitments the corpus holds for the same instrument.}
  \label{tab:coalition}
  \small
  \setlength{\tabcolsep}{4pt}
  \begin{tabular}{@{}p{0.20\columnwidth}p{0.74\columnwidth}@{}}
    \toprule
    \multicolumn{2}{@{}l}{\textit{Proposals emitted by the simulation}} \\
    NVIDIA $\rightarrow$ Microsoft & Collaborate to provide GPU infrastructure and
      computational resources for the NAIRR pilot, giving researchers access to AI compute. \\
    Microsoft $\rightarrow$ NVIDIA & Collaborate on AI infrastructure for the NAIRR pilot,
      leveraging NVIDIA's GPU technology to accelerate model training. \\
    \midrule
    \multicolumn{2}{@{}l}{\textit{What the record holds for the same instrument}} \\
    NVIDIA & \texttt{announce\_investment}: \$$30$M committed, \$$24$M of it as DGX Cloud
      compute, with software licences to national supercomputing centres. \\
    Microsoft & \texttt{announce\_investment}: \$$20$M in Azure credits, plus model access
      delivered through the firm's own cloud. \\
    \bottomrule
  \end{tabular}
\end{table}

\subsection{Actor relevance prediction (RQ2)}

The actor layer first asks \emph{who is reached by the policy?}, reported in
Table~\ref{tab:actor-main}. Training-free joint inference is stronger here, comparing the full
cast against the whole policy state, while actor-decomposed environments lose cross-actor
context. A weight-level LoRA on the grounded record reaches relevance F1 $0.62$.

\begin{table}[t]
  \centering
  \small
  \setlength{\tabcolsep}{3pt}
  \caption{\textbf{Actor relevance prediction} across supervision and simulation environments,
  as relevant-class F1. Gold+Silver adds evidence-grounded supervision with evaluation still on
  held-out Gold.}
  \label{tab:actor-main}
  \begin{tabular}{@{}p{1.4cm}p{1.55cm}p{2.3cm}c@{}}
    \toprule
    \textbf{Training} & \textbf{Environment} & \textbf{Model setting} & \textbf{F1} \\
    \midrule
    None        & Joint              & zero-shot          & 0.39 \\
    None        & Independent agents & actor-local        & 0.28 \\
    None        & Collaborative      & one board revision & 0.27 \\
    Gold/Silver & Joint              & 72B in-context     & 0.43 \\
    Gold+Silver & Independent actor  & 7B LoRA, temporal  & \textbf{0.62} \\
    \bottomrule
  \end{tabular}
\end{table}

The main implication is not that actor decomposition is uniformly better or worse. Relevance is
a global selection problem, so joint inference benefits from seeing all actors at once, whereas
actor-level impact is local and benefits from an actor's own behavioural history. The benchmark
therefore separates these targets rather than treating ``actor quality'' as a single score.

\subsection{Actor-level impact prediction (RQ2--RQ3)}

Impact is the central downstream task because it links a policy and actor response to a signed
consequence. Table~\ref{tab:impact-main} crosses dataset and environment settings. Among
training-free methods on the 34-role grounded corpus, joint inference reaches macro-F1 $0.77$,
compared with $0.65$ for independent agents and $0.64$ after a single collaborative revision.
Three rounds of communication recover part of the decomposition gap ($0.70$), indicating that
interaction can restore some context that was removed when the policy state was partitioned.

Weight-level grounding is the strongest intervention. On the human-curated pool a Qwen2.5-7B
LoRA lifts impact direction from macro-F1 $0.604$ untrained to $0.673$ on Gold supervision and
$0.630$ on Silver, with the union reaching $0.716$: the tiers are close once the prompts match,
and each still contributes what the other misses. On the role-level projection of the graph
over the full corpus, trained on Gold and Silver together, the same model reaches $0.85$, well
above the per-role majority prior.

\begin{table}[t]
  \centering
  \small
  \setlength{\tabcolsep}{3pt}
  \caption{\textbf{Actor-level impact prediction} across supervision and inference settings,
  macro-F1 for impact direction on a temporal split. The upper block is training-free inference
  on the role-level Gold surface (``Comm-3'' is three communication rounds); the lower block
  varies only the supervision tier, with every arm evaluated on the same held-out
  human-curated Gold pairs. $^{\dagger}$A different surface: $1{,}331$ (bill, role) pairs over the twelve
  functional roles the graph's actor types collapse into, on the whole $>$2k corpus. Its test
  labels combine Gold and Silver, so it is reported separately rather than beside the
  block above.}
  \label{tab:impact-main}
  \begin{tabular}{@{}p{2.05cm}p{1.55cm}p{2.2cm}c@{}}
    \toprule
    \textbf{Evaluation / supervision} & \textbf{Environment} & \textbf{Model setting} &
    \textbf{F1} \\
    \midrule
    \multicolumn{4}{@{}l}{\textit{Training-free inference, role-level Gold}} \\
    None                       & Joint              & zero-shot          & 0.77 \\
    None                       & Independent agents & actor-local        & 0.65 \\
    None                       & Collaborative      & one revision       & 0.64 \\
    None                       & Communicating      & Comm-3             & 0.70 \\
    None                       & Graph hybrid       & structural prior   & 0.76 \\
    \midrule
    \multicolumn{4}{@{}l}{\textit{Weight-level supervision, scored on human-curated Gold}} \\
    None                       & Independent actor  & 7B base, no adapter & 0.604 \\
    Gold                       & Independent actor  & 7B LoRA             & 0.673 \\
    Silver                     & Independent actor  & 7B LoRA             & 0.630 \\
    Gold + Silver              & Independent actor  & 7B LoRA             & \textbf{0.716} \\
    \midrule
    \multicolumn{4}{@{}l}{\textit{Role-level graph projection, $>$2k corpus}$^{\dagger}$} \\
    Gold + Silver              & Independent actor  & 7B LoRA             & \textbf{0.85} \\
    \bottomrule
  \end{tabular}
\end{table}

These results separate two roles of the benchmark. \emph{Predictive validity} asks whether
the final actor-level impact is correct. \emph{Mechanistic resolution} asks whether the system
also exposes the actions, messages, coalitions, and intermediate pathways that lead to that
prediction. Joint inference is currently stronger on the former in training-free settings;
actor-decomposed simulation is richer on the latter. Historical supervision can substantially
reduce the predictive cost of decomposition.

\subsection{Multi-agent interaction and communication (RQ4)}

Collaboration pays off exactly where agents differ: when each agent sees only its own exposure
clause, one round of exchange revises stances and improves on independent inference, which
identifies private evidence as the condition that makes deliberation worth running.
Communication is a property of the governance environment, not only an accuracy intervention.
Additional rounds give no monotonic gain, but the protocol changes the interaction structure:
targeted convince/ally communication produces strong same-stance targeting (homophily $0.77$),
while adversarial debate preserves impact accuracy better than targeted or broadcast
communication. The resulting graph contains coalitions and actor-to-actor responses a joint
forecast does not expose.

\paragraph{Coalition formation: agents committing to joint action.}
The communication environment's distinctive output is not a revised label but a
\emph{proposal}: each actor may address a named partner with a concrete joint move and a
reason for it. A proposal is \emph{reciprocated} when the named partner, reasoning
independently from its own evidence, proposes back to the proposer. Reciprocation is the
operational test of persuasion in \benchname, because neither agent sees the other's proposal
before making its own; agreement has to be reconstructed from each actor's separate position
rather than negotiated in a shared transcript.

Over the fine-tuned actor pool this produces $74$ proposals from $47$ proposers to $31$
partners across $15$ instruments, of which $14$ are reciprocated, forming $7$ mutually agreed
coalitions. No actor proposes to itself. The most-sought partners are the parties that hold
what others need: China ($7$ approaches), the European AI Office ($6$), and NVIDIA, Microsoft
and OpenAI ($5$ each). Coalitions form both within blocs (Anthropic$\leftrightarrow$OpenAI on
SB~1047) and across the firm--government boundary (Anthropic$\leftrightarrow$the Governor of
California on SB~53; Waymo$\leftrightarrow$the United Kingdom on the Automated Vehicles Act).

\Cref{tab:coalition} gives one such coalition in full, on an instrument dated $2024$ and
therefore held out of training: NVIDIA and Microsoft each independently name the other as the
compute partner for the NAIRR pilot, and the record confirms both committed, at \$$30$M and
\$$20$M.

\Cref{tab:pitch} shows what one agent actually says to another: not a bare label, but the joint
action, what it \emph{offers}, what it \emph{needs} back and what becomes possible only
together. The chipmakers' agent trades substitution data and allied lobbying leverage for a
legal safe harbour, and the executive agent, which alone would have tightened, commits to
calibration instead. The pitch, not the label, is what changed.

Coalitions form along existing lines: across $77$ committed coalitions among $13$
record-grounded actors, \emph{none} crosses the U.S.--China boundary, while instructing the
agents to reach across it yields $37$ of $72$. The contrast is a property of the instruction
rather than an emergent finding: the bloc boundary is a strong default prior, not a constraint.

The same example bounds the mechanism, since a matched pair is easy to overread as a validated
prediction. The record documents six firms making \emph{parallel} commitments to one federal
programme, not a bilateral NVIDIA--Microsoft agreement: the simulation recovers who commits and
that the two are natural counterparts, but renders as a pairwise deal what the world organised
through a convening institution. Reciprocation evidences a shared joint interest, not a
contract.

An analyst may care both about the aggregate forecast and about which actors coordinate,
oppose, lobby or retaliate, so multi-agent interaction stays first-class even where it is not
the highest-scoring predictor.

\paragraph{Case studies.} Four extended studies in the appendix carry the
mechanism further than aggregate scores can: a \emph{US--China semiconductor} study running
thirteen record-grounded actors on the 2022 export controls, logging every pitch, coalition and
stance revision; a \emph{worked dossier} giving the gold record for one bill per reached actor;
a \emph{walkthrough} of that bill through each inference mode; and a \emph{gold anchor} study
scoring our labels for eleven governance failures against the official inquiry that named the
failed link.

\subsection{Cross-task findings}

Three findings recur. \textbf{Evidence can be routed per target}: antecedents give passage its
largest gain, while impact is best served by the instrument's own mechanism.
\textbf{Behavioural grounding is the reliable lever}: an actor's dated record carries the
signal narrative elaboration does not. \textbf{Supervision and interaction compose}: joint
context suits global selection, actor decomposition yields explicit behavioural and relational
output, and Gold and Silver together beat either alone. Every predicted edge is typed and
dated, so it can be registered before the outcome is observed.

\section{Limitations}
\benchname remains a first version, and its scope is bounded in three ways. The corpus is
English-heavy with substantial US coverage, so legislative passage is partly confounded by
jurisdiction, which is why every passage result is reported against a jurisdiction-aware
baseline. Actor histories are reconstructed from the public record, which represents visible
behaviour, lobbying, litigation, public statements and formal enforcement, better than private
negotiation, so the actor layer models what an outside analyst could have observed rather than
everything that occurred. Finally, the benchmark directly scores passage, actor relevance and
impact direction; the mechanisms and world-state edges the simulator also emits are typed and
inspectable, and scoring them prospectively, once the outcomes they predict have resolved, is
the natural next step.

\section{Conclusion}
\benchname turns governance simulation into an evidence-grounded prediction problem, scoring
the trajectory from policy to affected actor to actor-level impact. The useful architecture
proves target-dependent, and a simulation is judged on the record.

\clearpage
\appendix
\setcounter{secnumdepth}{2}
\renewcommand{\thesection}{\Alph{section}}
\setcounter{section}{0}

\section*{Appendix overview}
\noindent The appendices below extend the main paper. Section letters and table
numbers (A1, A2, \ldots) are the ones the body cites. The first sections are the
paper's appendices: corpus construction and provenance, the experimental setup,
evidence sources, the extended results tables, additional figures, the multi-agent
analysis, the case study, the worked dossier and inference walk-through, and the
prompt templates. The sections after those document the precursor system whose
failure-mechanism vocabulary the main text reuses as its consequence label space.
References are shared with the main paper.
\renewcommand{\thetable}{A\arabic{table}}
\renewcommand{\thefigure}{A\arabic{figure}}
\setcounter{table}{0}
\setcounter{figure}{0}
  \section{Data collection, language, and expert-analysis provenance}

  \subsection{Collection procedure and corpus composition}
  \label{app:collection}
  Ingestion is a two-step, URL-first procedure. The fetch step retrieves the nominated source
  as HTML or extracts PDF text with Poppler, detects the language, checks the candidate against
  the corpus for duplication, and stages the result for inspection. The add step admits
  the record only after a curator has read the staged text and declared its title, issuing
  authority, jurisdiction and jurisdiction type, instrument type, original and normalised legal
  status, normative character, issue date, and a source-grounded description of its AI
  provisions.

  \benchname holds $1{,}233$ instruments from $350$ publisher domains, each with its full text.
  Of these, $431$ carry $641$ independently published expert analyses
  (Appendix~\ref{app:expert-analysis}); language coverage is in Appendix~\ref{app:translation}.
  Each document stores a SHA-256 of its text and its extraction provenance, and each source its
  URL, publisher, and official-source status. Legal status is harmonised by declared mapping
  rules. Before release the corpus must pass schema validation, a text audit for extraction
  artefacts and translation coverage, and duplicate detection. Table~\ref{tab:sources} lists the
  remaining evidence sources and the fields they populate. Outcome labels are verified against
  the GovInfo BILLSTATUS collection, $82{,}967$ bills across Congresses 114--119, of which $558$
  are AI-related and $14$ enacted.

  \subsection{Language coverage and translation provenance}
  \label{app:translation}
  Of the $1{,}233$ instruments, $1{,}162$ have English as the language of their stored text and
  $71$ span $20$ other languages (Table~\ref{tab:lang}). Language is detected at ingestion
  rather than inferred from jurisdiction, and the source-language title is retained in
  \texttt{title\_original} beside the English \texttt{title\_official}. Some instruments,
  chiefly Chinese, are obtained from English-language publishers such as CSET and DigiChina;
  their stored text is English and they are counted as English, so the $71$ counts instruments
  whose stored text is non-English.

  Of those $71$, three carry an English version issued by the originating body, $57$
  non-English instruments are translated into English with Gemini 3.5 Flash, and $12$ carry no
  English rendering and are retained only in their source language. All $57$ translations were
  produced in a single corpus-level pass, so the procedure is consistent across ingestion dates,
  though this does not imply uniform quality across languages.

  A rendering never replaces the authoritative text. It is stored as a separate document under
  \texttt{document\_role: translation} with its own language code and a SHA-256 reference to its
  source, so the two can be compared. Translations may enter model context but define no scored
  target; every target is coded from the authoritative text or from an independently published
  expert analysis.

  \begin{table}[t]
    \centering
    \small
    \setlength{\tabcolsep}{5pt}
    \begin{tabular}{@{}lcccc@{}}
      \toprule
      Language & Instr. & Published & Machine & None \\
      \midrule
      Chinese (zh)       & $23$ & --  & $11$ & $12$ \\
      Spanish (es)       & $18$ & --  & $18$ & --   \\
      German (de)        & $3$  & --  & $3$  & --   \\
      Dutch (nl)         & $3$  & --  & $1$  & $2$  \\
      Danish (da)        & $2$  & --  & $2$  & --   \\
      French (fr)        & $2$  & --  & $2$  & --   \\
      Greek (el)         & $2$  & --  & $1$  & $1$  \\
      Indonesian (id)    & $2$  & --  & $2$  & --   \\
      Hebrew (he)        & $2$  & $2$ & --   & --   \\
      Italian (it)       & $2$  & --  & $2$  & --   \\
      Korean (ko)        & $2$  & --  & $1$  & $1$  \\
      Thai (th)          & $2$  & --  & $2$  & --   \\
      Other ($8$ langs.) & $8$  & $1$ & $6$  & $1$  \\
      \midrule
      \textbf{Total}     & $\mathbf{71}$ & $\mathbf{3}$ & $\mathbf{51}$ & $\mathbf{17}$ \\
      \bottomrule
    \end{tabular}
    \caption{\textbf{Non-English instruments and their attached English renderings.} Rows count
    instruments whose \emph{stored} text is non-English. \emph{Published} is an English text
    issued by the originating body or an established translator; \emph{Machine} is a Gemini 3.5
    Flash translation; \emph{None} means no English rendering is attached. Instruments obtained
    from English-language publishers such as CSET and DigiChina count as English and fall
    outside this table.
    \emph{Other} is Portuguese, Hungarian, Persian, Japanese, Serbian, Russian, Slovak, and
    Vietnamese, one instrument each.}
    \label{tab:lang}
  \end{table}

  \subsection{Human expert-analysis provenance}
  \label{app:expert-analysis}
  The $641$ analyses come from $441$ distinct publishers, including law firms, think tanks, news
  organisations, academic institutions, industry and civil-society organisations, and
  independent policy analysts. They were not commissioned, rewritten, or generated for the
  benchmark. An instrument may carry several, preserving distinct legal, policy, and sectoral
  readings rather than collapsing them into one benchmark-authored summary.

  Each analysis is stored separately from the instrument it discusses and linked to it. Its
  provenance records the source URL and publisher for every analysis, the publication date for
  $609$ of $641$, the author for $438$, and the title where the source states one. Publisher
  provenance is recorded by name rather than assigned to a coded category, and no publisher type
  is treated as carrying greater evidentiary weight. Analyses form the human-analysed evaluation
  subset but do not alter the instrument text or its metadata.

  \section{Label taxonomy}
  \begin{table}[t]
    \centering
    \small
    \setlength{\tabcolsep}{4pt}
    \begin{tabular}{@{}l p{2.9cm} c c c@{}}
      \toprule Tier          & Source                              & Human-ver.\  & Train & Head.\ eval \\
      \midrule Observed gold & official recorded outcome / action  & observable   & yes   & yes         \\
      Human gold             & human annotation $+$ cited evidence & yes          & yes   & yes         \\
      Evidence silver        & separate LLM $+$ retrieved evidence & no           & yes   & no          \\
      Prediction             & GPS simulator output                & no           & no    & scored      \\
      \bottomrule
    \end{tabular}
    \caption{\textbf{Label taxonomy.} \benchname separates a \emph{human-validated
    evaluation core} (observed and human-gold tiers) from an \emph{evidence-grounded
    silver expansion} (separate-LLM labels with retrieved evidence and provenance).
    Silver labels enter training but not headline evaluation; predictions are the scored simulator
    output, not a label. A reviewer can read the benchmark's integrity from this table:
    no headline result rests on silver labels alone.}
    \label{tab:labels}
  \end{table}

  \section{Experimental setup in full}
  \label{app:setup-main}

  Our evaluation is organized by what kind of ground truth each claim can have, and is
  asymmetric between the two questions by design.

  \subsection{Datasets and splits}
  \textbf{Train/test split.} Every forecasting claim uses a strict \emph{temporal} split: train
  $<$2024, test $\ge$2024, not random (Table~\ref{tab:splits}). The LLM predictors are temporally
  separated by \emph{model cutoff} (\texttt{gpt-3.5-turbo-0613}, trained to September 2021),
  so no LLM is fitted and ``train'' marks only the exclusion boundary. The two \emph{fitted}
  models are the weight-level LoRA fine-tune (main paper) and the graph-structural
  impact hybrid; the latter trains on the sub-2k pre-2024 prior and is a caveated, non-recommended
  baseline (Results).

  \textbf{Leakage-free passage.} Leakage-freeness and label resolution are in tension: a measure
  avoids temporal contamination only because it is recent, but is labelled \emph{failed} only
  once its legislature adjourns (up to two years), whereas adoption is observable at
  once. Post-cutoff resolved negatives are therefore the binding constraint, and the
  reachable pool contains only three (one vetoed state bill, two withdrawn international
  measures). We consequently exclude a fully contamination-controlled passage arm from
  the primary results and retain $68$ dated pending predictions for prospective
  evaluation after the 119th Congress adjourns (3 January 2027).

  \subsection{Evaluation metrics}
  Each prediction is scored with the metric standard for its type: \textbf{passage} by
  \emph{balanced accuracy} (jurisdiction-stratified, enacted-class P/R vs.\ always-no) \citep{nay2017predicting};
  \textbf{actor relevance} and \textbf{impact direction} by \emph{macro-F1}; the \textbf{action}
  multilabel target by micro/macro/family-F1 with no-action recall. Impact additionally reports
  accuracy and balanced accuracy (Tables~the main paper,~\ref{tab:respimpact}).

  \subsection{Settings}
  \textbf{Predictor models.} The primary temporally separated model is \texttt{gpt-3.5-turbo-0613}
  (cutoff September 2021, predating the $\geq$2024 window). As robustness checks we add \texttt{DeepSeek-V3}
  and \texttt{Qwen2.5-72B} ($\sim$2024 cutoffs, partial leakage) and, as a leaky upper
  bound, \texttt{Claude Sonnet-4.5}; \texttt{Qwen2.5-7B-Instruct} is the weight-level
  LoRA arm (main paper) and \texttt{gpt-oss-20b} the independent impact judge (Table~\ref{tab:respimpact}).
  Headline relevance/impact evaluation uses held-out Human Gold labels; Evidence-Silver labels are produced by a \emph{separate} annotator (Sonnet-4.5, not a predictor) and are used only for additional supervision,
  so no model is scored against its own labels; leakage is audited in Table~\ref{tab:crossmodel}.

  \textbf{Inputs.} A predictor sees only outcome-scrubbed context, the bill's name and summary,
  with outcome tells, expert analysis, gold labels, and evidence quotes withheld: plus the
  role roster (relevance) or related-actor list (impact); antecedents and personas enter
  only where an ablation studies them. Fitted models take outcome-free embeddings of the bill
  title and actor text.

  \textbf{Decision modes and prompts.} Over identical inputs we compare three decision
  modes: \emph{zero-shot} (one joint call), \emph{independent} agents (each actor
  answers alone), and \emph{collaborative} agents (each sees a shared board and revises, or
  communicates over several rounds): plus an in-context \emph{few-shot} variant conditioned
  on $k\in\{0,16\}$ labelled pre-2024 examples. Verbatim prompt templates for every task
  and mode, keyed to the blocks of Table~\ref{tab:results}, are in Appendix~\ref{app:prompts}.

  \subsection{Graph and non-LLM baselines}
  Three families of method consume this representation, all trained (where trained at all)
  on pre-2024 measures and tested on $\ge$2024 (Table~\ref{tab:results}). \emph{(i) LLM}: reads
  the typed context and predicts the scored attributes directly (relevant roles, welfare signs),
  training-free or with in-context few-shot conditioning; the primary predictor. \emph{(ii)
  GNN}, a one-hop bill--actor message-passing network for relevance. \emph{(iii) Graph-structural
  prior}, a rule reading the sign off typed edges (legal-action polarity $\times$
  variable valence $\times$ actor role), fused with the LLM guess (the impact hybrid). Families
  (ii)--(iii) are fitted on the thin sub-2k pre-2024 prior and reported as caveated
  baselines; the graph's primary value is as the representation the LLM reasons over.

  \subsection{Label quality}
Not all labels have the same evidential status. We distinguish \emph{Observed Gold}
(official outcomes and recorded actions), \emph{Human Gold} (human annotation with cited
evidence), \emph{Evidence Silver} (separate-LLM coding constrained by retrieved evidence),
and simulator predictions (Table~\ref{tab:labels}). Headline results use held-out Gold labels.
Silver labels are used only as additional supervision or retrieval context and are evaluated
against Gold on an overlap subset before being admitted to training. For the action task,
\emph{verified-no-public-action} is distinguished from \emph{insufficient-evidence}, so a
missing record is never automatically scored as inaction.

  \section{Experimental setup: fine-tuning, decoding, and tests}
  \label{app:setup}
  \subsection{Fine-tuning and decoding}
  \textbf{Weight-level fine-tuning.} Every LoRA arm uses one recipe, varied only in its
  training file. Base model \texttt{Qwen2.5-7B-Instruct}; LoRA rank $16$, $\alpha{=}32$,
  dropout $0.05$, adapters on all attention and MLP projections
  (\texttt{q,k,v,o,gate,up,down\_proj}), $40.4$M trainable parameters ($0.53\%$ of $7.7$B).
  Training is bf16 with gradient checkpointing, per-device batch size $1$ and gradient
  accumulation $8$ (effective batch $8$), AdamW at learning rate $2\times10^{-4}$, cosine
  schedule with warmup ratio $0.03$, three epochs, maximum sequence length $2{,}048$ tokens.
  Loss is computed on the assistant completion only: prompt tokens are masked to
  $-100$, and over-length examples are left-truncated so the answer and the generation prompt
  survive. Hardware is two NVIDIA RTX 3090 cards ($24$\,GB each) in one workstation with a
  $20$-core Intel i7-12700K and $62$\,GB of RAM; one adapter trains on one card, and the
  $k$-fold arms run two folds concurrently. The host runs Debian~13 (trixie), kernel
  $6.12.73$, with NVIDIA driver $595.58.03$; training runs in a Conda environment on
  Python~$3.12.13$ with \texttt{torch}~$2.12.0$ (CUDA~$13.0$),
  \texttt{transformers}~$5.9.0$ and \texttt{peft}~$0.19.1$. A small arm ($\sim$100 rows) trains
  in under a minute, the largest ($\sim$2{,}200 rows) in about fifteen.

  \textbf{Decoding.} All local evaluation is greedy (\texttt{do\_sample=False}), so the only
  stochasticity in a fine-tuned arm is training. Generation budgets are the smallest that fit
  the answer schema: $16$ new tokens for impact direction, $24$ for the $k$-fold action verb,
  $80$ for the multilabel action list, $90$ for the multi-agent turns. Hosted models are
  called at temperature $0$ through a single API layer; note that this does \emph{not} make
  them deterministic; byte-identical reruns disagree on a fifth of predictions on the
  smallest test.

  \textbf{Repetition.} Every fine-tuned arm is trained several times under independent
  initialisations and reported as the mean, so no single run carries a claim; the spread is
  given in the text where it bears on one. In-context arms resample the exemplar draw and
  report the majority vote over draws. Label construction is made deterministic: an earlier
  tie-break over a set iteration order silently flipped gold impact labels between runs, and a
  fixed hash order with a deterministic tie-break removed it.

  \subsection{Statistical tests and leakage controls}
  \textbf{Statistical tests.} Paired comparisons on the same test items use an exact McNemar
  test over the discordant pairs, reported as $n_{01}/n_{10}$ with its $p$-value; we report
  the counts as well as $p$ because several comparisons have fewer than ten discordant items,
  where a $p$-value alone would overstate what the test can resolve. Interval estimates on
  aggregate scores are bootstrap percentile intervals. We do not report AUC anywhere: the
  targets are decisions at a fixed operating point, and on class-skewed splits AUC would
  flatter arms that never commit to the minority label. Balanced accuracy and macro-F1 are
  used instead, and passage is additionally reported jurisdiction-stratified because a single
  ``is US-federal'' feature outperforms the forecaster on the unstratified register.

  \textbf{Leakage controls.} Splits are temporal everywhere (train $<$2024, test $\ge$2024),
  never random. Persona track records are truncated to strictly-earlier events than the
  example they condition, and exposure-only memories, which carry anachronistic dates, are
  excluded from personas entirely. Silver rows whose (bill, actor) key appears in a gold test
  set are dropped from every training pool. For the annotator-detail feature, the evidence
  text is supplied to exemplars and to the test item in separate arms, and the arm that
  supplies it at test time is labelled as such, since it changes the task from forecasting to
  reading.

  \section{Evidence sources and causal-graph detail}

  \begin{table}[t]
    \centering
    \small
    \setlength{\tabcolsep}{4pt}
    \begin{tabular}{@{}p{0.45\linewidth}p{0.49\linewidth}@{}}
      \toprule Source (website / dataset)                                 & Field it populates                                  \\
      \midrule GovInfo BILLSTATUS bulk                                    & bill text, sponsor, stage; passage label            \\
      OECD.AI policy database                                             & non-US AI policies + adoption status                \\
      US state legislatures                                               & state AI bills + status                             \\
      CRS reports \& expert analyses (law-firm, think-tank, news, \ldots) & antecedent events; per-actor welfare-impact records \\
      Voteview roll-call votes                                            & legislator stance \& behaviour                      \\
      SEC / investor filings                                              & firm resources                                      \\
      Senate LDA lobbying disclosures                                     & actor lobbying positions                            \\
      UN Comtrade                                                         & trade flows (foundry / chip actors)                 \\
      World Bank (WDI)                                                    & country world state (GDP, R\&D, military)           \\
      CSET / ETO (Zenodo)                                                 & country AI capability (papers, patents, investment) \\
      Historic causal studies                                             & graded causal maps (antecedent$\to$bill$\to$impact) \\
      AI 2027 \& AI 2040 (AI Futures Project)                             & scenario-prior persona layer                        \\
      \bottomrule
    \end{tabular}
    \caption{\textbf{Data collection: each source and the benchmark field it populates.} No
    value is fabricated, a reviewer can open the document behind any field. The
    scenario forecasts (bottom row) seed a forward-looking prior that is always weighted
    below the documented record.}
    \label{tab:sources}
  \end{table}

  \begin{figure}[t]
    \centering
    \includegraphics[width=0.9\columnwidth]{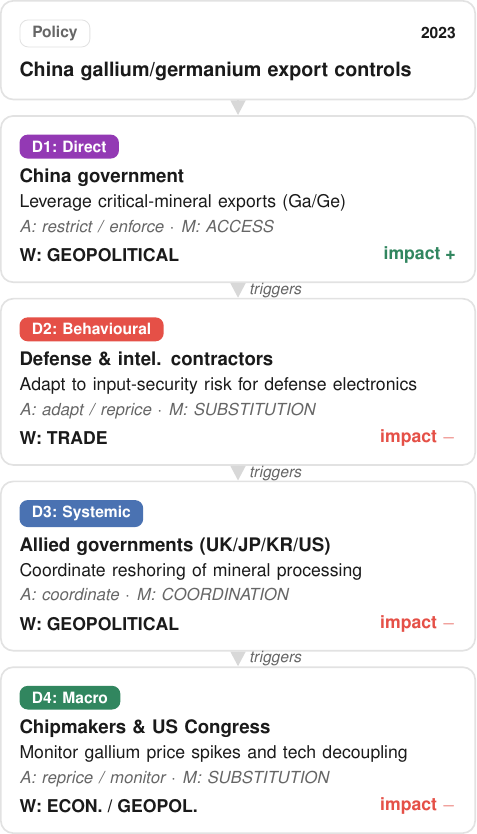}
    \caption{\textbf{An exemplar typed causal graph} from the full corpus: China's 2023 gallium
    and germanium export controls. Each causal depth (D1 direct mandate through D4 macro consequence)
    is a typed actor\,$\to$\,action\,(mechanism)\,$\to$\,world-variable\,$\to$\,impact
    record, and the \emph{triggers} edges chain each stage to the response it provokes: an
    access restriction (geopolitical$+$ for China) propagates to a defense-supply shock, allied
    reshoring, and finally a price spike and technology decoupling. The rail colour encodes
    the world perspective and the sign encodes benefit/harm. Every node derives from a
    verbatim-sourced record.}
    \label{fig:causalgraph}
  \end{figure}

  \begin{table}[t]
    \centering
    \small
    \setlength{\tabcolsep}{4pt}
    \begin{tabular}{@{}llccl@{}}
      \toprule Task                     & Split     & Train   & Test       & Predictor \\
      \midrule Relevance / impact (918) & temporal  & $24$ b. & $157$ b.   & LLM       \\
      Impact direction                  & 5-fold CV &:     & $130$ imp. & hybrid    \\
      \quad same, temporal              & temporal  & $13$ b. & $62$ b.    & hybrid    \\
      \bottomrule
    \end{tabular}
    \caption{\textbf{Train/test for the forecasting experiments.} Temporal $=$ train dated
    $<$2024, test $\geq$2024, not random. LLM predictors are temporally separated by the
    September-2021 model cutoff and are not fitted (``train'' marks only the exclusion boundary);
    the impact hybrid is the only fitted model, reported both under $5$-fold cross-validation
    and out-of-time (sub-$2$k prior, caveated).}
    \label{tab:splits}
  \end{table}

  \section{Extended results tables}
  These tables give the full model$\times$mode breakdowns and cross-model audits summarized
  in the main text (Results).

  \begin{table*}
    [t]
    \centering
    \footnotesize
    \setlength{\tabcolsep}{4pt}
    \begin{tabular}{@{}llccccc@{}}
      \toprule Setting                                                                                                                                                         & Mode / variant & GPT-3.5$^{\ast}$ & DeepSeek             & Qwen          & GNN  & Hybrid \\
      \midrule \multicolumn{7}{@{}l}{\emph{A. Core benchmark}: macro-F1 (relevance base rate $\sim$14\%; impact majority F1 $0.39$)}                                         \\
      Actor relevance (34-way)                                                                                                                                                 & zero-shot      & \textbf{0.39}    & 0.37                 & 0.31          & 0.30 &:    \\
                                                                                                                                                                               & independent    & 0.25             & \textbf{0.28}        & 0.26          &,   &,     \\
                                                                                                                                                                               & collaborative  & 0.25             & \textbf{0.27}        & 0.26          &,   &,     \\
      Impact direction                                                                                                                                                         & zero-shot      & \textbf{0.77}    & 0.71                 & 0.71          &:  & 0.76   \\
                                                                                                                                                                               & independent    & \textbf{0.65}    & 0.60                 & 0.64          &,   &,     \\
                                                                                                                                                                               & collaborative  & 0.64             & 0.61                 & 0.62          &,   &,     \\
      \midrule \multicolumn{7}{@{}l}{\emph{B. Training data in context}: impact direction, few-shot $k{=}0 \rightarrow k{=}16$}                                              \\
                                                                                                                                                                               & zero-shot      & 0.78$\to$0.77    & 0.72$\to$0.77        & 0.70$\to$0.76 &,   &,     \\
                                                                                                                                                                               & independent    & 0.64$\to$0.68    & 0.61$\to$0.63        & 0.62$\to$0.69 &,   &,     \\
                                                                                                                                                                               & collaborative  & 0.64$\to$0.69    & 0.63$\to$0.66        & 0.62$\to$0.68 &,   &,     \\
      \midrule \multicolumn{7}{@{}l}{\emph{C. Agent specification}: impact direction, independent / collaborative}                                                           \\
      Identity persona                                                                                                                                                         & ind.\ / col.\  & 0.64 / 0.65      & 0.61 / 0.60          & 0.56 / 0.62   &,   &,     \\
      Rich persona                                                                                                                                                             & ind.\ / col.\  & 0.70 / 0.73      & 0.45 / 0.51          & 0.64 / 0.62   &,   &,     \\
      Rich persona $+$ few-shot                                                                                                                                                & ind.\ / col.\  & 0.71 / 0.71      & \textbf{0.72 / 0.74} & 0.70 / 0.70   &,   &,     \\
      \midrule \multicolumn{7}{@{}l}{\emph{D. Antecedent events}: impact direction, bill-only / $+$antecedents ($17$ bills that carry them)}                                 \\
                                                                                                                                                                               & zero-shot      & 0.81 / 0.76      & 0.85 / 0.78          & 0.81 / 0.75   &,   &,     \\
                                                                                                                                                                               & independent    & 0.78 / 0.76      & 0.60 / 0.61          & 0.77 / 0.76   &,   &,     \\
                                                                                                                                                                               & collaborative  & 0.78 / 0.78      & 0.72 / 0.62          & 0.84 / 0.73   &,   &,     \\
      \midrule \multicolumn{7}{@{}l}{\emph{E. Earlier 18-role corpus}: F1; \emph{model cols: GPT-3.5 / DeepSeek / Sonnet\,(UB)}}                                             \\
      Relevance (18-way)                                                                                                                                                       & zero-shot      & \textbf{0.66}    & 0.62                 & 0.71          & 0.58 &:    \\
                                                                                                                                                                               & independent    & 0.56             & 0.61                 & 0.59          &,   &,     \\
                                                                                                                                                                               & collaborative  & 0.55             & 0.60                 & 0.63          &,   &,     \\
      Impact direction                                                                                                                                                         & zero-shot      & \textbf{0.79}    & 0.78                 & 0.82          &,   &,     \\
                                                                                                                                                                               & independent    & 0.59             & 0.52                 & 0.73          &,   &,     \\
                                                                                                                                                                               & collaborative  & 0.60             & 0.53                 & 0.71          &,   &,     \\
      \midrule \multicolumn{7}{@{}l}{\emph{F. Release named-actor pool}: impact-direction macro-F1, DeepSeek-V3, $18$ multi-actor post-2023 \emph{gold} bills ($98$ actors)} \\
      Impact direction                                                                                                                                                         & zero-shot      &,               & \textbf{0.71}        &,            &,   &,     \\
                                                                                                                                                                               & independent    &,               & 0.62                 &,            &,   &,     \\
                                                                                                                                                                               & collaborative  &,               & 0.58                 &,            &,   &,     \\
      \bottomrule
    \end{tabular}
    \caption{\textbf{All \benchname prediction results in one table} (blocks A--D on the $3
    4$-role grounded corpus, $153$ post-2024 test bills except D's $17$; block E on the earlier
    $18$-role corpus). The three model columns are $^{\ast}$GPT-3.5 (temporally separated,
    Sept-2021 cutoff), DeepSeek and Qwen ($\ge$2024-cutoff); block E swaps Qwen for
    Claude Sonnet-4.5, a leaky upper bound (see its header). The two trained-baseline
    columns are the paper's non-LLM predictor families, each a single fitted configuration
    with no decision modes (sub-$2$k prior, caveated): \textbf{GNN}, our one-hop bill--actor
    relevance model (block A F1 $0.30$; block E F1 $0.58$), and \textbf{Hybrid}, the graph-structural
    impact-direction model, the bill's legal-action polarity crossed with the measurable
    variable's valence, fused with a DeepSeek per-pair guess via a gradient-boosted tree
    (block A impact F1 $0.76$). Each sits on its one task's row; every other cell is ``: ''.
    \emph{Reading the blocks:} (A)~joint zero-shot wins relevance and impact on every
    model, the multi-agent modes win nothing, and the trained baselines trail the best
    LLM (GNN $0.30$ on relevance; Hybrid $0.76$ on impact vs.\ GPT-3.5 $0.77$, though it
    edges out its DeepSeek base $0.71$); (B)~few-shot conditioning lifts every model and
    mode, most in the multi-agent cells; (C)~a rich persona alone is a gamble, it helps
    GPT-3.5 and Qwen but crashes DeepSeek, while rich$+$few-shot is reliably strong on
    all three; (D)~antecedents do not help impact and hurt most under collaboration on the
    open models; (E)~on the $18$-role corpus zero-shot again wins both tasks and the
    trained GNN (F1 $0.58$) trails every LLM; (F)~the same joint$>$independent$>$collaborative
    ordering reproduces on the human-curated release named-actor pool (DeepSeek-V3
    $0.71/0.62/0.58$), and when the collaborative agents may commit \emph{joint actions} they
    form $42$ record-grounded coalitions across the four richest bills (co-signed
    frameworks, shared funds), the interaction structure a single prediction does not express.
    The cross-model leakage audit is reported in Table~\ref{tab:crossmodel}.}
    \label{tab:results}
  \end{table*}

  \begin{table}[t]
    \centering
    \small
    \begin{tabular}{@{}llcccc@{}}
      \toprule Predictor   & Cond.      & Pass. & Rel.\ F1      & Imp.\ F1      & Imp.\ acc     \\
      \midrule DeepSeek-V3 & none       & 0.55  & 0.39          & 0.56          & 0.59          \\
      DeepSeek-V3          & fine-tuned &:   & 0.42          & 0.58          & 0.58          \\
      Qwen2.5-72B          & none       & 0.51  & 0.40          & \textbf{0.67} & \textbf{0.68} \\
      Qwen2.5-72B          & fine-tuned &:   & \textbf{0.43} & 0.66          & 0.66          \\
      \bottomrule
    \end{tabular}
    \caption{\textbf{All three policy-level targets per open model.} \emph{Rel.\ F1} and
    \emph{Imp.} are actor relevance and impact direction on the full $>$2k corpus ($644$ post-2023
    test bills, $1{,}331$ (bill, role) impact pairs, over the $12$ functional roles both
    the gps918 roster and the auto-crawled corpus collapse into); \emph{Pass.} is legislative-passage
    balanced accuracy on the $329$-bill US register (title-only, leak-free), which is unconditioned
    and so is reported once per model (GPT-3.5 reference $0.55$). \emph{Rel.\ F1} is relevant-class
    F1: recall-dominated, since only impact-recorded roles are annotated positive (DeepSeek
    recall $0.69$/$0.52$, Qwen $0.90$/$0.79$ for none/fine-tuned). \emph{Imp.\ F1} is macro-F1
    of the welfare sign, \emph{Imp.\ acc} its accuracy. \emph{Fine-tuned} $=$ in-context conditioning
    on a \emph{bill-specific} prior: the relevant actors and their welfare signs drawn from
    each bill's nearest prior bills, retrieved by TF-IDF similarity (year $<$ target). Passage
    from the title alone sits near the majority baseline ($0.50$), consistent with the
    jurisdiction confound; relevance and impact are well above their trivial baselines.
    Open-weights predictors only.}
    \label{tab:relimp2k}
  \end{table}

  \begin{table}[t]
    \centering
    \small
    \begin{tabular}{@{}llcccc@{}}
      \toprule Predictor             & Cond.      & mic-F1        & mac-F1        & fam-F1        & no-act        \\
      \midrule majority (all-comply) &,         & 0.12          & 0.01          &,            & 0.00          \\
      \midrule DeepSeek-V3           & none       & 0.15          & 0.14          & 0.22          & 0.02          \\
      DeepSeek-V3                    & fine-tuned & 0.18          & 0.16          & 0.21          & \textbf{0.64} \\
      Qwen2.5-72B                    & none       & 0.13          & 0.12          & 0.21          & 0.05          \\
      Qwen2.5-72B                    & fine-tuned & \textbf{0.25} & \textbf{0.21} & \textbf{0.30} & 0.02          \\
      \bottomrule
    \end{tabular}
    \caption{\textbf{Actor-action multilabel benchmark} ($2{,}466$ test pairs: $1{,}201$ acted
    $+$ $1{,}265$ verified-no-action, over $820$ bills). \emph{Fine-tuned} $=$ in-context
    conditioning on the actor's \emph{full documented track record}: every prior bill
    it acted on (title and action set) plus the aggregate action distribution, temporally
    restricted to earlier bills. \emph{no-act} $=$ recall on the verified-no-action pairs
    (does the agent correctly stay silent). Conditioning improves every model's F1, but only
    DeepSeek learns restraint; Qwen predicts actions best yet over-attributes. Open-weights
    predictors only.}
    \label{tab:action}
  \end{table}

  \begin{table}[t]
    \centering
    \small
    \begin{tabular}{@{}llcccc@{}}
      \toprule Predictor   & History form         & mic-F1        & mac-F1        & fam-F1        & no-act \\
      \midrule DeepSeek-V3 & action repertoire    & 0.19          & 0.17          & 0.23          & 0.60   \\
      DeepSeek-V3          & structural rationale & 0.05          & 0.07          & 0.09          & 0.91   \\
      DeepSeek-V3          & stated motive        & \textbf{0.23} & \textbf{0.19} & \textbf{0.29} & 0.27   \\
      \midrule Qwen2.5-72B & action repertoire    & \textbf{0.24} & \textbf{0.20} & \textbf{0.29} & 0.01   \\
      Qwen2.5-72B          & structural rationale & 0.21          & 0.18          & 0.27          & 0.05   \\
      Qwen2.5-72B          & stated motive        & 0.22          & 0.19          & 0.28          & 0.01   \\
      \bottomrule
    \end{tabular}
    \caption{\textbf{Form of the historical conditioning} on the actor-action benchmark ($2
    {,}466$
    test pairs), all at a fixed depth of ten prior bills. \emph{Action repertoire} $=$
    the terse list of action-types the actor used; \emph{structural rationale} $=$ each
    past action with the world-state variable it targeted, welfare direction, and co-actors;
    \emph{stated motive} $=$ the actor's own-voice first-person reason per past action, written
    by the separate annotator (Sonnet-4.5) grounded in the record. The structural
    rationale hurts both models (DeepSeek collapses into near-total passivity, no-action
    recall $0.91$ with micro-F1 $0.05$); the stated motive is best for DeepSeek but does
    not beat the terse repertoire for Qwen. Open-weights predictors only.}
    \label{tab:actionwhy}
  \end{table}

  \begin{table}[t]
    \centering
    \small
    \begin{tabular}{@{}lcc@{}}
      \toprule Impact measurement                    & macro-F1      & acc           \\
      \midrule Self-report, in-context (DeepSeek)    & 0.56          & 0.59          \\
      Self-report, LoRA fine-tune (7B)               & \textbf{0.85} & \textbf{0.86} \\
      Response $\to$ independent judge (gpt-oss-20b) & 0.44          & 0.62          \\
      \midrule majority-sign baseline                &:           & 0.62          \\
      \bottomrule
    \end{tabular}
    \caption{\textbf{Two ways to measure impact} on the same $1{,}331$ post-2023 (bill, actor)
    pairs. \emph{Self-report} asks the actor whether the bill benefits/harms it (welfare sign);
    \emph{response $\to$ judge} instead predicts the actor's concrete response and has
    an independent open-weights model (gpt-oss-20b, not the responder) infer the impact from
    that response. The response$\to$judge chain drops to the majority-sign baseline on
    welfare-sign recovery, the two-stage pipeline compounds error where the sign is
    directly readable, but yields a richer output: the actor's action plus a signed,
    per-dimension world-impact attribution (aggregate direction governance/rights/safety/innovation-positive,
    matching the corpus world vector).}
    \label{tab:respimpact}
  \end{table}

  \begin{table}[htp]
    \centering
    \small
    \setlength{\tabcolsep}{4pt}
    \begin{tabular}{@{}lccc@{}}
      \toprule Model (knowledge cutoff) & Audit           & Recog.          & Gen.\ recall    \\
                                        & ($/6$)          & (base$\to$+P+A) & (+P+A / none)   \\
      \midrule qwen3:4b (2024)          & $1/6^{\dagger}$ & $0.49{\to}0.70$ & $0.19$ / $0.30$ \\
      gpt-3.5-turbo-0613                & $4/6$           & $0.75{\to}0.88$ & $0.35$ / $0.35$ \\
      Claude Sonnet 4.5 (2025)          & $6/6$           & $0.83{\to}0.80$ & $0.55$ / $0.45$ \\
      \bottomrule
    \end{tabular}
    \caption{\textbf{Cross-model robustness and leakage.} Audit: number of $2023{+}$
    outcome facts recovered by direct probes (of 6); every reachable model leaks, from
    \texttt{gpt-3.5-turbo-0613} at $4/6$ to the frontier model at $6/6$. Recognition: 4-option
    accuracy over each bill's real generated events, no-persona $\to$ persona$+$antecedents
    (P$+$A). Generation: generated-action recall, P$+$A vs.\ a neutral analyst. The persona
    effect is \emph{model-dependent}: grounding helps recognition on weaker models but is
    saturated on Sonnet ($0.83{\to}0.80$, flat-to-slightly-negative), and helps generation
    only on the strongest model, where, being fully contaminated, it is largely
    memorisation retrieval rather than forecasting. $^{\dagger}$qwen recovers only $1/6$
    on direct probes yet reproduces post-cutoff events at high coverage ($0.89$), so a low
    direct-probe count does not certify a model clean.}
    \label{tab:crossmodel}
  \end{table}

  \section{Additional benchmark and results figures}

  \begin{figure}[t]
    \centering
    \includegraphics[width=\columnwidth]{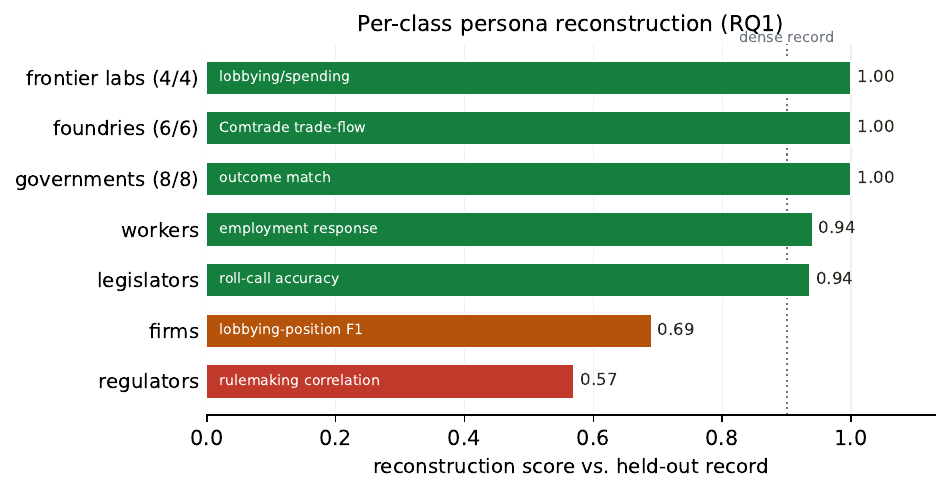}
    \caption{\textbf{Per-class persona reconstruction} against held-out records (RQ1).
    Classes with dense, structured records (governments, foundries, frontier labs, workers,
    legislators) reconstruct accurately; classes whose behaviour is sparser or more
    discretionary (firms, regulators) are weaker. The per-class metric is shown inside
    each bar.}
    \label{fig:recon}
  \end{figure}

  \begin{figure}[t]
    \centering
    \includegraphics[width=\columnwidth]{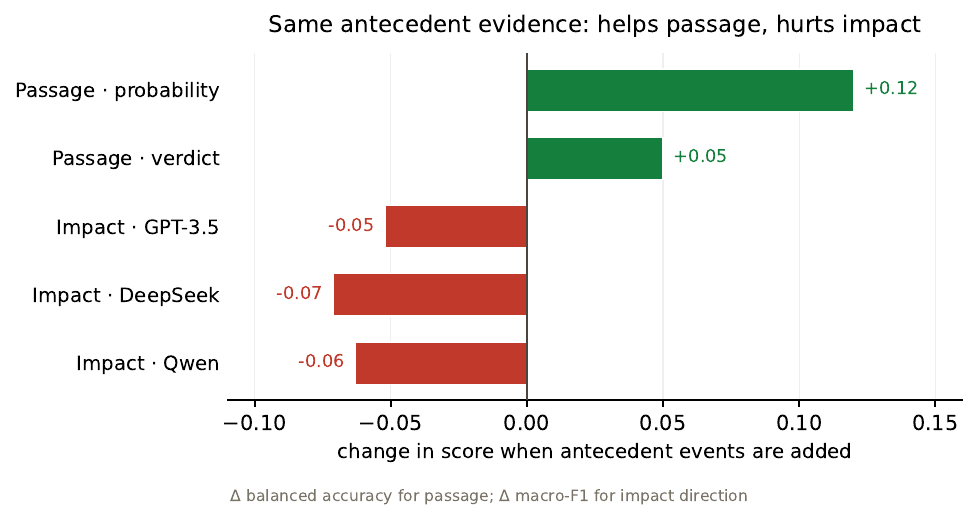}
    \caption{\textbf{Evidence must match the grain of the target.} Adding the same antecedent
    events improves legislative passage prediction (green; balanced accuracy, temporally
    separated model, calibrated and verdict prompting) but degrades impact-direction
    prediction (red; macro-F1, zero-shot) across all three models. The passage forecast depends
    on political timing, which antecedents supply; the welfare sign is structural, for
    which antecedents are distractor context.}
    \label{fig:grain}
  \end{figure}

  \begin{figure}[t]
    \centering
    \includegraphics[width=\columnwidth]{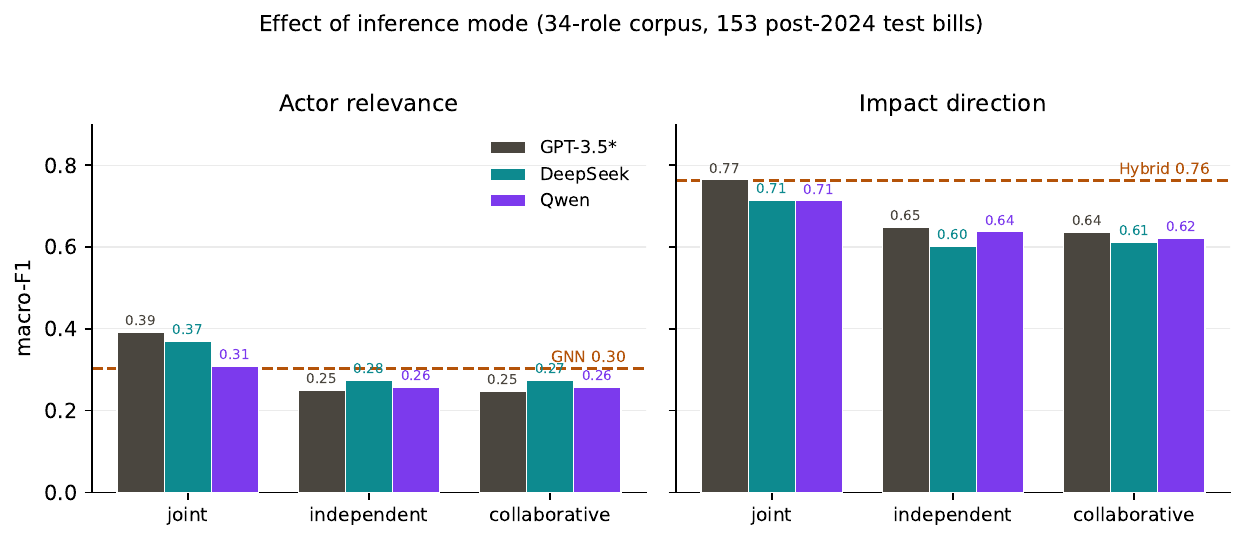}
    \caption{\textbf{Effect of inference mode} (macro-F1, $34$-role corpus, $153$ post-2024
    test bills). Joint inference is strongest on both tasks for every model; decomposing into
    independent or collaborative per-actor agents removes cross-actor context and lowers performance.
    The trained non-LLM baselines (GNN on relevance, graph-structural Hybrid on impact; dashed)
    trail the joint LLM.}
    \label{fig:modes}
  \end{figure}

  \begin{table}[t]
    \centering
    \small
    \begin{tabular}{@{}ll@{}}
      \toprule Node type        & Output signature                              \\
      \midrule Antecedent event & event, date, provenance grade                 \\
      Policy provision          & mechanism, scope, status                      \\
      Actor state               & role, resources, stance                       \\
      Actor action              & type (15-way), target, probability            \\
      Mechanism                 & family (12-way)                               \\
      World-state change        & dimension (10-way), variable, sign            \\
      Impact                    & affected actor, variable, sign $\pm$, horizon \\
      \bottomrule
    \end{tabular}
    \caption{\textbf{Node schema.} Every vertex is a typed function $f:\,$in-edges $\to$
    out-edges with the output signature shown; every edge carries $\langle$relation, variable,
    sign $\pm$, probability, lag$\rangle$ and a provenance tag binding it to a named source
    field.}
    \label{tab:schema}
  \end{table}

  \section{Multi-agent interaction and extended analysis}
  \subsection{Multi-agent interaction (RQ4)}
\begin{figure}[t]
    \centering
    \includegraphics[width=\columnwidth]{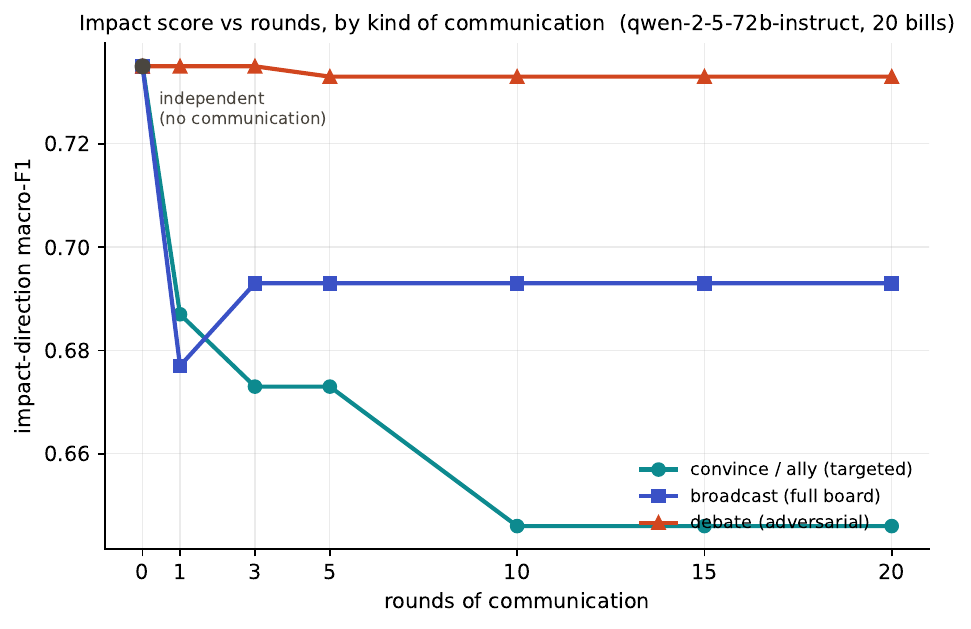}
    \caption{\textbf{Impact-direction macro-F1 vs.\ rounds of communication, by kind} (open-arm
    Qwen2.5-72B, $20$ grounded bills; the ordering reproduces on DeepSeek). All three regimes
    start at the independent, no-communication score ($0.735$) and none rises above it: targeted
    \emph{convince/ally} erodes to $0.65$, \emph{broadcast} settles at $0.69$, and adversarial
    \emph{debate} holds at $0.73$. Communication protocol has a larger effect on performance
    than communication depth within the evaluated range.}
    \label{fig:commkinds}
  \end{figure}

  \textbf{Effect of communication depth.} Figure~\ref{fig:commkinds} reports impact-direction
  performance over $0$--$20$ communication rounds for the two open arms on all $151$
  multi-actor test bills. Balanced accuracy improves during the first five rounds and then
  plateaus (DeepSeek $0.59\!\to\!0.65$, Qwen $0.62\!\to\!0.65$), whereas macro-F1 shows no
  consistent improvement over the independent-agent condition (DeepSeek peaks near round
  ten at $0.62$ then declines to $0.61$; Qwen remains near its independent value). Stance
  churn continues to rise with additional rounds (DeepSeek $0.19\!\to\!0.29$).
  Additional communication therefore increases interaction without a corresponding
  improvement in predictive performance.

  \textbf{Communication protocol ablation.} Fixing the mechanism at communication rather than
  depth, we compare three protocols over $20$ rounds on the open arms (Figure~\ref{fig:commkinds}):
  targeted \emph{convince/ally}, full-board \emph{broadcast}, and adversarial \emph{debate}.
  None exceeds the independent baseline ($0.735$). Convince/ally progressively degrades
  macro-F1 ($-0.09$ by round ten), broadcast settles just below baseline ($-0.04$), and
  debate remains approximately stable (churn $2$--$4\%$). Across both models, adversarial
  debate preserves impact-direction performance more effectively than targeted or broadcast
  communication; the communication protocol matters more than the number of rounds.

  \subsection{Analysis: mechanisms behind the results}
  The preceding section reports \emph{what} each mode predicts; here we explain \emph{why},
  through the interaction structure the agents expose, the grounding that drives actor
  prediction, and a qualitative case.

\begin{figure}[t]
    \centering
    \includegraphics[width=\columnwidth]{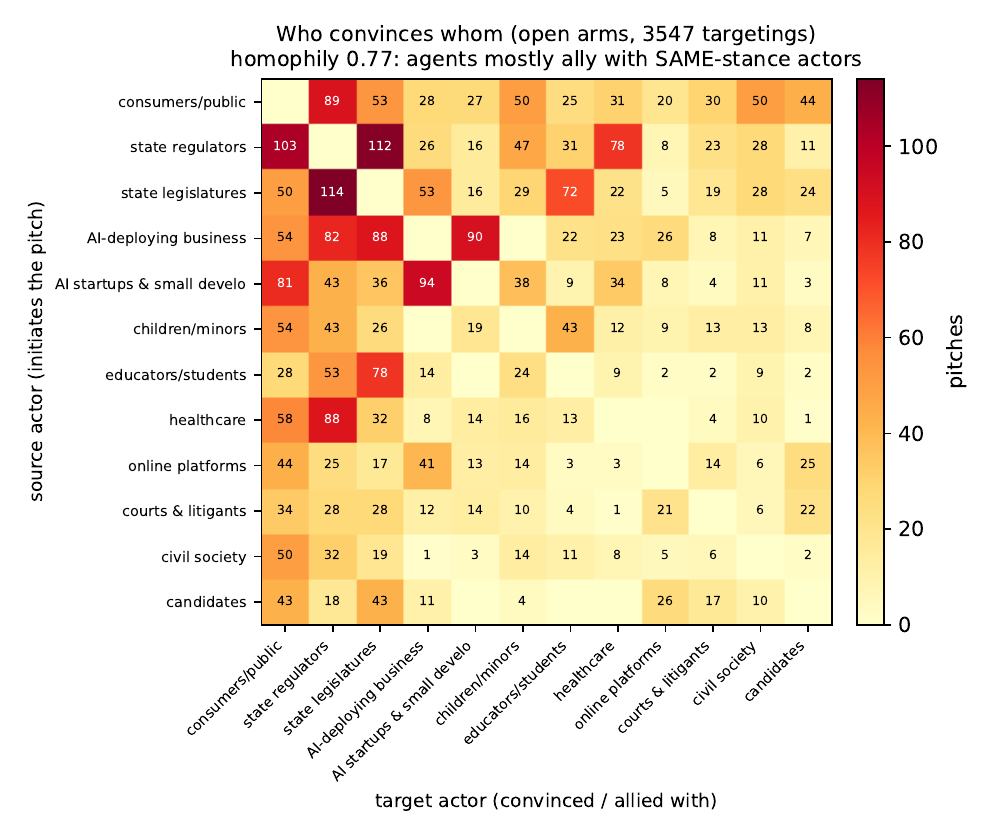}
    \caption{\textbf{Who agents choose to convince/ally with} (open arms, $7{,}138$ targetings
    from $4{,}254$ pitches over the $3$-round communication; cell $=$ pitches from the row
    actor to the column actor, top-$12$ actors by activity). Agents overwhelmingly target \emph{same-stance}
    actors (homophily $0.77$): they build coalitions rather than persuade opponents. The
    heaviest edges are a government/regulatory core (state legislatures $\leftrightarrow$
    regulators) and an industry bloc (business $\leftrightarrow$ startups), with consumers/the
    public the most-courted target.}
    \label{fig:coalition}
  \end{figure}

  \textbf{Coalition formation.} To characterise what communication does, we instrument the
  convince/ally step (Figure~\ref{fig:coalition}). Across $4{,}254$ pitches from the two open
  arms, an agent almost always selects a target ($1$--$4\%$ pitch to none), and $77\%$ of
  targets already hold the \emph{same} stance as the source (homophily $0.77$ pooled, $0.70$
  on DeepSeek-V3 and $0.84$ on Qwen-72B):
  agents predominantly reinforce same-stance coalitions rather than convert opponents.
  The heaviest edges form a government/regulatory core (state legislatures $\leftrightarrow$
  regulators) and an industry bloc (business $\leftrightarrow$ startups). This coalition-reinforcing
  topology is consistent with the absence of accuracy gains from additional communication
  rounds.

\begin{figure}[t]
    \centering
    \includegraphics[width=\columnwidth]{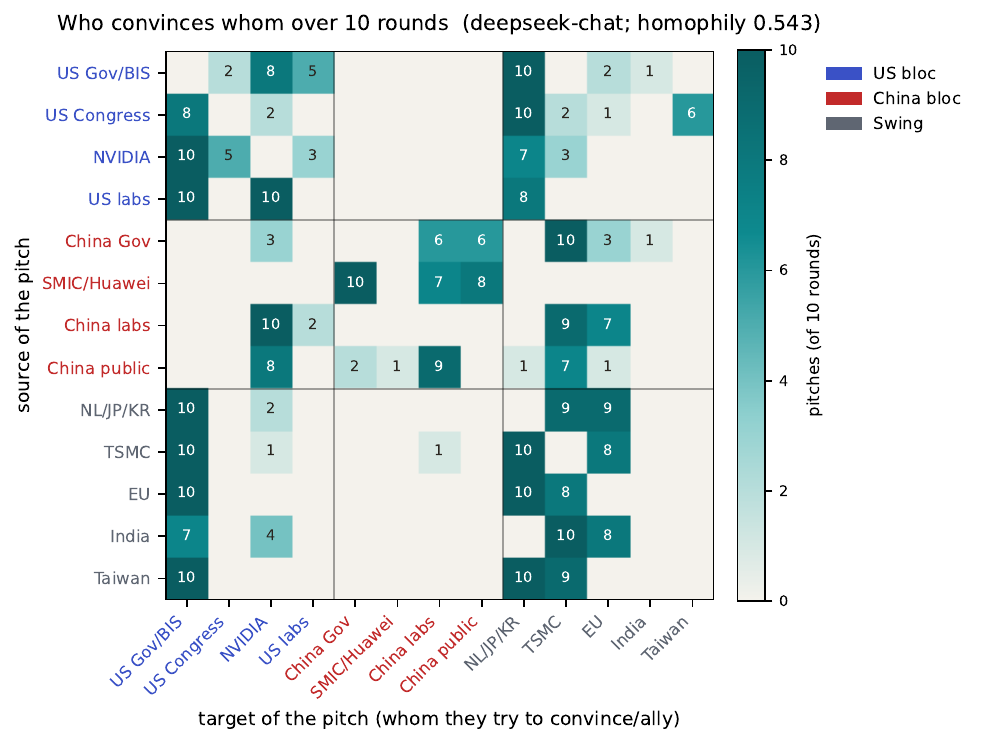}
    \caption{\textbf{Who convinces whom in the US--China chip-control deliberation} ($13$
    record-grounded agents ordered U.S.\,$\vert$\,China\,$\vert$\,swing; cell $=$
    pitches from the row actor to the column actor over $10$ rounds; DeepSeek). Pitches cluster
    within blocs (homophily $0.54$; $0.72$ under Qwen), and the heaviest cross-bloc column
    is the U.S.\ chip-seller (NVIDIA), lobbied by Chinese labs, users and government
    alike. Almost no stance is converted ($1/370$ moves), so the structure is coalition
    maintenance, not persuasion.}
    \label{fig:uschina}
  \end{figure}

  \textbf{Case study: semiconductor export controls.} We instantiate $13$ evidence-grounded
  actors associated with the 2022 U.S.\ semiconductor export controls to examine whether
  the communication topology observed at scale also appears in a policy-specific
  simulation. Across both open models, ten communication rounds produce stable coalition
  structure with limited stance conversion (Figure~\ref{fig:uschina}). Joint-action simulation
  similarly favours within-bloc coordination unless cross-bloc interaction is explicitly required.
  This case illustrates how actor-decomposed inference provides interaction-level
  outputs that complement, rather than replace, the stronger aggregate predictions from
  joint inference; the detailed coalition analysis is in Appendix~\ref{app:casestudy}.

  \textbf{When they can \emph{act}, not just argue, coalitions still stop at the bloc line.}
  A stance debate only exchanges words, so we change the mechanism to elicit \emph{joint
  action}: each of the $13$ agents first names the move it would take \emph{alone}, then
  proposes concrete joint actions to named partners: co-signed proposals, reciprocal deals,
  shared funds and fabs, which partners accept, counter, or reject, with accepted proposals
  becoming committed coalitions. Over six rounds the agents commit $77$ coalitions and
  every agent joins one, yet \emph{not a single coalition pairs a U.S.\ actor with a Chinese
  one}. The joint actions are moreover a different \emph{kind} of object than the solo
  moves: alone an actor tweaks a threshold or lobbies, but together the roster pools
  capital into two rival build-outs, a Western secure-supply bloc (a China-barred Arizona
  fab, a U.S.--Taiwan packaging line, an allied resilience pact, a tiered DUV framework) and
  a Chinese self-reliance bloc (a \$50B national chip fund, a \$20B sovereign accelerator
  fund, a joint open-model ecosystem), with the swing states consolidating into the Western
  one. This sharpens the earlier lesson: shared \emph{preference} crosses the divide (in the
  debate every commercial actor wanted ``calibrate''), but shared \emph{commitment} does
  not, so deliberation among grounded actors manufactures rival coalitions rather than a common
  solution. The bloc line is a preference, not an inability: when the mechanism instead
  \emph{requires} each agent to include a rival partner, $37$ of $72$ committed
  coalitions cross it, and the deals converge on a managed-competition regime, a mutually
  verified sub-threshold ``safe-harbor'' chip tier, a legacy-node ($28$nm$+$) carve-out
  with a joint R\&D fund, and a market-access-for-ending-retaliation truce, while the frontier
  is not put on the table. The agents can bargain across the divide; left to themselves they
  simply prefer not to.

  \textbf{Effect of actor grounding.}\label{sec:grounding} Actor descriptions alone do not
  reliably improve multi-agent prediction. On impact direction (independent mode, all
  $151$ test bills; Table~\ref{tab:results}, block~C), rich interest profiles improve
  performance for GPT-3.5 ($0.64\!\to\!0.70$) and Qwen ($0.56\!\to\!0.64$) but
  substantially reduce it for DeepSeek ($0.61\!\to\!0.45$). Conditioning the same actor representations
  on pre-cutoff behavioural examples removes much of this cross-model instability and yields
  consistently stronger performance across models and inference modes (up to $0.74$; Table~\ref{tab:results}
  (block C)). This result distinguishes \emph{actor grounding} from persona complexity: adding
  descriptive detail is insufficient unless the agent is anchored to observed behaviour.

  \textbf{Context aggregation.} The joint advantage follows from context aggregation. A single
  call observes the whole bill and all $34$ candidate actors, exploiting the relational
  structure both tasks depend on: welfare direction is relative, a measure restricting platforms
  simultaneously benefits regulators, consumers, and courts, so splitting the cast into self-focused
  agents drops impact from $0.77$ to $0.65$, and communication recovers part of it by re-injecting
  shared context. For relevance, a self-interested agent over-claims that it is affected,
  answering ``yes'' far more than the true $\sim\!5$-of-$34$ rate and collapsing precision
  ($0.39\!\to\!0.25$), whereas the joint call calibrates across the roster. Multi-agent structure
  therefore helps only when the target is a joint property of the cast.

  \section{US--China semiconductor case study}
  \label{app:casestudy} This appendix expands the case study reported in the Results. We
  instantiate $13$ record-grounded actors around the 2022 U.S.\ advanced-chip export controls:
  the U.S.\ and Chinese governments, U.S.\ chipmakers (NVIDIA) and frontier labs,
  Chinese foundries (SMIC/Huawei), Chinese AI labs and public, and swing states (the ASML/Japan/Korea
  toolmakers, TSMC, the EU, India, and Taiwan). Each actor is conditioned on its documented
  record, motivation (objectives, red-lines, AI-stance), dated antecedents, and a
  scenario prior. All runs use open-arm models (DeepSeek-V3, Qwen2.5-72B).

  \textbf{Deliberation.} Over $10$ rounds of convince/ally, stances barely move ($1/370$
  moves under DeepSeek, $13/390$ under Qwen). The emergent cleavage follows interest rather
  than nationality: the two U.S.\ political actors hold ``expand,'' whereas a cross-bloc
  commercial coalition: U.S.\ and Chinese chipmakers and labs, the toolmakers, TSMC,
  the EU, and India: converges on ``calibrate.'' Two actors keep the controls because
  they benefit: China's government, which reframes containment as a self-reliance
  mandate, and Taiwan, whose ``silicon shield'' depends on its fabs remaining
  indispensable. Every cross-bloc pitch targets the U.S.\ chip-seller, lobbied to loosen thresholds.

  \textbf{Joint action.} When agents propose and commit to concrete joint actions, $77$ coalitions
  form but none crosses the US--China line; the roster instead pools capital into two rival
  build-outs, a Western secure-supply bloc (a China-barred Arizona fab, a U.S.--Taiwan packaging
  line, an allied resilience pact, a tiered DUV framework) and a Chinese self-reliance
  bloc (national chip funds and a joint open-model ecosystem), with the swing states
  joining the Western one.

  \textbf{Forced cross-bloc bargaining.} Requiring each agent to include a rival partner yields
  $37$ of $72$ cross-bloc coalitions, converging on a managed-competition regime: a mutually
  verified sub-threshold ``safe-harbor'' chip tier, a legacy-node ($28$nm$+$) carve-out
  with a joint R\&D fund, a market-access-for-ending-retaliation truce, and a compute-for-safety
  exchange, while the advanced frontier is not traded. The bloc line is thus a
  preference rather than an inability.

  \section{Worked dossier: the grounded record for one bill}
  \label{app:dossier} Before the predictive walkthrough (Appendix~\ref{app:walkthrough}),
  we show the \emph{gold} it is scored against: the benchmark's record for a bill is,
  per reached actor, its role, its action, a documented response with source and evidence
  grade, the \emph{mechanism} by which the instrument reaches it (the exposure channel),
  a \emph{counterfactual} stating what the instrument actually changed, and the welfare
  \emph{sign} of the impact on that actor. Table~\ref{tab:dossier} gives seven of the
  twelve actors reached by the \textbf{Chips and Science Act}; every cell is copied from
  the record (a browsable dossier per bill is generated by \texttt{gen\_bill\_dossier.py}).

  \begin{table*}
    [t]
    \centering
    \small
    \setlength{\tabcolsep}{5pt}
    \begin{tabular}{@{}p{2.5cm}p{4.4cm}cp{6.0cm}@{}}
      \toprule Actor (role)                             & Documented response                                                       & Sign & Why: mechanism / counterfactual                                                                                                \\
      \midrule NVIDIA \newline (excluded\_from)         & Marks the first domestic Blackwell wafer at TSMC Arizona                  & $0$  & Credit written for \emph{facilities, not fabless designers}; absent CHIPS its leading-edge capacity stays concentrated in Taiwan. \\
      TSMC Arizona \newline (benefits\_from)            & Accepts a \$6.6B direct-funding award (+\$5B loans)                       & $+$  & Direct incentive $+$ loan authority for US fabs; the Act's contribution is read as scale and schedule, not existence.             \\
      US Commerce \newline (implementer)                & Stands up the CHIPS office; \$39B NOFO                                    & $+$  & The Act appropriates \$52.7B and hands Commerce the job of designing and policing the incentives.                                 \\
      Intel \newline (benefits\_from)                   & Grants restructured into a $\sim$10\% US-government equity stake          & $-$  & Received the money but at the cost of independence; the Act changed \emph{who holds the equity}, not the build timetable.         \\
      EU \newline (lobbied\_against)                    & EU Chips Act (Reg.\ 2023/1781) enters into force                          & $-$  & Policy diffusion: a US statute produced a matching EU statute within thirteen months,  a defensive subsidy.                     \\
      Japan / Korea / China \newline (lobbied\_against) & Rapidus (\$3.9B); ``supercluster'' (\textrm{\textwon}622tn); Big Fund III & $-$  & Counter-subsidy: rivals commit public funds to hold domestic capacity against a foreign subsidy.                                  \\
      Huawei, ZTE \newline (excluded\_from)             & No public action on record                                                &,   & Coded absence,  a verified non-response, not missing evidence.                                                                  \\
      \bottomrule
    \end{tabular}
    \caption{\textbf{Grounded dossier for one bill} (Chips and Science Act; 7 of 12 reached
    actors, verbatim from the record). Each actor carries a response (with source/grade,
    omitted here), an impact \emph{sign}, and the \emph{why}: the exposure mechanism and
    a counterfactual. The signs are relational,  the same statute that benefits domestic
    fabs and the implementing agency ($+$) triggers rival counter-subsidies abroad ($-$),
    and records two coded non-responses.}
    \label{tab:dossier}
  \end{table*}

  \textbf{World-state roll-up.} Actor-level impacts aggregate to a bill-level \emph{world-state
  vector} over the benchmark's world axes: each impact's variable maps to one of ten axes
  (economic/fiscal, labour, innovation, competition, safety, rights, governance, trade,
  geopolitical, environment), and signs are summed per axis. For the Chips and Science
  Act the vector is net \emph{economic/fiscal} $-3$, \emph{labour} $+2$, \emph{trade} $+1$,
  \emph{governance} $+1$, \emph{geopolitical} $-1$: concentrated domestic industrial
  gains, but a net-negative \emph{global} fiscal read once every rival's counter-subsidy
  is counted,  the subsidy race made quantitative. The same roll-up over China's Interim
  Measures for Generative AI is dominated by \emph{governance capacity} $+13$ (thirteen firms
  file registrations or launch only after security clearance), the signature of a licensing
  regime rather than an industrial-policy one. This is the aggregate object a bill-level
  impact label would flatten away, and it is read directly off the per-actor record.

  \section{Worked example: one bill through each inference mode}
  \label{app:walkthrough} This appendix walks a reviewer, step by step, through what GPS produces
  for a \emph{single} input bill under each inference mode, so that the abstract schema of
  the main paper becomes concrete. We use a real run of the record-grounded
  actor system on the \textbf{2022 U.S.\ advanced-chip export controls} (BIS, 7 Oct 2022;
  bar advanced AI chips and the tools to make them from China), with ten evidence-grounded
  actors. Every stance, pitch, and movement count below is taken verbatim from the logged
  transcript (\texttt{sim\_uschina\_deliberation.json}); nothing here is illustrative-only.

  \textbf{Step 0: Input.} The system receives the outcome-scrubbed bill context
  $\langle$\textsc{ctx}$\rangle$ and the evidence-grounded graph $G_{\le t}(b)$: the ten
  actors, each with a dated state (role, prior positions, resources, relationships)
  restricted to information available before 7 Oct 2022. All three modes below see \emph{this
  same graph}; they differ only in what each LLM call is shown. The required output schema
  is, for every actor, a \emph{response} (the action it takes) and the welfare \emph{direction}
  of the bill's impact on it ($+$ benefits / $-$ harms), which together populate
  $\hat G^{\ast}(b)$.

  \textbf{Step 1: Independent mode ($f_{\mathrm{ind}}$).} Each actor is queried \emph{alone}
  from its own persona, with no sight of any other actor. It returns its response and
  its self-impact in one pass. Table~\ref{tab:walkthrough} (columns 2--3) gives all ten
  outputs: the two U.S.\ political actors take \emph{expand} and read the bill as beneficial
  to them; the commercial actors on both sides take \emph{calibrate} (seek looser
  thresholds) and read it as harmful (e.g.\ U.S.\ chipmakers, consistent with NVIDIA's later
  \$5.5B H20 write-down); China's government takes \emph{maintain}, reframing containment
  as a self-reliance mandate it benefits from, as does Taiwan, whose ``silicon shield'' depends
  on its fabs staying indispensable. This mode yields the per-actor predictions directly,
  but no actor's answer is informed by any other's.

  \textbf{Step 2: Collaborative mode, convince/ally regime ($f_{\mathrm{collab}}$).}
  The \emph{same} ten actors now run over communication rounds. In each round every
  actor reads a shared board of the others' current positions, chooses which actors to
  \emph{convince or ally with}, sends a targeted argument, and may revise. This exposes
  structure the independent pass cannot represent. A representative real pitch:
  \begin{quote}
    \small \textbf{U.S.\ chipmakers} $\to$ \textbf{U.S.\ Executive/BIS}: ``Current
    controls push China to develop domestic alternatives, eroding the U.S.\ lead. Adjust thresholds
    to allow legal exports while maintaining security.''
  \end{quote}
  Over ten rounds the actors exchange $270$ such pitches. The interaction has clear topology: pitches cluster within blocs (homophily $0.48$) and the most-lobbied target is the U.S.\ chip-seller,
  courted across the divide to loosen thresholds (Table~\ref{tab:walkthrough}, column~4).
  But the \emph{predictions do not change}: \textbf{$0$ of $270$ pitches move a stance}; every
  actor's final response and impact direction equal its independent answer (column~5).
  Persuasion is bounded by material interest,  actors argue vigorously and build same-stance
  coalitions, but interest-anchored positions do not convert. (Fixing depth and varying the
  protocol reproduces this: targeted convince/ally in fact \emph{erodes} macro-F1 relative
  to acting independently, Figure~\ref{fig:commkinds}.)

  \textbf{Step 3: Joint mode ($f_{\mathrm{joint}}$), for contrast.} A single call
  sees the whole bill and all ten actors at once and emits every actor's response
  together. Because welfare direction is a \emph{relative} property of the cast (the same
  rule that harms sellers helps the containment coalition), this shared view calibrates the
  signs across the roster and gives the strongest aggregate score of the three modes (main paper);
  the decomposed modes trade that cross-actor context for an explicit, inspectable interaction
  record.

  \textbf{What each setting buys the reviewer.} Independent mode is the per-actor
  predictor: one response and one impact sign per actor, cheaply and in parallel. The collaborative/convince
  setting adds an interaction layer,  who lobbies whom, which coalitions form, what is offered: that is valuable as a \emph{model of the politics}, but on this bill (and in
  aggregate, Figure~\ref{fig:commkinds}) it does not improve the forecast, because grounded
  actors reinforce coalitions rather than convert. Joint mode gives the best aggregate
  accuracy by pooling context but hides that structure. The three are therefore complementary
  readings of one graph, not competing architectures: choose the mode by whether the question
  is \emph{what each actor does}, \emph{how they try to move each other}, or \emph{the
  single most accurate world state}.

  \begin{table}[t]
    \centering
    \small
    \setlength{\tabcolsep}{4pt}
    \begin{tabular}{@{}lccp{2.5cm}c@{}}
      \toprule                                   & \multicolumn{2}{c}{Independent ($f_{\mathrm{ind}}$)} & \multicolumn{2}{c}{Collaborative / convince ($f_{\mathrm{collab}}$)} \\
      \cmidrule(lr){2-3}\cmidrule(lr){4-5} Actor & Response                                             & Impact                                                              & Pitches to            & Final (moved?)   \\
      \midrule U.S.\ Exec / BIS                  & expand                                               & $+$                                                                 & (most-lobbied target) & expand \,(no)    \\
      U.S.\ Congress                             & expand                                               & $+$                                                                 & allies, BIS           & expand \,(no)    \\
      U.S.\ chipmakers                           & calibrate                                            & $-$                                                                 & BIS, Congress         & calibrate \,(no) \\
      U.S.\ frontier labs                        & calibrate                                            & $-$                                                                 & BIS, chipmakers       & calibrate \,(no) \\
      China gov / MOFCOM                         & maintain                                             & $+$                                                                 & foundry, labs         & maintain \,(no)  \\
      SMIC / Huawei (fab)                        & calibrate                                            & $-$                                                                 & China gov             & calibrate \,(no) \\
      Chinese AI labs                            & calibrate                                            & $-$                                                                 & chip-seller           & calibrate \,(no) \\
      Chinese users / public                     & calibrate                                            & $-$                                                                 & chip-seller           & calibrate \,(no) \\
      Allied toolmakers$^{\dagger}$              & calibrate                                            & $-$                                                                 & BIS, chip-seller      & calibrate \,(no) \\
      TSMC / global supply                       & calibrate                                            & $+$                                                                 & BIS                   & calibrate \,(no) \\
      \bottomrule
    \end{tabular}
    \caption{\textbf{One bill through two inference modes} (2022 U.S.\ chip export controls;
    real transcript). Independent mode returns each actor's response and self-impact
    from its persona alone. The collaborative/convince regime runs the same ten actors over
    ten rounds of targeted persuasion ($270$ pitches, within-bloc homophily $0.48$, the
    U.S.\ chip-seller the most-lobbied cross-bloc target),  yet \emph{$0/270$ pitches
    change a stance}, so every final response equals the independent one. Interaction adds
    structure, not a different forecast, when actors are interest-anchored. $^{\dagger}$ASML/Japan/Korea.}
    \label{tab:walkthrough}
  \end{table}

  \section{Prompts and context per evaluation mode}
  \label{app:prompts} Each paragraph below gives the exact prompt for one setting of the consolidated
  results table (Table~\ref{tab:results}), keyed to its block. All settings share the same
  \emph{outcome-scrubbed context}: the predictor sees only the bill name and summary,
  with outcome tells (\texttt{enacted}, \texttt{vetoed}, \texttt{signed into law}, \ldots)
  scrubbed, and not the expert analysis, the gold label, or the evidence quote. We write
  $\langle$\textsc{ctx}$\rangle$ for that context and $\langle$\textsc{roster}$\rangle$
  for the $34$ actor labels. The default \emph{identity persona} is only the actor's
  identity: ``\texttt{You ARE "\{actor\}", an actor in AI governance, reasoning only
  from your own interests.}'', and the \emph{rich persona} of block C replaces it (defined
  below). No setting is given any actor's gold direction or relevance. The templates are
  verbatim skeletons (JSON reply formats abbreviated).

  \paragraph{Block A: Zero-shot (one joint call).}
  \begin{itemize}
    \itemsep2pt

    \item \textbf{Relevance:} ``\texttt{A neutral policy analyst reviews this AI bill: $\langle$ctx$\rangle$.
      Which of these actors are MATERIALLY affected (bear costs, gain/lose, must change behaviour)?
      Actors: $\langle$roster$\rangle$. JSON: \{"related":[...]\}}''

    \item \textbf{Impact direction:} ``\texttt{$\ldots$For each actor, is the impact
      BENEFICIAL(+) or HARMFUL($-$) to it? Actors: $\langle$gold-related actors$\rangle$.
      JSON: \{"dir":\{"<actor>":"+/$-$"\}\}}''
  \end{itemize}

  \paragraph{Block A: Independent agents (each actor answers alone, no communication).}
  \begin{itemize}
    \itemsep2pt

    \item \textbf{Relevance:} ``\texttt{$\langle$persona$\rangle$ A bill is proposed: $\langle$ctx$\rangle$.
      Are YOU materially affected? JSON: \{"affected": true/false\}}''

    \item \textbf{Impact:} ``\texttt{$\langle$persona$\rangle$ Bill: $\langle$ctx$\rangle$.
      Does this bill BENEFIT(+) you (funding, market, protection, lighter burden) or
      HARM($-$) you (cost, liability, lost access, restriction)? JSON: \{"dir":"+/$-$"\}}''
  \end{itemize}

  \paragraph{Block A: Collaborative, single revision (each actor sees the board once, then
  revises).}
  Round 1 is the independent call above. Round 2: ``\texttt{$\langle$persona$\rangle$
  Bill: $\langle$ctx$\rangle$. Other parties' first read: $\langle$board of all actors'
  round-1 answers$\rangle$. In light of the others, [are YOU affected? / your sign?] JSON:
  \{$\ldots$\}}'' (Blocks C and D also run this single-revision collaborative call, with
  the rich persona and the antecedent block respectively.)

  \paragraph{Collaborative, 3-round communication (convince/ally, then update; Table~\ref{tab:results}
  and Figures~\ref{fig:commkinds}--\ref{fig:coalition}).}
  Shared independent init, then three rounds of two steps each. \emph{Convince/ally:} ``\texttt{$\langle$persona$\rangle$
  Bill: $\langle$ctx$\rangle$. Positions: $\langle$board$\rangle$. You want others to
  adopt your reading and may ally. Whom do you CONVINCE/ALLY with, and what do you argue?
  JSON: \{"targets":[...],"message":"<one sentence>"\}}'' \emph{Update:} ``\texttt{$\langle$persona$\rangle$
  Bill: $\langle$ctx$\rangle$. Others' positions: $\langle$board$\rangle$. Messages to you:
  $\langle$inbox$\rangle$. Weighing their arguments/alliances, your UPDATED sign. JSON:
  \{$\ldots$\}}'' Only the messages actors send each other carry influence; no external information
  enters, so the regime isolates the effect of communication itself.

  \paragraph{Block B: Few-shot (in-context conditioning on the training data; also
  Table~\ref{tab:results}).}
  The block-A impact prompt of the chosen mode, prefixed with a labelled-example block: ``\texttt{Labelled
  examples (actor, bill $\to$ welfare impact on that actor): $\langle K$ balanced
  examples drawn from the pre-2024 train split, each \texttt{- actor: \{a\}; bill: \{text\}
  -> +/$-$}$\rangle$}'' followed by the usual per-actor impact question. Every example is
  dated before $2024$ and the test bill is $\geq$2024 and bill-disjoint from the
  examples, so the conditioning is temporally separated. Reported at $k{=}0$ (no examples)
  and $k{=}16$.

  \paragraph{Block C. Agent specification: rich persona (independent / collaborative).}
  Identical to the block-A independent and single-revision collaborative impact prompts,
  except the identity persona is replaced by the \emph{rich interest-persona}: ``\texttt{You
  ARE \{actor\} -- \{what the actor does; what it wants; what it fears\}}'' (e.g.\ ``\texttt{insurers
  -- use AI in underwriting, claims and prior-authorization; face anti-discrimination and
  disclosure duties; want actuarial freedom, fear mandates}''). The \emph{rich $+$ few-shot}
  row additionally prefixes the block-B example block. No outcome information enters; only
  the persona wording changes.

  \paragraph{Block D: Antecedent events (zero-shot / independent / collaborative).}
  Identical to the block-A impact prompts of the given mode, with a pre-bill antecedent block
  inserted after $\langle$ctx$\rangle$: ``\texttt{$\ldots$Events that preceded and motivated
  the bill: $\langle$up to $10$ lines, each \texttt{- [date] event}$\rangle\ldots$}'' The
  antecedents (real motivating events, quote-verified, post-outcome sources removed) are themselves
  scrubbed of outcome tells, and the pair is compared with the block-A prompt (no antecedents)
  on the same bills.

  The remaining paragraphs give the prompts for the targets and models beyond Table~\ref{tab:results}:
  the multilabel action task (Tables~\ref{tab:action},~\ref{tab:actionwhy}), the weight-level
  LoRA fine-tune (main paper), impact via an independent judge (Table~\ref{tab:respimpact}),
  title-only passage, and the full-corpus relevance/impact prior (Table~\ref{tab:relimp2k}).
  All keep the same outcome-scrubbed $\langle$\textsc{ctx}$\rangle$.

  \paragraph{Actor-action, multilabel with explicit negatives (Table~\ref{tab:action}).}
  Per (bill, actor), over the $15$-action vocabulary $\langle$vocab$\rangle$: ``\texttt{$\langle$persona$\rangle$$\langle$history$\rangle$
  Bill: $\langle$ctx$\rangle$. List EVERY public action you take in response to this bill
  (zero or more). Choose only from: $\langle$vocab$\rangle$. If you take no public
  action, return an empty list. JSON: \{"actions":[...]\}}'' The empty-list option is what
  makes the \emph{verified-no-action} negatives scorable. $\langle$history$\rangle$ is empty
  in the base row and, when conditioned, is one of the three forms below.

  \paragraph{Conditioning forms for the action target (Table~\ref{tab:actionwhy}).}
  All three are temporally restricted to bills dated before the target. \emph{Action
  repertoire} (flat track record): ``\texttt{Your documented track record on earlier
  bills (most-recent first): $\langle$(year) title: action-types; $\ldots\rangle$. Across
  your record your actions break down as: $\langle$distribution$\rangle$.}'' \emph{Structural
  rationale}: each prior action with the world-state variable it targeted, its welfare direction,
  and the co-actors: ``\texttt{(\{y\}) \{title\}: you \{action\} [\{type\}] targeting \{variable\}
  (\{dir\} to you) amid \{other actors\}.}'' \emph{Stated motive} (the actor's own-voice
  reason per prior bill, written by the separate annotator grounded in the record,
  \texttt{NONE} where unsupported): ``\texttt{(\{y\}) \{title\}: \{action-types\} --
  Because $\langle$stated motive$\rangle$.}''

  \paragraph{Weight-level LoRA fine-tune (main paper).}
  The fine-tuned model uses the \emph{same} prompts as its in-context counterpart; only
  the weights differ. \emph{Relevance} (one joint call per bill over the $12$ functional roles
  $\langle$menu$\rangle$): ``\texttt{For this bill, which stakeholder roles does it MATERIALLY
  affect? Bill: $\langle$ctx$\rangle$. Choose any number from these roles: $\langle$menu$\rangle$.
  JSON: \{"roles":[...]\}}'' \emph{Impact} (per (bill, role)): ``\texttt{You ARE the "\{role\}"
  stakeholder role (\{gloss\}). Bill: $\langle$ctx$\rangle$. Does this bill BENEFIT(+)
  or HARM($-$) you? JSON: \{"dir":"+/$-$"\}}'' \emph{Action}: the action prompt above
  with stated-motive conditioning. Training targets are the gold JSON (role set / sign /
  action set); the loss is on the completion tokens only, on pre-2024 bills, with the
  $\geq$2024 test bills held out and bill-disjoint.

  \paragraph{Impact via response and an independent judge (Table~\ref{tab:respimpact}).}
  Two stages. \emph{Response} (open responder): ``\texttt{You ARE the "\{role\}"
  stakeholder role (\{gloss\}). Bill: $\langle$ctx$\rangle$. In 1--2 sentences, what concrete
  PUBLIC action(s) do you take in response to this bill? State what you DO, not how you
  feel.}'' \emph{Judge} (a \emph{separate} open model, gpt-oss-20b, not the responder and
  not shown the gold): ``\texttt{An AI-governance bill prompted a stakeholder to respond.
  Actor: "\{role\}" (\{gloss\}). Bill: $\langle$ctx$\rangle$. The actor's response: "$\langle$response$\rangle$".
  Assess the IMPACT OF THIS RESPONSE. JSON: \{"self\_dir":"+/$-$", "world\_dimension":$\langle$one
  of ten$\rangle$, "world\_dir":"+/$-$", "magnitude":"small/moderate/large"\}}'' \texttt{self\_dir}
  is scored against the gold welfare sign; the signed magnitudes aggregate into the
  world-state vector.

  \paragraph{Legislative passage, title-only (leak-free).}
  ``\texttt{Forecast whether this proposed AI-related bill will be ADOPTED / enacted
  into law, judging ONLY from its title (most such bills die in committee). JURISDICTION:
  \{juris\}. BILL TITLE: \{title\}. JSON: \{"adopted": true/false\}}'' The title-only
  form drops the constructed summary, which leaks the outcome for most US bills, so it
  is the leak-free passage prompt reported per open model in Table~\ref{tab:relimp2k}.

  \paragraph{Bill-specific retrieval prior (fine-tuned relevance / impact; Table~\ref{tab:relimp2k}).}
  For each test bill we retrieve its nearest prior bills by TF-IDF cosine (year $<$
  target) and inject only their relevant actors and welfare signs, not a global
  frequency list. \emph{Relevance prior:} ``\texttt{In the most similar earlier bills, the
  roles materially affected were: $\langle$role (count), $\ldots\rangle$.}'' \emph{Impact
  prior:} ``\texttt{On the most similar earlier bills, the "\{role\}" welfare direction was
  benefit(+) \{n\}, harm($-$) \{m\}.}''

\clearpage
\section{Data-to-Agent Validation Matrix}
\label{app:datamatrix}
The benchmark's organizing principle: \emph{every simulated actor has a real-data
source and an independent validator; every bill a real database entry with an
official status; every antecedent event a dated public document}, so an actor's
state is \emph{derived from observable data and scored against its historical
actions}, not prompted. Table~\ref{tab:datamatrix} states, per component, the real
data it draws on, what it predicts, the independent label it is scored against,
and its status: \textbf{pilot} = instantiated and validated on real data in this
paper; \textbf{target} = specified with its data source and released as the
benchmark, not yet run.

\begin{table*}[t]
\centering
\small
\setlength{\tabcolsep}{3pt}
\begin{tabular}{@{}p{0.10\linewidth}p{0.155\linewidth}p{0.215\linewidth}p{0.185\linewidth}p{0.215\linewidth}p{0.055\linewidth}@{}}
\toprule
Stage & Component & Derived from (real data) & Predicts & Validated against & Status \\
\midrule
S1 & Bill record & Congress.gov, GovInfo &, (input) & official status codes & \textbf{pilot} \\[2pt]
S2 & Actor pool & bill records, hearings, filings, reporting & recurring participants & named mentions in the record & \textbf{pilot} (recall $1.00$) \\[2pt]
S2 & Cast selection & bill $+$ actor pool & who decided \emph{this} bill & actors the record names as participating & \textbf{pilot} (R $0.93$ / P $0.34$) \\[2pt]
S3 & Antecedent events & dated pre-introduction sources &, (input) & publication dates precede introduction & \textbf{pilot} \\[2pt]
S4 & Legislator persona & sponsorship, cosponsorship, roll-call, committee, district & cosponsor / support / vote & actual roll-call action & \textbf{pilot} ($0.94$) \\[2pt]
S4 & Company persona & SEC EDGAR, LDA lobbying & support / comply / oppose & later filings \& lobbying & \textbf{pilot} (F1 $0.69$) \\[2pt]
S4 & Agency persona & rulemaking \& enforcement record & rulemake / enforce / defer & Federal Register, enforcement records & \textbf{pilot} ($0.57$) \\[2pt]
S5 & Passage (zero-shot) & bill $+$ antecedent events & pass / fail / stage & official legislative status & \textbf{pilot} (bal.\ $0.804$) \\[2pt]
S5 & Passage (multi-agent) & $+$ cast and personas & pass / fail / stage & official legislative status & \textbf{pilot} (bal.\ $0.959$, contaminated) \\[2pt]
S5 & Consequences & simulation transcript & failure mechanism & implementation record, judged & \textbf{pilot} (weaker) \\
\midrule
S2 & Committee agent & committee histories & hearing / report / delay & committee timeline \& dates & target \\
S4 & Worker / adopter & QCEW, BTOS, occupation exposure & employment, AI adoption & later QCEW / BTOS periods & target \\
S5 & Employment effect & AI adoption $+$ policy shock & jobs / wages & QCEW (event study) & target \\
S5 & Innovation effect & policy, R\&D, compute & patents / publications & USPTO PatentsView/AIPD & target \\
S5 & Price effect & input exposure $+$ shock & consumer/producer prices & BLS CPI/PPI; BEA I-O & target \\
S5 & Repair ranking & failure mechanism $+$ revisions & best targeted repair & enacted amendment outcome & target \\
\bottomrule
\end{tabular}
\caption{\textbf{Data-to-agent validation matrix,} organized by pipeline stage.
Each component's real-data source, prediction, and independent validator.
\textbf{pilot} rows are instantiated and scored on real data in this paper;
\textbf{target} rows are specified with their data source and released as the
benchmark (Congress.gov, SEC EDGAR, LDA, BTOS, QCEW, PatentsView, BLS/BEA),
each to be scored on temporally held-out data. The agents are thus falsifiable
predictive models, not role-play, the distinction the main text turns on.}
\label{tab:datamatrix}
\end{table*}

\textbf{Evidence tiers.} Not all sources carry equal weight (Table~\ref{tab:tiers}). Crucially, \emph{LLM-generated labels are never gold} (Tier~6), only candidate annotations requiring external validation, which is why GPS's validators are the historical record (votes, laws, inquiries, enforcement, government statistics), not model judgments.

\begin{table}[h]
\centering
\small
\begin{tabular}{@{}p{0.09\linewidth}p{0.52\linewidth}p{0.28\linewidth}@{}}
\toprule
Tier & Evidence type & Use \\
\midrule
1 & official vote, law, regulation, inquiry, enforcement, admin statistics & gold outcome/action label \\
2 & government survey, patent, trade, labor, incident database & quantitative state \& consequence \\
3 & independent academic dataset (documented coding) & mechanism / external validation \\
4 & company filing, lobbying report, testimony & actor state, stance, action \\
5 & reputable media report & event discovery, supplementary \\
6 & LLM-generated label & \emph{never gold}; candidate needing validation \\
\bottomrule
\end{tabular}
\caption{Evidence reliability hierarchy. GPS uses Tier-1-2 sources as validators wherever possible and treats LLM output as Tier-6.}
\label{tab:tiers}
\end{table}

\section{The Precursor System}
\label{app:legacy}
This appendix documents the system that preceded the pipeline in the main text. It assembled governance measures, incidents, and historical analogues into source-grounded causal chains, certified those chains edge-by-edge under an adversarial protocol, validated the resulting failure-mechanism labels against incident attributes and official inquiries, and paired them with a world-state adoption model backtested on hand-described decisions. Three parts follow: the chain substrate and its problem model, the validation protocols and their detailed metrics, and the pilot adoption model.

The main text depends on exactly two things from it, the failure-mechanism vocabulary used as the consequence label space (\S S5) and the inquiry-anchored validation of that vocabulary (Appendix~\ref{app:gold}). Everything else is retained for completeness and is \emph{not} re-claimed as a result of this paper. The module was previously named \textsc{Cast}; we drop the name here.

\textbf{What is superseded.} The adoption backtest below (48 hand-described decisions, described neutrally to role-agents) is the small-scale precursor of the bill-level benchmark in the main text, which replaces both its corpus and its world-state input with real bills, official status labels, and per-bill antecedent events. It is reported here as provenance, not as a current claim. The coupled hindcast and the expert-panel decision sets are independent of the new pipeline and stand as reported.

\subsection{Question 2: If It Passes, Will It Work?}
Suppose Parliament passes the retraining levy anyway. Will unemployment actually fall? Not necessarily, the levy can be undercut: firms move the work offshore, so it leaves the rule's scope; the money never reaches the workers most at risk; or the obligation carries no penalty. The failure-mechanism component identifies \emph{which} of these mechanisms is most likely to break a given measure. The mechanisms are not ad hoc: seven recurring governance-failure modes drawn from free-riding, regulatory capture, nonproliferation, the pacing problem, the security dilemma, and systemic risk \citep{ostrom1990governing,stigler1971regulation,sastry2024compute,hatz2025nuclear,collingridge1980,hendrycks2025superintelligence,perrow1984normal,hacker2025systemic}.

\subsubsection{Unit of Analysis}
The unit of analysis is a governance measure, event, or incident represented as a chain rather than as a flat label. For each row, we store extracted fields and author-coded fields separately. Extracted fields include source name, URL, date, verbatim excerpt, and explicitly named actors. Author-coded fields include actor subset, governance gap, capability or hazard movement, failed control, endpoint, pathway, closure variable, bottleneck, and candidate repair.

The chain schema is:
\[
 x \mapsto (A, g, m, c, e, p, z, b, r),
\]
where $x$ is the source excerpt, $A$ is the required actor subset set, $g$ is the governance gap, $m$ is capability or hazard movement, $c$ is the failed control, $e$ is the endpoint, $p$ is the pathway, $z$ is the closure variable, $b$ is the bottleneck, and $r$ is the repair.

\subsubsection{Corpus and Sources}
The pilot dataset contains 57 source-grounded items and 136 actor-role rows across 14 domains (Table~\ref{tab:corpus}): recent AI-specific items (governance measures, model releases, incidents, capability updates) and 30 historical analogues across cyber, financial, export-control, nuclear, public-health, aviation, and infrastructure governance, each included only when the source supports all four causal steps and the row records a limit of analogy. The benchmark splits into a \emph{certified core} (27 chains passing the strict edge test), an \emph{exploratory} set (30 with uncertain links), an \emph{adversarial} set (negative controls), and an \emph{external} test set (silver incidents and official-inquiry gold, frozen after the taxonomy). It is a deliberate \emph{pilot}: 57 audited cases fix the task and metrics; scaling to a community benchmark ($\sim$200 cases, $\sim$100 adoption decisions; Appendix~\ref{app:roadmap}) keeps the hand-audited character rather than trading it for volume.

\begin{table}[t]
\centering
\small
\begin{tabular}{p{0.34\linewidth}p{0.56\linewidth}}
\toprule
Property & Value \\
\midrule
Items / actor-role rows & 57 / 136 \\
Domains & 14 AI and analogue domains \\
Pathways & 7 source-grounded pathways, 6-11 items each \\
Historical analogues & 30 across 9 domains, each with a limit of analogy \\
Splits & random, temporal, and domain-transfer \\
Source grounding & 100\% of rows have a verbatim excerpt and URL \\
\bottomrule
\end{tabular}
\caption{Summary of the source-grounded pilot dataset. The corpus is intentionally small and audit-focused rather than a large benchmark.}
\label{tab:corpus}
\end{table}

\subsubsection{Field Provenance}
Each field is extracted, author-coded, or rule-derived, and this distinction is central: extracted fields (source name, URL, date, verbatim excerpt, explicitly named actors) make the dataset auditable; author-coded fields (actor subsets, the four chain steps, pathway, closure variable, bottleneck, repair) are hypotheses; silver labels are produced from independent incident attributes by pre-specified rules and provide external consistency checks. Field provenance is recorded per row in the released data.

\subsubsection{Grounding and Splits: Guarding Against Hallucination and Bias}
Two worries, that a dataset about AI could itself be AI-hallucinated, and that its splits could be arranged to flatter the results, are addressed and verified reproducibly. No model generates the data: every row is hand-authored from a named source with a verbatim excerpt, URL, and date (all 57 rows, 45 distinct sources, no duplicate excerpts; the build script rejects any row lacking an excerpt), and a reviewer can click any URL to check the quote. Language models enter only downstream, as blind coders and evaluators on this fixed data. Splits are deterministic, not hand-picked, a random hash of the item id, a temporal split by year, and a domain-transfer split by source domain, and recomputing the rules reproduces the stored splits exactly (0/57 mismatches), so nothing can be cherry-picked. The history-to-present matcher split shares zero source URLs and zero excerpts across train and test; the per-split pathway balance (appendix, Table~\ref{tab:balance}) is uneven, which we disclose as one reason strict single-label temporal transfer is hard.

\subsubsection{Causal-Chain Examples}
Table~\ref{tab:examples} gives expert-facing examples. Experts rate each chain step rather than only the final pathway label.

\begin{table*}[t]
\centering
\small
\begin{tabular}{p{0.14\linewidth}p{0.18\linewidth}p{0.48\linewidth}p{0.13\linewidth}}
\toprule
Case & Source signal & Causal chain & Pathway / variable \\
\midrule
Open-weight release & A model provider makes weights available for broad use & release control does not bind redistribution $\rightarrow$ weights diffuse to downstream actors $\rightarrow$ takedown cannot reach local copies $\rightarrow$ non-revocable dual-use capability & open-weight proliferation / containment \\
Ebola analogue & delayed emergency declaration & early signals not escalated $\rightarrow$ outbreak outruns response $\rightarrow$ emergency authority delayed $\rightarrow$ early control fails & governance lag / timeliness \\
Compute reporting & domestic compute reporting rule & rule covers domestic providers $\rightarrow$ training can move offshore $\rightarrow$ domestic reporting cannot observe uncovered compute $\rightarrow$ unsafe scaling outside regime & foreign free-riding / coverage \\
Critical-infrastructure dependency & many deployers rely on shared AI component & single provider/component becomes common dependency $\rightarrow$ failure propagates across users $\rightarrow$ fallback/diversity control absent $\rightarrow$ correlated service failure & systemic dependency / resilience \\
\bottomrule
\end{tabular}
\caption{Examples of the cause-effect chain format. Each row is designed for audit: the expert can agree with the pathway while rejecting a specific step.}
\label{tab:examples}
\end{table*}

\subsection{Problem Model: Actors, Pathways, and Closure}
\subsubsection{Actor Subsets}
A governance measure is complete only if the actor subsets required for implementation, verification, and enforcement are covered. We use ten role-level actor subsets (frontier developers, compute providers, regulators, evaluators, model hubs, open-weight actors, foreign states, deployers, and critical-infrastructure operators/affected publics), mapped by the governance lever an actor controls, not by identity (an actor can belong to several).

\subsubsection{Failure Pathways}
Start with the retraining levy. Firms subject to it can shift the work to contractors abroad, so the levied activity slips outside the rule's reach, a \emph{coverage} failure (\emph{foreign free-riding} when the decisive actor is offshore); and if the obligation carries no real penalty, the levy is an \emph{enforcement gap}. These are two of \textbf{seven} recurring governance-failure mechanisms (foreign free-riding, enforcement gap, capability leakage, open-weight proliferation, governance lag, rivalry acceleration, systemic dependency), each a chain of real events escalating to a catastrophic endpoint if left unclosed (Table~\ref{tab:realevents}). A side example shows the range: a frontier training license binds domestic labs, but an insider can move model weights abroad, unrecallable once gone, \emph{capability leakage}, the containment failure from the A.Q.\ Khan network (2004) to a 2024 AI trade-secret theft. The taxonomy is testable: membership is decided by an explicit five-part rule (appendix), not by resemblance to a label.

\subsubsection{Closure Model}
For a measure $m$ and pathway $p$, the component computes a closure score
\[
C(m,p)=r(m,p)a(m,p)v(m,p)e(m,p)t(m,p)h(m,p),
\]
where $r$ is relevance, $a$ is actor coverage, $v$ is verification, $e$ is enforcement, $t$ is timeliness, and $h$ is containment or revocability when distribution matters. The pathway remains open when a necessary factor is near zero. The diagnosed bottleneck is the factor whose repair produces the largest reduction in diagnostic failure score.

The multiplicative form is a conservative bottleneck model, not a probability model; we read only relative diagnosis, and the diagnosed bottleneck is invariant across min-link and additive-with-threshold aggregators on the seven scenario factor vectors. \emph{\textbf{Policy takeaway:}} most implementation failures trace to a few recurring mechanisms, so a government can screen a proposal against seven known failure modes before harm, not after.

\subsection{Question 3: Where Will It Fail, and How to Fix It?}
Applied to the retraining levy, the workflow converts its labels into closure factors and performs a \emph{counterfactual sweep}: holding the scenario fixed, it strengthens one closure link at a time and measures the change in the diagnostic failure score (read only for relative diagnosis, not as a catastrophe probability). For the levy the links whose repair helps most are \emph{coverage} and \emph{enforcement}, so the fix is to redesign the instrument, a funded transition fund with mandatory displacement reporting and tax incentives, which firms cannot offshore around and which routes support to the workers at risk, exactly what the brief recommends.

\subsection{How We Evaluate GPS}
``Is GPS good?'' has one structured answer: GPS is a pipeline whose every stage is a concrete decision question with its own test and data (Table~\ref{tab:evalframe}), each mapping to one of \textbf{four validity dimensions}, predictive, diagnostic, decision, and simulation. More broadly, \emph{every agent and transition has a real-data source and an independent validator}, the historical record, not LLM judgment, as the label (Appendix~\ref{app:datamatrix}); this pilot validates the decision-critical rows and releases the rest as the benchmark. Throughout we report Wilson intervals and treat small-$n$ checks as plausibility signals.

\textbf{Perceive, model, decide, a pipeline, not a single answer.} A recommendation is only as sound as the world GPS starts from, so we evaluate every stage, from \emph{perceiving} the actors, through \emph{modeling} the stakeholders and causal chain, to the \emph{decision}, both in isolation and end-to-end, with all metrics collected in Table~\ref{tab:evalframe}. Because stages depend on each other, we do \emph{not} multiply their accuracies (errors correlate; later stages recover); instead a sensitivity analysis (stable in $98.5\%$ of $\pm0.10$ trials) flags the $2/15$ flip-prone decisions for human review. GPS also attaches a \emph{confidence} to each answer from its validation strength, adoption and hindcast \emph{high}, pathway \emph{medium}, repair \emph{advisory}, so an upstream error is surfaced, not silently propagated (Appendix~\ref{app:sensitivity}).

\begin{table}[t]
\centering
\small
\setlength{\tabcolsep}{3pt}
\begin{tabular}{@{}p{0.29\linewidth}p{0.37\linewidth}p{0.26\linewidth}@{}}
\toprule
Stage / layer (validity) & How it is validated (data) & Result \\
\midrule
Perception, world understanding & actor-subset identification, 57 cases & F1 $0.73$ (recall $0.94$) \\[2pt]
Stakeholder modeling & coverage vs.\ author-coded key, 4 policies & $0.63$ \\[2pt]
Adoption, \emph{predictive} & 48 real 2023-2025 decisions, neutral & \textbf{$0.77$} (Brier $0.11$) \\[2pt]
Failure mechanism / implementation, \emph{diagnostic} & pathway \& bottleneck: SECC blind cross-model, silver, gold anchor & pathway $0.77$-$0.90$; gold $8/11$; bottleneck contested \\[2pt]
Repair & objective improvement; expert review (future) & advisory \\[2pt]
Decision, \emph{decision} & expert-panel plurality, 15 sets & \textbf{$14/15$} \\[2pt]
Simulation, \emph{simulation} & coupled hindcast, 17 real decisions & \textbf{$0.88$} ($15/17$) \\[2pt]
Robustness & objective sensitivity; adoption paraphrase & $98.5\%$ stable; $0.78\!\pm\!0.035$ \\
\bottomrule
\end{tabular}
\caption{\textbf{All GPS validation metrics in one place}, by stage and validity layer (the precursor pipeline's input$\to$output flow). The failure \emph{type} reproduces robustly while the fine-grained repair link is contested (mechanism row); repair quality and the long-run \emph{policy-ranking} target await the register (Appendix~\ref{app:roadmap}). Wilson intervals are in the text; small-$n$ rows are plausibility signals, not powered estimates.}
\label{tab:evalframe}
\end{table}

\subsubsection{Strict Edge-Backed Chain Certification (SECC), Diagnostic}
The component does not assume a full policy-failure chain can be validated automatically end-to-end; it decomposes the diagnosis into five edges and certifies which survive skeptical scrutiny. We keep a deliberately \emph{strict} criterion: a chain is certified only if \emph{all five} edges, policy gap $\rightarrow$ movement, movement $\rightarrow$ failed control, control $\rightarrow$ endpoint, endpoint $\rightarrow$ pathway, pathway $\rightarrow$ bottleneck, withstand an adversarial judge that tries to refute each and defaults to ``refuted'' when evidence is insufficient. A first pass judging bare chain steps certifies \textbf{0/57}: the early edges are robust but the interpretive tail is weak (Table~\ref{tab:validity}), the final edges were left implicit.

We therefore introduce \emph{Strict Edge-Backed Chain Certification} (SECC), which makes each edge's evidence explicit (a verbatim excerpt, the codebook warrant, a stated alternative, and for the bottleneck a targeted-versus-wrong repair contrast) before the same all-five-edges test. Certification rises to \textbf{27/57} ($0.47\,[0.35,0.60]$; Table~\ref{tab:validity}), but this measures whether each edge is \emph{defensible once its warrant is explicit}, not blind survival. We therefore also run the harder test: a \emph{blind, cross-model} re-certification. A \emph{different} model (\textsc{claude-opus-4-8}; chains coded with \textsc{claude-sonnet-4-6}), given only the chain facts and the neutral label space, no warrant, no preferred alternative, independently assigns the pathway and bottleneck, agreeing with the coded labels on endpoint$\rightarrow$pathway $0.81$ and pathway$\rightarrow$bottleneck $0.77$ (all five edges $36/57$, $0.63\,[0.50,0.75]$; Table~\ref{tab:validity}). This is inter-model \emph{agreement given the coded chain}, not raw-fact recovery, the harder neutral-facts test is the gold anchor below. We do not force the rest to pass: the remaining 30 chains are released as \emph{exploratory} (Table~\ref{tab:secc}).

\begin{table}[t]
\centering
\small
\begin{tabular}{p{0.30\linewidth}p{0.44\linewidth}c}
\toprule
Set & Criterion & $n$ \\
\midrule
Full corpus & source excerpt + chain & 57 \\
SECC-certified & all five edges survive strict refutation & 27 \\
Exploratory & $\ge$1 edge fails after SECC & 30 \\
Priority target & failing edge is pathway/bottleneck & 23 \\
\bottomrule
\end{tabular}
\caption{Chain certification under SECC. The strict all-five-edges criterion is unchanged from the first pass (which certified 0/57); making each edge explicit lifts certification to 27/57. Exploratory chains are retained for the silver and agentic checks, not discarded or forced to pass.}
\label{tab:secc}
\end{table}

\subsubsection{Diagnostic Layers, Results}
The \emph{diagnostic} dimension is checked without gold labels along the data, problem-model, and output criteria by which an applied AI-for-social-impact contribution is ordinarily judged (Table~\ref{tab:results}): the problem model is certified edge-by-edge (SECC), the output is aligned to independently coded incident attributes (silver) and to what emerges when role-agents play each scenario out (agentic), and both are backed by transfer and negative-control stress tests. A paired test separates the two labels where samples are small.

\emph{Internal sanity checks} (the counterfactual sweep recovering the coded bottleneck, repair discrimination, the new-pathway trigger, grounding, and closure-aggregator invariance) all pass as expected unit tests (Table~\ref{tab:results}, appendix).

\subsubsection{Transfer, Baselines, and Negative Controls, Diagnostic}
A central claim is that \emph{new AI governance failures often resemble older governance failures} in cyber, public health, export controls, finance, and infrastructure. Most recent events' nearest precedent is a pre-2019 case, a 2025 open-weight release matches historical proliferation, a compute-reporting gap matches export-control free-riding, and a mechanism-aware model matches the right pathway family across the history-to-present split at $\approx0.8$ top-2 (Table~\ref{tab:match}), though surface retrieval does not (next subsection). Recent AI failures are largely \emph{new technologies with old governance mechanisms}. A blind zero-shot labeler over all 57 chains reaches pathway accuracy $0.56$ ($\kappa$ 0.32), confirming the assignment is non-trivial, and supplying non-AI historical exemplars lifts held-out pathway accuracy from $0.75$ to $0.83$, direct evidence the pathways recur rather than being AI-specific stories. Because the near-misses are confusable \emph{pairs}, a \emph{contrastive few-shot} mapper (one exemplar per candidate pathway with its discriminator) raises leave-one-out top-1 from $0.49$ ($k$NN) to $0.61$ (top-2 $0.83$).

Three deterministic stress tests reposition surface methods as baselines (Table~\ref{tab:labelfree}). A keyword classifier matches embedding retrieval at pathway naming ($0.54$ vs.\ $0.51$ top-1, both far above the $0.19$ majority) yet passes only $3/7$ adversarial negative controls and transfers weakly across domains ($0.37$): pathway naming is partly lexical, but surface features cannot read the invariant, so the membership rule and reasoning-based certification carry the diagnosis. Abstaining on low-margin cases lifts accuracy on the confident $40\%$ to $0.65$.

The dataset also supports first-draft memo generation (Table~\ref{tab:memo} format): a model proposes actors, pathway, and a candidate bottleneck as decision support, not ground truth (actor-role $\mathrm{F1}\,0.61$, pathway top-1 $0.67$ on twelve held-out events).

\subsubsection{Silver-Standard External Alignment, Diagnostic}
Because no dataset gives gold governance-pathway labels, we build a silver benchmark from 26 AI Incident Database reports: we extract \emph{independent} incident attributes (misuse vs.\ failure, cross-jurisdiction actor, unauthorized transfer, over-reliance, mitigation type), derive a silver pathway and bottleneck by pre-specified rules, and apply the codebook blind to them. The pathway label receives silver support, the codebook matches on 20/26 (top-2 $0.77\,[0.58,0.89]$; top-1 $0.46$), but the fine-grained bottleneck does not ($\approx0.19\,[0.09,0.38]$), because observed responses are reactive whereas the codebook asks which upstream link would have closed the pathway \emph{before} harm. A \emph{paired} test settles the asymmetry: on 7/26 incidents the codebook gets the pathway right while missing the bottleneck and on none the reverse (McNemar $p=0.016$); against the $0.385$ majority baseline, pathway reaches baseline while the bottleneck falls below it, the contested label the agentic and gold checks target.

\textbf{External grounding.} The pathways are orthogonal to harm-type catalogs \citep{mcgregor2021aiid,slattery2024riskrepo}: re-coding 26 incidents with an ``other'' option maps only 7/26 (the rest are product harms), so the taxonomy does not force-fit, and endpoints are calibrated against expert elicitation \citep{grace2024thousands} (Appendix~\ref{app:external}).

\textbf{Gold anchor.} For 11 governance failures with an authoritative official inquiry, we take the inquiry's named failure as gold. A blind LLM matches it on 8/11 ($0.73\,[0.43,0.90]$); our codebook on 4/8 of the corpus cases ($0.50\,[0.22,0.78]$), small $n$, so read as an anchor, not an estimate. Historical failures split on \emph{enforcement} (under-weighted by our coding, so the bottleneck stays contested even against gold); three algorithmic scandals (Dutch childcare algorithm, Robodebt, Post Office Horizon) are all \emph{verification} failures the model recovers (Appendix~\ref{app:gold}).

\subsubsection{Agentic-Simulation Validation, Diagnostic}
The silver and panel layers check the bottleneck label against \emph{static} attributes and \emph{re-coders}. The strongest independent test asks a different question: when the actors actually \emph{play out} the scenario, does the same failure link emerge on its own? We run a short LLM multi-agent simulation for each of seven governance scenarios. Role-agents, a frontier developer (deploy fast while appearing compliant), a regulator (prevent harm using only the powers the measure actually grants), and a scenario-specific third actor (a foreign compute provider, a subcontractor, a rival state, $\ldots$), each take a concrete action over two rounds, given \emph{only} the situation and their incentive, never the coded bottleneck. A separate extractor then reads the transcript and names the single closure link whose failure drove the bad outcome. Agreement with the independently coded bottleneck is a behavioral, label-free check that the diagnosed link is what the incentives actually produce, not merely our annotation.

The emergent failure link matches the coded bottleneck on \textbf{4/7} scenarios (timeliness, containment, verification, and resilience recovered exactly). To rule out a same-model artifact, the agents and the extractor were the same model, we re-run with a \emph{cross-model} split (agents \textsc{claude-sonnet-4-6}, extractor \textsc{claude-opus-4-8}); agreement rises to \textbf{5/7}, so the emergent link is not an artifact of one model judging itself. With only 7 scenarios both intervals are wide and span chance ($4/7=0.57\,[0.25,0.84]$; $5/7=0.71\,[0.36,0.92]$): this is a behavioral \emph{plausibility} check, not a powered estimate. The misses are diagnostic, not random: the agents converge on the \emph{earliest} link an actor can exploit while the codebook records the downstream link that ultimately breaks, localizing exactly where the fine-grained bottleneck label is contestable, the same link the silver and panel layers flag.

\subsubsection{Question 1: Will the Proposal Become Law? (Adoption)}
The government proposes the mandatory retraining levy. Will it pass? GPS predicts only \textbf{11\%}, firms call the cost too high; workers back it but the government fears slower growth. The reason is that adoption turns not on technical merit but on the \emph{shared interest} of politicians and the public: politicians optimize re-election, the public optimizes concrete concerns, polls put worry on deepfakes (66-87\%), jobs (56\%), and bias (55\%) \citep{pew2025ai,aipi2024}, so a measure passes when it addresses a concern the public holds at low industry cost. Stakeholder role-agents (including a public agent) model this; on a hard backtest of 48 real 2023-2025 decisions (Table~\ref{tab:backtest}) described \emph{neutrally} the model recovers outcomes at $0.77$ (Brier $0.11$; $0.95$ clear, $0.66$ hard), with 8/8 leakage-free synthetic and 5/5 counterfactual flips confirming mechanism over recall, and accuracy stable ($0.78\!\pm\!0.035$; $4/48$ flips) across neutral paraphrases by a different model (Appendix~\ref{app:sensitivity}). \textbf{Composite.} Multiplying adoption by implementation effectiveness locates the achievable-\emph{and}-effective zone, deepfake and scam bans are in it, voluntary commitments adopted-but-toothless, a frontier license effective-but-unadoptable (Appendix~\ref{app:feasibility}). \emph{\textbf{Policy takeaway:}} public support and industry cost jointly decide whether an effective AI rule becomes law, so rules aimed at a felt harm (scams, deepfakes) are both more passable and more impactful.

\subsubsection{Does GPS Reproduce Governance History? (Hindcast)}
The simulator is a dynamical system,
\[ W_{t+1}=f(W_t,\,A_t,\,P_t), \]
where $W_t$ is the world state, $A_t$ the agent actions (company, citizen, and government moves), and $P_t$ the policy; each stage computes one component of $f$. Stages 1-2 are one \emph{transition}: a world state $W_t$ (ten coded variables with data sources, AI capability, the economy, public opinion, industry, politics, international competition) and a measure $P_t$ map to adoption, implementation, and consequences, updating $W_{t+1}$. The test is whether \emph{coupling} the stages, driving adoption from the evolving state, not a per-case constant, preserves accuracy. Hindcasting 2022-2025 (concern rises $0.41\!\to\!0.80$, tracking polls from 57\% to 72\%), the coupled model recovers 15/17 real decisions ($0.88$), with $n{=}17$ this shows coupling \emph{preserves} accuracy (its CI overlaps the 48-case backtest), not that it improves it, and reproduces the 2024 consumer-law surge, once institutional route (executive, authoritarian, regulatory-union) is modeled. Forward 2026-2030 were \emph{conditional} stress-tests: under a cyber incident, labor displacement, or an international race, consumer measures stay feasible while frontier-safety controls stay hard. \emph{\textbf{Policy takeaway:}} GPS reproduces recent governance history closely enough for scenario analysis, stress-testing which rules stay feasible under a shock, not prediction.

\subsubsection{Does GPS Recommend Sensible Policies?}
GPS ultimately claims to help \emph{choose} measures, so we test that end-to-end. On 15 decision sets (a goal and three candidate measures each, across compute, elections, labor, model release, hiring, medical AI, and more), GPS ranks candidates by its objective while five \emph{single-lens} experts (economist, legal, security, safety, political) each pick from their remit alone. The lenses conflict and split on 10 of 15 sets. GPS's balanced choice matches the divided panel's plurality on \textbf{14/15} ($0.93\,[0.70,0.99]$; e.g.\ compute-provider KYC over both a hard license and toothless commitments); the one miss is medical AI, where GPS down-weights a costly pre-market approval the experts still favor. We do \emph{not} claim GPS matches \emph{human} experts: both sides are LLMs, so the honest reading is narrow, a deterministic aggregator internally consistent with the plurality of LLM single-lens agents, robust to the panel model and coding. Whether it tracks \emph{human} judgment needs a human study we have not run, the result's central limitation. \emph{\textbf{Policy takeaway:}} GPS steers scarce political effort toward rules that can both pass and cut real harm.

\subsection{Detailed Evaluation Metrics}

\begin{table}[t]
\centering
\small
\begin{tabular}{p{0.31\linewidth}p{0.33\linewidth}p{0.24\linewidth}}
\toprule
Test & Setup & Result \\
\midrule
\multicolumn{3}{l}{\emph{Internal consistency (mechanics)}}\\
Bottleneck diagnosis & sim vs.\ coded link, 17 cases & 0.94 / $\kappa$ 0.93 \\
Repair discrimination & correct vs.\ best wrong, 11 cases & win 1.00 / margin 0.24 \\
New-pathway flag & 5 known / 5 novel & 0/5 FP, 5/5 TP \\
Source grounding & field audit, 10 events & 1.00 \\
Closure-form invariance & 3 aggregators, 7 scenarios & 7/7 identical \\
\midrule
\multicolumn{3}{l}{\emph{Automated recoding \& transfer}}\\
Zero-shot chain labeling & blind LLM, 57 items & pathway 0.56 / $\kappa$ 0.32 \\
Actor-behavior-pathway synthesis & LLM applies rules, 12 events & actor-role F1 0.61 / pathway 0.67 \\
Pathway matching (history $\to$ 23-26) & retrieval, temporal & top-2 0.77 / 0.86 \\
Contrastive few-shot mapping & 1 exemplar/pathway, LOO & top-1 0.61 / top-2 0.83 \\
\midrule
\multicolumn{3}{l}{\emph{External alignment (silver standard)}}\\
Pathway (AIID) & silver rule vs.\ blind codebook, 26 & top-2 0.77 / top-1 0.46 \\
Preventive bottleneck & vs.\ cause-implied silver, 26 & 0.19 \\
Response bottleneck & vs.\ mitigation-implied silver, 21 & 0.19 \\
\midrule
\multicolumn{3}{l}{\emph{Agentic simulation}}\\
Emergent vs.\ coded bottleneck & LLM role-agents, 7 scenarios & 4/7 = 0.57 \\
\bottomrule
\end{tabular}
\caption{Evaluation results. Internal-consistency values are deterministic checks that the machinery behaves as intended (near-perfect is an expected unit-test outcome, not generalization); the remaining blocks are external and behavioral signals, automated recoding and history-to-present transfer, silver-standard alignment to independently coded incident attributes, and an agentic simulation that plays each scenario out and compares the emergent failure link to the coded bottleneck. Pathway labels reproduce substantially while the fine-grained bottleneck label is the harder target; 95\% Wilson CIs are given in the text, and a paired McNemar test on the silver incidents separates the two labels at matched top-1 ($p=0.016$).}
\label{tab:results}
\end{table}

\begin{table}[h]
\centering
\small
\begin{tabular}{p{0.56\linewidth}c}
\toprule
Label-free stress test & value \\
\midrule
\multicolumn{2}{l}{\emph{Baselines (pathway top-1)}}\\
chance (1/7) & 0.14 \\
majority class & 0.19 \\
keyword/rule classifier & 0.54 \\
embedding retrieval (kNN) & 0.51 (top-2 0.77) \\
\midrule
\multicolumn{2}{l}{\emph{Calibration (accuracy at coverage)}}\\
coverage 1.00 / 0.60 / 0.40 & 0.51 / 0.53 / 0.65 \\
\midrule
\multicolumn{2}{l}{\emph{Leave-one-domain-out (surface retrieval)}}\\
historical analogues $\rightarrow$ AI cases & top-1 0.19 / top-2 0.37 \\
(mechanism-aware model, same transfer) & top-2 0.86 \\
\midrule
\multicolumn{2}{l}{\emph{Negative controls (should-not-fit)}}\\
surface predictor avoids the invalid pathway & 3/7 \\
\bottomrule
\end{tabular}
\caption{Deterministic label-free stress tests. Surface methods (keyword, embedding retrieval) name pathways moderately but fail the invariant checks, negative controls (3/7) and strict domain transfer (top-2 0.37), because they cannot read the mechanism; a mechanism-aware model recovers the same transfer at 0.86. This positions surface methods as baselines and motivates the mechanism-level membership rule and reasoning-based certification. Abstention on low-confidence cases lifts accuracy on the confident core.}
\label{tab:labelfree}
\end{table}

\begin{table}[h]
\centering
\small
\begin{tabular}{p{0.42\linewidth}cc}
\toprule
Matcher (setting) & metric & value [95\% CI] \\
\midrule
Embedding kNN, leave-one-out & top-1 & 0.53 [0.40, 0.65] \\
Embedding kNN, leave-one-out & top-2 & 0.77 [0.65, 0.86] \\
Embedding kNN, temporal $\le$2022$\to$23-26 & top-2 & 0.48 [0.31, 0.66] \\
RAG + LLM, temporal (pred in gold) & acc & 0.69 [0.51, 0.83] \\
RAG + LLM, temporal (either/either) & acc & 0.86 [0.69, 0.95] \\
RAG + LLM, temporal & top-1 / macro-F1 & 0.45 / 0.31 \\
Recent event nearest precedent is $\le$2019 & rate & 0.72 [0.54, 0.85] \\
\bottomrule
\end{tabular}
\caption{Pathway matching across time. Deployed-tool retrieval is deterministic; the retrieval-augmented LLM uses claude-sonnet-4-6 and is non-deterministic (model version reported). Matching a recent event to the correct pathway family is reliable ($\approx0.8$ at top-2 / either); exact single-label and per-class balance remain hard.}
\label{tab:match}
\end{table}

\begin{table}[h]
\centering
\small
\begin{tabular}{p{0.34\linewidth}ccc}
\toprule
Pathway & train & dev & test \\
 & ($\le$2019) & (20-23) & ($\ge$24) \\
\midrule
foreign free-riding & 2 & 2 & 5 \\
enforcement gap & 1 & 3 & 5 \\
capability leakage & 5 & 1 & 2 \\
open-weight proliferation & 1 & 5 & 0 \\
governance lag & 3 & 3 & 5 \\
rivalry acceleration & 6 & 1 & 0 \\
systemic dependency & 4 & 2 & 1 \\
\midrule
total & 22 & 17 & 18 \\
\bottomrule
\end{tabular}
\caption{Per-split pathway balance for the temporal split. The distribution is uneven, some pathways (e.g.\ open-weight proliferation, rivalry acceleration) are rare or absent in a given split, which is one honest reason strict single-label temporal transfer is hard. Splits are recomputed from deterministic rules and match the stored splits exactly (0/57 mismatches); the temporal history-to-present matcher split shares zero source URLs and zero excerpts across train/test.}
\label{tab:balance}
\end{table}

\begin{table}[h]
\centering
\small
\begin{tabular}{p{0.34\linewidth}ccc}
\toprule
Chain edge & bare & +SECC & blind$\times$model \\
\midrule
gap $\rightarrow$ movement & 0.91 & 0.83 & 0.90 \\
movement $\rightarrow$ control & 0.79 & 0.91 & 0.88 \\
control $\rightarrow$ endpoint & 0.63 & 1.00 & 0.86 \\
endpoint $\rightarrow$ pathway & 0.39 & 0.77 & 0.81 \\
pathway $\rightarrow$ bottleneck & 0.16 & 0.65 & 0.77 \\
\midrule
chains with all 5 edges intact & 0/57 & 27/57 & 36/57 \\
\bottomrule
\end{tabular}
\caption{Per-edge survival across three protocols. \emph{bare}: same-model judge sees only the chain steps and tries to refute each edge (default ``refuted''), an adversarial floor. \emph{+SECC}: the judge is given the codebook warrant and a stated alternative, measuring whether each edge is defensible \emph{once its warrant is explicit} (not independent evidence). \emph{blind$\times$model}: the follow-up we promised, now run, a \emph{different} model (\textsc{claude-opus-4-8}; chains coded with \textsc{claude-sonnet-4-6}) is given only the chain facts and the neutral label space, no warrant and no preferred alternative, and independently assigns the pathway and bottleneck; the number is inter-model \emph{agreement} with the coded label. A different model reads the same mechanism the same way on endpoint$\rightarrow$pathway $0.81$ and pathway$\rightarrow$bottleneck $0.77$ (all five edges $36/57$). This is agreement given the coded chain, not recovery from raw facts (the gold anchor is that harder test).}
\label{tab:validity}
\end{table}

\subsubsection{Decision-Objective Sensitivity}
\label{app:sensitivity}
The decision objective multiplies three hand-coded quantities, so we test whether its ranking is an artifact of the coding or the aggregation form. Adding uniform noise to every coded input (salience, implementation, risk-reduction) and re-ranking over 400 trials per set, GPS's top-1 choice is unchanged in $99.9\%$ of trials at $\pm0.05$, $98.5\%$ at $\pm0.10$, $94.4\%$ at $\pm0.15$, and $89.1\%$ at $\pm0.20$. Swapping the multiplicative form for a min-link form reproduces all $15/15$ winners and an additive form $12/15$, so the ranking is largely form-independent. Only two sets are genuinely fragile (top-1 flips in $\geq\!10\%$ of $\pm0.10$ trials): election-deepfake governance and AI incident response, where two candidates score nearly equally, we flag these rather than present a spurious winner. This does not make the coded inputs objective; it shows the \emph{ranking} does not hinge on their exact values, and the objective's weights should be set by the deploying office, not us.

\textbf{Adoption prompt-sensitivity.} A benchmark others run must not swing on phrasing, so we re-run the 48-case adoption backtest across three \emph{neutral} phrasings of each case: the original plus two paraphrases generated by a \emph{different} model (\textsc{claude-opus-4-8}) that preserve every fact and add no outcome cue. Accuracy is $0.78\pm0.035$ (by phrasing $0.81/0.79/0.73$), and only $4/48$ cases change their binary prediction across phrasings, so the score is stable to neutral rewording. (A separate, non-neutral ``cued'' phrasing does score higher; we report the neutral number precisely because cueing is what a usable benchmark must avoid.)

\textbf{World understanding (perception).} Before GPS reasons it must \emph{see} the world. On the 57 corpus cases, a blind model given only the neutral summary identifies the involved role-level actor subsets (A1-A10) at recall $0.94$, precision $0.59$, F1 $0.73$ ($117$ hits, $8$ misses, $80$ over-inclusions): it almost never misses an involved actor but tends to over-include plausible ones, the safe error direction for a tool whose job is to not overlook an affected party.

\textbf{Stakeholder coverage.} GPS's consequence map is only useful if it names the \emph{right} affected parties. For four policies (GPU export, worker retraining, frontier license, deepfake liability) we author-code the expected stakeholders in three tiers (primary/secondary/tertiary) and ask a blind model to list a policy's affected parties; it recovers the key at $0.63$ ($19/30$), and the errors are structured: primary, most-affected actors are found reliably while secondary and tertiary ripples are missed more often. The key is author-coded (an anchor, not gold), so this bounds coverage rather than certifying it; scaling to a graded, independently-built stakeholder key is future work.

\subsection{Pilot Adoption and Feasibility Model}
\label{app:feasibility}
This simulation reframes the agentic layer from \emph{technical desirability} to \emph{political feasibility}. Each stakeholder-agent optimizes an incentive-weighted utility $U=\alpha\,\mbox{Re-elect}+\beta\,\mbox{Party}+\gamma\,\mbox{Econ}+\delta\,\mbox{NatSec}+\epsilon\,\mbox{Safety}$ in which AI safety ($\epsilon$) is usually not the dominant term (Table~\ref{tab:stakeholders}). We do not ask ``is this policy good?'' but ``given these incentives, will it be adopted?''. Stakeholders report a stance in $[-2,+2]$ and the salience they assign AI in the current world state; a government-agent, weighting stakeholders by electoral power, returns an adoption probability. Sweeping world-state regimes, including a post-incident shock anchored to a real corpus event (the 2026 grid-attack), exposes the focusing-event dynamics behind most real policy change (Table~\ref{tab:feasibility}).

\begin{table}[h]
\centering
\small
\begin{tabular}{p{0.30\linewidth}p{0.60\linewidth}}
\toprule
Stakeholder & Primary incentive (AI safety is minor) \\
\midrule
Politician & re-election, party success, then economy / national security \\
Voter & jobs, cost of living, healthcare, personal safety \\
AI company & profit, market share, speed to deploy \\
Nat.-security agency & strategic advantage; avoid capability surprise \\
Research lab & scientific progress; open access \\
\bottomrule
\end{tabular}
\caption{Stakeholder incentives used by the feasibility agents. AI safety enters each utility only through its effect on the stakeholder's primary concerns.}
\label{tab:stakeholders}
\end{table}

\begin{table}[h]
\centering
\small
\begin{tabular}{p{0.40\linewidth}cc}
\toprule
World-state regime & License & Reporting \\
\midrule
Baseline (AI a niche issue) & 0.08 & 0.18 \\
Post-incident shock (grid attack) & 0.82 & 0.82 \\
High unemployment (AI-blamed) & 0.72 & 0.72 \\
GDP declining (AI a growth engine) & 0.08 & 0.72 \\
Pre-election (electoral calculus) & 0.18 & 0.62 \\
\bottomrule
\end{tabular}
\caption{Adoption probability by world-state regime for a heavy measure (a 2028 frontier-training license) and a light one (compute-run reporting). Both spike after a focusing event and under AI-blamed unemployment; they diverge when GDP contracts, the burdensome license conflicts with the growth priority while the light rule does not. Incentive-conditioned scenario estimates from role-agents (\texttt{claude-sonnet-4-6}), not empirical predictions; the ordering and the regime shift are the signal, not the absolute values.}
\label{tab:feasibility}
\end{table}

\begin{table}[h]
\centering
\small
\begin{tabular}{p{0.62\linewidth}cc}
\toprule
Real decision (described \emph{neutrally} to agents) & pred & out. \\
\midrule
Frontier safety-test + liability bill, AI-hub state & 0.35 & fail \\
Binding GPAI duties amid a voluntary-code push & 0.62 & adopt \\
Executive AI directive under existing authority & 0.82 & adopt \\
Comprehensive federal AI law, split chambers & 0.12 & fail \\
Binding duties vs.\ offered voluntary commitments & 0.08 & fail \\
Statutory AI rules vs.\ regulators (tech economy) & 0.55 & fail \\
First-mover AI anti-discrimination law (non-AI state) & 0.35 & adopt$^\ast$ \\
Ban on real-time public facial recognition & 0.41 & partial \\
Disclosure of AI content in election ads & 0.72 & adopt \\
Ban AI voices in unsolicited robocalls & 0.82 & adopt \\
Full AI duties applied to open-source developers & 0.08 & partial$^\ast$ \\
Pre-launch review of generative-AI services & 0.41 & partial \\
Post-election signal of binding frontier rules & 0.55 & pending \\
Broad private liability for AI harms & 0.18 & fail \\
AI disclosure via a consumer-protection law & 0.82 & adopt \\
Prohibit AI for enumerated harmful purposes & 0.72 & adopt \\
Protect voice / likeness from AI cloning & 0.75 & adopt \\
Disclose generative-AI training-data summaries & 0.35 & adopt$^\ast$ \\
Criminalize non-consensual intimate deepfakes & 0.85 & adopt \\
Framework duties for high-impact / generative AI & 0.72 & adopt \\
Light AI-promotion law, no penalties & 0.07 & partial$^\ast$ \\
High-risk AI duties bill (light-touch governor) & 0.23 & fail \\
Ban algorithmic rent-price coordination & 0.62 & fail$^\ast$ \\
Ten-year moratorium on state AI laws & 0.27 & fail \\
Comprehensive AI law (government falls mid-session) & 0.28 & fail \\
Risk-based AI framework (one chamber approved) & 0.72 & pending \\
Comprehensive risk-tiered AI law (union) & 0.82 & adopt \\
Consent for AI replicas of performers & 0.74 & adopt \\
\bottomrule
\end{tabular}
\caption{Hard backtest against 48 real 2023-2025 AI-governance decisions described \emph{neutrally}, facts only, no outcome-telegraphing, spanning adoptions, vetoes, a procedural death, weakened, and pending cases across US states, national governments, and unions (representative cases shown). \emph{out.}: adopt / fail / partial or pending (graded 0.5, scored correct when $0.25\le\mathrm{pred}\le0.75$). Accuracy $0.77$ (37/48): $0.95$ on clear-incentive cases but $0.66$ on hard ones; Brier $0.11$; probabilities sit near $0.5$ on contested cases. The misses (11/48, some marked $^\ast$) concentrate on surprising first-mover adoptions and non-binding international declarations that expert observers also found hard to call. LLM predictions vary slightly by run; a version with cued descriptions scores higher but is not the honest number.}
\label{tab:backtest}
\end{table}

\textbf{Shared-interest backtest.} A second backtest tests the reframed claim directly: measures are adopted when they lie in the \emph{shared-interest zone}, high public salience at low industry cost, not when experts rank them important. Public salience is grounded in polling, not assigned per case: deepfakes/misinformation $\approx0.85$, scams $\approx0.80$, job loss $0.56$, bias $0.55$ \citep{pew2025ai,aipi2024}. Coding eleven real measures on (public salience, industry cost), every shared-interest measure was adopted (3/3) and expert-favored high-cost, low-salience measures mostly failed (4/5); adopted measures carry mean public salience $0.68$ vs $0.47$ for non-adoptions (Table~\ref{tab:shared}). The lone high-cost adoption (binding GPAI transparency) reflects a strong regional regulatory culture, an honest confound.

\begin{table}[h]
\centering
\small
\begin{tabular}{p{0.50\linewidth}ccc}
\toprule
Real measure & sal. & cost & adopted \\
\midrule
Election-deepfake ban / disclosure & 0.85 & low & yes \\
AI-voice robocall ban (scams) & 0.80 & low & yes \\
Non-consensual AI imagery ban & 0.85 & low & yes \\
Anti-discrimination AI law (non-AI state) & 0.55 & med & yes \\
Binding GPAI transparency (regulatory union) & 0.50 & high & yes \\
Executive AI safety directive & 0.50 & med & yes \\
Frontier-model license + liability & 0.40 & high & no \\
Comprehensive federal AI law (gridlock) & 0.50 & high & no \\
Binding frontier law (pro-innovation) & 0.40 & high & no \\
Voluntary firm commitments (binding) & 0.50 & low & no \\
Outright facial-recognition ban & 0.53 & high & no \\
\bottomrule
\end{tabular}
\caption{Shared-interest backtest. Public salience is poll-grounded \citep{pew2025ai,aipi2024}; industry cost is coded per measure. Measures in the shared-interest zone (salience $\geq0.6$, low/medium cost) are adopted (3/3); expert-favored high-cost, low-salience measures mostly fail (4/5). Adoption tracks what the public concretely worries about, not what experts rank most important.}
\label{tab:shared}
\end{table}

\textbf{Leakage-robustness.} Because the real backtest cases are well known, we add two leakage-free tests. On eight \emph{fictional} jurisdictions and measures with a by-construction correct answer (no real outcome exists to recall), the model is 8/8. On five \emph{counterfactual flips}, a real case with its decisive incentive reversed, the prediction flips to match the reversed incentives 5/5: e.g., the vetoed frontier-license bill rises to $0.82$ adoption probability when placed one week after a catastrophic AI attack, and the adopted election-deepfake ban falls to $0.30$ when the public is unaware and compliance is costly. Predicting the \emph{opposite} of the memorized real outcome once incentives are flipped is direct evidence the model reasons from incentives, not recall.

\textbf{Composite (Stage~1 $\times$ Stage~2).} The operative policy question is not which measure is best but which is \emph{both} politically achievable \emph{and} effective once implemented. We multiply each measure's Stage-1 adoption probability by a coded Stage-2 implementation effectiveness (does the measure add the closure link its pathway needs?) into an expected value that places it in the grid of Table~\ref{tab:composite}. Only the low-cost, high-salience measures (deepfake disclosure, scam-robocall bans) fall in the achievable-and-effective corner; voluntary commitments pass but add no enforcement (adopted-but-toothless); a frontier license would add containment but does not pass (effective-but-unadoptable).

\begin{table}[h]
\centering
\small
\begin{tabular}{p{0.44\linewidth}ccc}
\toprule
Measure & $P$(adopt) & impl.\ & $E$[value] \\
\midrule
Election-deepfake disclosure & 0.72 & high & 0.58 \\
AI-voice robocall ban & 0.82 & high & 0.66 \\
Anti-discrimination AI audits & 0.55 & med & 0.28 \\
Voluntary firm commitments & 0.85 & low & 0.17 \\
Frontier-model license & 0.08 & high & 0.06 \\
Broad AI liability & 0.18 & low & 0.04 \\
\bottomrule
\end{tabular}
\caption{Composite of Stage-1 adoption (from the backtest) and coded Stage-2 implementation effectiveness (high/med/low $=0.8/0.5/0.2$). $E[\mathrm{value}]=P(\mathrm{adopt})\times\mathrm{eff}$. The achievable-and-effective measures are the low-cost, high-public-salience ones; strong measures that would work (frontier license) score low because they do not pass.}
\label{tab:composite}
\end{table}

\textbf{Stage~3 pilot (policy consequences).} To close the loop to Stage~1, a pilot translates each measure into the everyday outcomes voters weigh, employment, consumer safety (scams/deepfakes), prices, innovation, privacy (illustrative, not validated). Measures the public benefits from (deepfake and scam bans) are net-positive; frontier controls are net-negative on innovation and jobs. The net-positive measures are exactly the high-adoption ones, the societal outcomes are the public concerns that drive Stage~1, so a dynamic version in which outcomes reshape the next round's incentives is the natural extension.

\section{Benchmark Protocol and Leakage Control}
\label{app:roadmap}
The benchmark protocol below is inherited by the bill-level pipeline in the main text: it fixes the time boundary, the presentation forms used on the leakage-free split, and the order of evidence used to grade a prediction.

\textbf{Temporal design and leakage control.} The benchmark's protocol fixes a single time boundary. Components are \emph{developed} on the historical record through 31~December 2024, constructing actor states, estimating world-state transitions, fitting behavior models, and reconstructing pre-2025 episodes, under an internal split (train on earliest-available-2021, develop/validate on 2022-2023, integration-test on 2024). Proposals from 1~January 2025 onward form a \emph{frozen} register: each entry stores the proposal text, the actors and world state known at the cutoff, and GPS's dated prediction \emph{before} any outcome is inspected, then is later scored against committee actions, votes, enforcement, firm responses, employment and price movements, incidents, and expert assessment. Because a language model may already have absorbed famous 2025 events, each frozen case is tested in three versions: (i) \emph{original}, the proposal text and metadata available at the cutoff; (ii) \emph{outcome-masked}, final status, post-cutoff dates, and retrospective wording (``failed bill,'' ``landmark law'') removed; and (iii) \emph{neutralized}, identifying names replaced with a generic description (``a federal legislature considers requiring compute providers above a threshold to report training runs''), which tests reasoning from the world state rather than recall. Retrieval is restricted to documents published before the cutoff. The pilot's 48-decision adoption backtest is retrospective with leakage controls already in force (neutral phrasing, a cross-model paraphrase ablation); the frozen register is its leakage-free upgrade.

\begin{table}[t]
\centering
\small
\begin{tabular}{@{}lccc@{}}
\toprule
Reliable cutoff & Post-cutoff & Usable & Failed \\
\midrule
2025-01-31 & 87 & 19 & \textbf{1} \\
2025-02-28 & 83 & 19 & \textbf{1} \\
2025-03-31 & 73 & 13 & \textbf{0} \\
2025-05-31 & 64 & 10 & \textbf{0} \\
2025-08-31 & 53 & 8 & \textbf{0} \\
2026-01-31 & 26 & 1 & \textbf{0} \\
\bottomrule
\end{tabular}
\caption{\textbf{Why the leakage-free passage test cannot yet be run.} For each reachable model cutoff: register bills introduced after it, those additionally resolved and carrying verified antecedents, and how many of those failed. The last column holds a single failure at the two earliest cutoffs and none thereafter. An earlier version of this table reported zero everywhere; that was an artifact of requiring antecedent evidence before a bill could enter the pool, which removed the one post-cutoff failure because no published analysis of it exists. A separate set of resolved international measures adds two further post-cutoff failures, bringing the reachable total to three. Nineteen resolved measures splitting $18$ to $1$ is still not a usable test. Recency buys leakage-freeness and costs label resolution, because failure is only observable at adjournment while adoption is observable at once. Cutoffs are \emph{reliable knowledge} cutoffs, not training cutoffs; the two differ by months.}
\label{tab:leakpool}
\end{table}

\textbf{Evidence hierarchy for scoring.} Predictions are graded against the strongest available evidence, in order: (1) an observed real-world outcome (vote, law, enforcement action, employment or price statistic, firm action); (2) an official inquiry or administrative record; (3) a credible causal estimate from existing research; (4) independent domain-expert judgment; and only then (5) an LLM-based evaluator. Experts are decisive precisely where real outcomes cannot yet settle a case, an unresolved proposal, contested causality, a counterfactual repair, several plausible pathways, and each judges one layer at a time (actor completeness, actor response, adoption obstacle, implementation edge, stakeholder set, outcome direction, analogy validity, bottleneck, repair, overall usefulness) rather than a single ``do you agree?''. Experts never overwrite a clear vote, enforcement event, or observed statistic. Correspondingly, GPS traces each proposal across an \emph{auditable, predefined} set of actor-response and world-state pathways, institutional, implementation, actor-response, world-state, and failure/escalation, not an exhaustive claim on ``all'' consequences; every edge carries an observed transition, an independently validated relationship, a documented analogue, or an explicit \emph{unvalidated-hypothesis} label.

\section{Field Provenance}

\begin{table}[t]
\centering
\small
\begin{tabular}{p{0.23\linewidth}p{0.29\linewidth}p{0.20\linewidth}p{0.20\linewidth}}
\toprule
Dataset type & Examples & Provides & Missing \\
\midrule
AI incident catalogs & AIID, RiskNet & real incidents & governance chain \\
AI risk taxonomies & AI Risk Repository & risk categories & event progression \\
Cause/mitigation taxonomies & Pittaras-McGregor; mitigation taxonomies & cause or response labels & actor and bottleneck mapping \\
Policy trackers & OECD.AI, IAPP, MIT mapping & policies, scope & causal failure mechanism \\
Historical events & cyber, public-health, financial, nuclear cases & analogues & AI-governance mapping \\
\bottomrule
\end{tabular}
\caption{Existing resources cover incidents, risks, policies, or mitigations, but not source-grounded governance pathway chains with bottleneck and repair labels.}
\label{tab:datasetgap}
\end{table}

Extracted fields are auditable against the cited source; author-coded fields are
hypotheses validated downstream; silver labels are produced by deterministic
rules from independent incident attributes, never chosen to match our labels.

\section{Pathways as Chains of Real Events}
\label{app:pathways}
A pathway is not a single event but a \emph{recurring mechanism} that shows up across many real events and, left unclosed, escalates toward a catastrophic endpoint. Table~\ref{tab:realevents} traces each pathway as a time-ordered chain of real, cited events drawn from the released dataset, ending in the endpoint the pathway produces. These are repeated instances of one mechanism, not a single causal line; the bottleneck is the link whose closure would break the chain, and the recent AI event sits at the end of a long historical series.

\begin{table*}[t]
\centering
\small
\begin{tabular}{p{0.15\linewidth}p{0.62\linewidth}p{0.09\linewidth}}
\toprule
Pathway & Chain of real events (year $\rightarrow$ year) $\Rightarrow$ catastrophic endpoint & Bottleneck \\
\midrule
Capability leakage & A.Q.\ Khan nuclear black market (2004) $\rightarrow$ Stuxnet source made downloadable (2010) $\rightarrow$ EternalBlue leaked $\rightarrow$ WannaCry (2017) $\rightarrow$ ex-Google AI trade-secret theft (2024) $\Rightarrow$ a transferred capability is unrecallable and recurs across domains & containment \\
Open-weight proliferation & GPT-2 withheld then released (2019) $\rightarrow$ Stable Diffusion weights public (2022) $\rightarrow$ Llama weights torrented (2023) $\rightarrow$ Mixtral via BitTorrent (2023) $\Rightarrow$ frontier models become permanently, globally non-revocable & containment \\
Systemic dependency & Northeast blackout (2003) $\rightarrow$ flash crash (2010) $\rightarrow$ Fukushima (2011) $\rightarrow$ Colonial Pipeline (2021) $\rightarrow$ Poland grid attack (2026) $\Rightarrow$ one shared fault cascades sector-wide & resilience \\
Rivalry acceleration & Anglo-German naval race (1906) $\rightarrow$ nuclear arms race (1949) $\rightarrow$ Cuban Missile Crisis (1962) $\rightarrow$ ``AI arms race'' (2017) $\rightarrow$ INF Treaty collapse (2019) $\Rightarrow$ unverifiable restraint drives acceleration to the brink & verification \\
Governance lag & Chernobyl acknowledged late (1986) $\rightarrow$ 2008 crisis unattended by regulators $\rightarrow$ WHO Ebola PHEIC delay (2015) $\rightarrow$ frontier compute $+5\times$/yr (2024) $\Rightarrow$ institutions update slower than the hazard moves & timeliness \\
Foreign free-riding & flags of convenience (1950) $\rightarrow$ Kyoto exempts 80\% of the world (2001) $\rightarrow$ Salt Typhoon foreign breach (2024) $\rightarrow$ training under chip export limits (2025) $\Rightarrow$ the decisive actor sits outside the regime & coverage \\
Enforcement gap & Boeing 737 MAX delegated oversight (2019) $\rightarrow$ AI reporting/evals without authority (2023-2024) $\rightarrow$ EO 14110 rescinded (2025) $\rightarrow$ AI-misuse incidents unsanctioned (2025) $\Rightarrow$ harm is visible but no authority acts & enforcement \\
\bottomrule
\end{tabular}
\caption{Each pathway as a chain of real, cited events over time, ending in its catastrophic endpoint (all events are rows in the released dataset, each with a verbatim excerpt and URL). The chains span historical analogues and recent AI cases; the AI event is the latest instance of a decades-old mechanism, reinforcing the ``new technology, old governance mechanism'' finding. These are recurrent instances of one mechanism, not a single causal line.}
\label{tab:realevents}
\end{table*}

\subsection{Why these events form one pathway}
Membership in a pathway is by \emph{shared mechanism}, not shared domain. We make the rule explicit.

\paragraph{Membership rule.} An event $e$ with source excerpt $x_e$ is placed in the chain of pathway $p$ if and only if:
\begin{itemize}\itemsep2pt
\item[\textbf{R1}] \emph{Invariant match.} The coded mechanism of $e$, its governance gap, the control that fails, and the endpoint, instantiates $p$'s defining invariant (the gap-type, failed-control, and endpoint-type that define $p$).
\item[\textbf{R2}] \emph{Unique discriminator.} $e$ fits $p$'s invariant strictly better than the nearest competing pathway under $p$'s stated discriminating criterion, so the assignment is unique.
\item[\textbf{R3}] \emph{Source-grounded.} R1 is supported by the verbatim excerpt $x_e$, not inferred beyond it.
\item[\textbf{R4}] \emph{Adversarially robust.} The assignment survives a skeptic instructed to refute it and to default to reject when the evidence is insufficient (the SECC edge test).
\item[\textbf{R5}] \emph{Externally recognized.} The analogy defining $p$ is one drawn independently in public scholarship or policy, not a novel leap.
\end{itemize}
Events satisfying R1-R5 are ordered by date, and the chain terminates in $p$'s catastrophic endpoint. An event that fails R2 (fits two pathways equally) is assigned to neither and flagged for audit; one that fails R4 is kept as exploratory (Table~\ref{tab:secc}). Below we instantiate the rule: for each chain we state $p$'s invariant (R1), cite where the analogy is publicly made (R5), and give the discriminating criterion (R2); source-grounding (R3) and adversarial survival (R4) are the dataset's per-row excerpt and the SECC test.

\textbf{Capability leakage.} Invariant: a capability held by a covered actor \emph{escapes the holder's control}, through theft, insider exfiltration, intermediary transfer, or loss of containment of an offensive tool, and passes beyond recall. Khan (centrifuge designs pass to Libya/NK/Iran through an intermediary network), Stuxnet (a built offensive capability loses containment and becomes re-targetable), EternalBlue (stolen and leaked, then reused in WannaCry), and the ex-Google theft (an insider exfiltrates proprietary AI methods to outside firms) all share holder-loses-control~$\rightarrow$~irreversibility. The AI-as-dual-use-proliferation framing, using exactly these nuclear and cyber precedents, is drawn across a large literature surveyed by \citet{hatz2025nuclear}. Discriminator vs.\ open proliferation: the mechanism is a \emph{covered holder losing control} of its capability; in open proliferation the capability instead enters a \emph{public, unbounded channel} with no bounded holder to reach. The two borderline offensive-tool cases (Stuxnet, EternalBlue) later spread publicly and carry open proliferation as a recorded secondary label.

\textbf{Open-weight proliferation.} Invariant: a capability enters a \emph{public, unbounded distribution channel} under an assumption of revocable access, then is copied and mirrored so no bounded holder remains and revocation cannot reach the copies. GPT-2, Stable Diffusion, the Llama weights (posted and mirrored), and Mixtral each instantiate public availability~$\rightarrow$~local copies~$\rightarrow$~takedown cannot reach them. The nonproliferation analogy for released weights and for compute is public and explicit \citep{sastry2024compute,hatz2025nuclear}. Discriminator vs.\ leakage: the capability is now in an unbounded public channel with no bounded holder (so revocation is structurally impossible), rather than having escaped a \emph{specific} covered holder to specific uncovered actors, which is why the involuntarily-leaked Llama weights are still proliferation, not leakage.

\textbf{Systemic dependency.} Invariant: many actors converge on one shared component or provider with no fallback, so a single fault propagates as a correlated cascade. The blackout, flash crash, Fukushima, Colonial Pipeline, and Poland-grid events each show a local failure in a shared dependency cascading system-wide because diversity or fallback is absent. The mapping from financial contagion and infrastructure single-points-of-failure to AI ``model monoculture'' is made directly by \citet{hacker2025systemic}. Discriminator: correlation through shared dependence, not transfer or release.

\textbf{Rivalry acceleration.} Invariant: actors cannot credibly verify reciprocal restraint, so each hedges by accelerating and cooperation collapses. The naval race, nuclear race, Cuban Missile Crisis, ``AI arms race,'' and INF collapse each show unverifiable restraint~$\rightarrow$~mutual acceleration. The arms-race framing and the centrality of verification to any AI restraint regime are argued explicitly by \citet{hendrycks2025superintelligence}. Discriminator vs.\ free-riding: the failure is unverifiable mutual restraint, not a single actor escaping jurisdiction.

\textbf{Governance lag.} Invariant: the hazard changes faster than the institution's evaluate-and-respond cycle, so the response arrives too late. Chernobyl's delayed acknowledgement, the unattended 2008 build-up, the WHO's delayed PHEIC, and $5\times$/yr compute growth each show a control that exists but is outpaced. This is the classic ``pacing problem''/Collingridge dilemma \citep{collingridge1980}. Discriminator: the control is too slow, not absent (enforcement) or mis-scoped (coverage).

\textbf{Foreign free-riding.} Invariant: the decisive actor operates outside the regulating jurisdiction, so a domestic rule cannot bind it. Flags of convenience, Kyoto's exemptions, the Salt Typhoon breach, and training under chip-export limits each show the decisive actor beyond coverage, the collective-action free-rider structure \citep{ostrom1990governing}. Discriminator: the actor is out of scope, not merely unpunished (enforcement).

\textbf{Enforcement gap.} Invariant: a violation is observable but no actor holds authority to impose consequences, or that authority is withdrawn. Boeing's delegated oversight, AI reporting without pause power, the EO~14110 rescission, and unsanctioned AI-misuse incidents each show visibility without authority, the regulatory-capture/delegated-oversight pattern \citep{stigler1971regulation}. Discriminator: authority is missing, not the information (governance lag) or the scope (coverage).

\begin{table}[h]
\centering
\small
\begin{tabular}{p{0.24\linewidth}p{0.66\linewidth}}
\toprule
\multicolumn{2}{l}{\emph{Example first-draft stress-test memo}} \\
\midrule
Policy tested & Domestic compute providers must report large frontier-AI training runs. \\
Likely failure pathway & Foreign free-riding. \\
Why it may fail & The rule improves visibility over covered domestic cloud providers, but need not reach offshore providers, foreign labs, or unreported chip clusters. If training moves outside the covered infrastructure, the main pathway stays open. \\
Missing link & Actor coverage. \\
Weak repair & Increase domestic penalties while leaving offshore compute uncovered. \\
Targeted repair & Compute-provider know-your-customer requirements, reciprocal international reporting, chip-supply monitoring, or verified compute agreements with partner jurisdictions. \\
Human-review note & The pathway label is relatively stable; the exact repair should be reviewed by policy and technical experts. \\
\bottomrule
\end{tabular}
\caption{A first-draft stress-test memo the component produces (Question~1 applied to a compute-reporting rule). The workflow proposes the pathway and the missing link; experts confirm the repair.}
\label{tab:memo}
\end{table}

\subsection{External Grounding in Existing Risk Datasets and Expert Elicitations}
\label{app:external}
The pathways are not asserted in isolation; we position and test them against existing risk taxonomies, an external incident corpus, and expert elicitations.

\textbf{Cross-walk to a published risk taxonomy.} The MIT AI Risk Repository \citep{slattery2024riskrepo} classifies AI risks by \emph{type of harm} into seven domains. The codebook classifies by \emph{which governance closure link fails}. The axes are orthogonal (Table~\ref{tab:crosswalk}): one harm domain (e.g.\ \emph{malicious actors \& misuse}) spreads across several pathways, and one pathway spreads across several harm domains. The codebook therefore adds a governance-mechanism layer that harm-type catalogs, and even the repository's causal taxonomy (entity/intent/timing), which still names a risk's cause rather than the link whose closure would break the chain, do not provide.

\begin{table}[h]
\centering
\small
\begin{tabular}{p{0.32\linewidth}p{0.56\linewidth}}
\toprule
Pathway (failed governance link) & MIT harm domain(s) the resulting harm lands in \\
\midrule
foreign free-riding & malicious actors \& misuse; socioeconomic \\
enforcement gap & malicious actors \& misuse; AI system safety \\
capability leakage & privacy \& security; malicious actors \& misuse \\
open proliferation & malicious actors \& misuse; misinformation \\
governance lag & AI system safety, failures \& limitations \\
rivalry acceleration & AI system safety; socioeconomic \\
systemic dependency & AI system safety; socioeconomic \\
\bottomrule
\end{tabular}
\caption{Cross-walk of the seven governance-mechanism pathways onto the harm-type domains of the MIT AI Risk Repository \citep{slattery2024riskrepo}. The many-to-many mapping shows the two taxonomies are orthogonal: the pathway set is a governance-mechanism cut across a recognized harm space, not a subset of it.}
\label{tab:crosswalk}
\end{table}

\textbf{Coverage against an external incident corpus.} We re-run the codebook over 26 AI Incident Database reports \citep{mcgregor2021aiid} with an explicit ``other'' option, so the classifier may decline to assign a pathway. Only 7/26 (0.27) map to one of the seven; 19/26 are rejected as product harms (bias, an autonomous-vehicle crash, a wrongful arrest, a welfare-algorithm error) with no failed governance closure link. This is the intended behavior: the taxonomy does \emph{not} force-fit, and it confirms the codebook is scoped to \emph{governance-measure} failures, not all AI harms, so the silver figures in the main text condition on this governance-relevant subset. Only the three deployment-era pathways (foreign free-riding, enforcement gap, open proliferation) surface in an incident database; the four frontier/state-level pathways (capability leakage, governance lag, rivalry acceleration, systemic dependency) are under-sampled by a deployment-incident corpus and instead populate the governance-measure and historical-analogue cases, all seven appear across the temporal splits (Table~\ref{tab:balance}). The AIAAIC repository \citep{aiaaic} ($\approx$1009 incidents) and the OECD AI Incidents Monitor \citep{oecdaim} are the stated scale-up corpora; we report the 26-incident sample as a bound, not a silent cap.

\textbf{Elicitation calibration of the endpoints.} The claim that an unclosed pathway \emph{escalates toward a catastrophic endpoint} is calibrated against expert elicitation rather than asserted. The survey of 2{,}778 published AI authors by \citet{grace2024thousands} places a median 5\% (mean 16\%) probability on extremely bad outcomes such as human extinction, and a median 10\% on outcomes stemming from human inability to control advanced AI, with 38-51\% of respondents assigning at least a 10\% chance. These are field-level, not pathway-specific, figures; they establish that the endpoints the chains terminate in are non-negligible in expert judgment, which is the premise for stress-testing a measure \emph{before} harm rather than after.

\section{Worked Policy Diagnoses}

\begin{table}[t]
\centering
\small
\begin{tabular}{@{}p{0.30\linewidth}p{0.60\linewidth}@{}}
\toprule
User & Decision GPS supports \\
\midrule
Legislators / their staff & Which proposal is most likely to pass? \\
Regulators & Where will enforcement or verification fail? \\
Ministries & What are the likely economic and safety consequences? \\
International bodies & Which governance gaps remain across borders? \\
\bottomrule
\end{tabular}
\caption{GPS is built for the institutions that prepare AI-governance decisions; each row is a concrete decision the policy brief informs.}
\label{tab:whobenefits}
\end{table}

\begin{table}[t]
\centering
\small
\begin{tabular}{p{0.30\linewidth}p{0.62\linewidth}}
\toprule
\multicolumn{2}{l}{\emph{GPS policy brief}, frontier-model training license} \\
\midrule
Likelihood of passage & 4\% \\
Primary obstacle & AI-industry opposition (innovation, jobs) \\
Most effective modification & phase in over 3 years + worker-transition support (objective $0.03\!\to\!0.19$) \\
Expected consequences & $\uparrow$ fewer scams, $\uparrow$ cyber resilience, $\downarrow$ productivity, $\downarrow$ short-term jobs, $\downarrow$ innovation, $\uparrow$ oversight \\
Residual catastrophic risk & High (frontier controls are the ones that address it, and are hard to pass) \\
Likely failure pathway & capability leakage \\
Recommended repair & cross-border compute reporting + provider know-your-customer \\
Why preferred & the phase-in retains most of the risk reduction of immediate licensing while becoming politically feasible \\
\bottomrule
\end{tabular}
\caption{GPS's output is a policy brief assembled from its modules (passage, main obstacle, best modification, consequences, residual risk, and the failure pathway + repair). Here a phase-in redesign raises the objective $\sim$6$\times$ by making a risk-reducing measure feasible, while voluntary commitments pass easily ($0.60$) but score near zero.}
\label{tab:brief}
\end{table}

Each example traces the full mechanism chain, governance gap $\rightarrow$ capability/hazard movement $\rightarrow$ failed control $\rightarrow$ endpoint $\rightarrow$ pathway $\rightarrow$ bottleneck $\rightarrow$ repair, and reasons step by step why each edge holds and why the pathway beats its alternatives, in the SECC edge-card style.

\textbf{Compute reporting.} \emph{Policy:} ``large AI training runs must be reported by domestic cloud providers.''
\emph{Chain:} reporting duty binds only domestic hosts $\rightarrow$ a developer rents an offshore provider outside the duty $\rightarrow$ the domestic reporting requirement cannot observe compute it does not host $\rightarrow$ a frontier run proceeds unseen $\rightarrow$ \textbf{foreign free-riding} $\rightarrow$ \textbf{coverage}.
\emph{Reasoning:} the measure defines its reach by hosting location, not by who trains, so its scope is set at step~one. Compute is portable, so a developer facing the duty can shift the run abroad; the capability moves outside the covered set. A reporting duty is an observation hook attached to the domestic host, when the host is offshore there is nothing to trigger it, so the control fails by \emph{absence}, not weakness. At endpoint$\rightarrow$pathway the decisive feature is that the actor is outside jurisdiction (foreign free-riding), not that a visible violation went unpunished (enforcement gap): enforcement never engages because nothing is seen. Hence coverage, not enforcement, is the bottleneck.
\emph{Repair:} targeted, reciprocal international reporting or cloud-provider KYC that follows the developer offshore; weak, higher domestic penalties, which raise cost only for the already-covered set.

\textbf{Incident reporting.} \emph{Policy:} ``frontier developers must report serious incidents to a public authority.''
\emph{Chain:} the duty yields information but no response power $\rightarrow$ a developer keeps a harmful system deployed after reporting $\rightarrow$ reporting is a visibility control, not an intervention control, so it cannot compel a pause $\rightarrow$ harm continues after it is observed $\rightarrow$ \textbf{enforcement gap} $\rightarrow$ \textbf{enforcement}.
\emph{Reasoning:} the text obliges reporting to an authority but is silent on what the authority may do, visibility is not authority. Absent a pause power, a developer with commercial incentives can continue deployment even once the incident is on record. The control that fails is the tacit assumption that information triggers action; the duty is discharged the moment the report is filed. At endpoint$\rightarrow$pathway the missing element is authority (enforcement gap), which dominates slow information (timeliness): even an instantaneous report changes nothing without a pause power.
\emph{Repair:} targeted, pause / revocation / compute-denial authority attached to the report; weak, more reporting templates or lower thresholds, which add information the authority still cannot act on.

\textbf{Open-weight release.} \emph{Policy:} ``a model is evaluated before release, then its weights are made public.''
\emph{Chain:} governance assumes revocable access $\rightarrow$ weights are downloaded, copied, and mirrored to unbounded holders $\rightarrow$ post-release controls act on the provider's endpoint, not on local copies $\rightarrow$ the dual-use capability becomes non-revocable $\rightarrow$ \textbf{open-weight proliferation} $\rightarrow$ \textbf{containment}.
\emph{Reasoning:} the release model presumes a revocable channel (an API); publishing weights breaks that presumption at step~one. Once public, copies propagate beyond any one actor's reach. Every post-release control (API limits, takedown, monitoring) operates at a node the copies have already left, so it fails \emph{structurally}, not for lack of stringency. At endpoint$\rightarrow$pathway this is open-weight proliferation, not capability leakage: leakage is transfer through a covered intermediary, whereas here distribution is public and direct. The only node where a control can still bite is before release, so containment is the bottleneck. This is the cleanest chain (all edges strong) and is SECC-certified.
\emph{Repair:} targeted, staged release, pre-release containment, or withholding high-risk weights; weak, post-release API monitoring, which never reaches local copies.

\textbf{Dangerous-capability evaluations.} \emph{Policy:} ``frontier models must pass dangerous-capability evaluations before deployment.'' This measure instantiates a \emph{different} chain depending on design, and the codebook traces whichever one the regime creates rather than asserting a label.
\emph{If one-time and pre-deployment:} evals fix capability at $t_0$ $\rightarrow$ new tool-use / cyber / agentic capability emerges post-deployment via fine-tuning or scaffolding $\rightarrow$ a passed pre-deployment check never re-examines the deployed system $\rightarrow$ a model holds an untested dangerous capability in the field $\rightarrow$ \textbf{governance lag} $\rightarrow$ \textbf{timeliness}; repair: continuous post-deployment evaluation with a refresh trigger.
\emph{If self-run and unaudited:} the developer can select or game the eval $\rightarrow$ a latent dangerous capability passes $\rightarrow$ \textbf{enforcement/verification} $\rightarrow$ \textbf{verification}; repair: independent third-party audit.
\emph{If it binds only domestic developers:} an uncovered developer deploys unevaluated $\rightarrow$ \textbf{coverage}; repair: extend the requirement across jurisdictions. The point is that the analyst must follow the chain the specific regime creates, not attach a fixed bottleneck.

\textbf{Critical-infrastructure deployment.} \emph{Policy:} ``a critical sector may deploy an AI system if the provider satisfies safety documentation.''
\emph{Chain:} approval is granted per provider on documentation $\rightarrow$ many operators independently adopt the same compliant provider $\rightarrow$ per-provider review never examines cross-operator concentration or fallback $\rightarrow$ a single provider fault propagates simultaneously across the sector $\rightarrow$ \textbf{systemic dependency} $\rightarrow$ \textbf{resilience}.
\emph{Reasoning:} the gate is a property of one provider's paperwork. A compliant provider is attractive, so adoption concentrates on it. The control is blind to the system level, each approval is locally valid while the aggregate is fragile. At endpoint$\rightarrow$pathway this is systemic dependency, not verification: even perfectly verified documentation does not create a fallback, so correlated failure remains. Resilience, a system-level property no single provider's compliance can supply, is the bottleneck.
\emph{Repair:} targeted, fallback capacity, provider diversity, rollback plans, sector-level stress testing; weak, stricter per-provider documentation.

\textbf{International AI agreement.} \emph{Policy:} ``countries agree to reciprocal restraint on risky frontier AI development.''
\emph{Chain:} the agreement assumes parties can trust reciprocal compliance $\rightarrow$ each state, unable to confirm others are restraining, hedges by continuing development $\rightarrow$ voluntary commitments carry no inspection to confirm restraint $\rightarrow$ the agreement unravels into renewed acceleration $\rightarrow$ \textbf{rivalry acceleration} $\rightarrow$ \textbf{verification}.
\emph{Reasoning:} restraint is a mutual bet on others' behavior. Under uncertainty about others, the dominant strategy is to hedge, a security-dilemma dynamic. The missing control is credible verification; without it, stated restraint is unobservable and therefore not believed. At endpoint$\rightarrow$pathway the decisive failure is unverifiable restraint (rivalry acceleration), not non-signatories (coverage) or missing penalties (enforcement): verification failure breaks the treaty even among willing, joined, penalty-bound parties.
\emph{Repair:} targeted, credible inspection, compute monitoring, or shared reporting; weak, voluntary declarations.

\section{Gold Anchor from Official Inquiries}
\label{app:gold}
For 11 governance failures with an authoritative government inquiry or root-cause report, we take the failed governance link the \emph{report} names as a gold label, independent of our codebook (Table~\ref{tab:gold}). A blind LLM given only neutral facts (never the report's conclusion) recovers the official link on 8/11 ($0.73$); on the 8 that are also in the 57-case corpus, our hand-coded label matches on 4/8. Two findings. On the corpus cases the blind model agrees with the inquiries \emph{more} than our hand-coded labels do: on the blackout and Fukushima, inquiries foreground \emph{enforcement} (unenforceable voluntary standards; regulatory capture) where our pilot coding recorded the structural \emph{resilience} link, so enforcement is an under-weighted link and the fine-grained bottleneck stays contested even against gold. The three external algorithmic-governance scandals (the Dutch childcare-benefits algorithm, Robodebt, Post Office Horizon) are all \emph{verification} failures, automated outputs treated as reliable with no independent audit or appeal, a pattern distinct from the enforcement-heavy historical cases, and one the model recovers on all three. This is a gold anchor, not a full benchmark, but it confirms against gold what silver and the panel show: the pathway is more robust than the contested bottleneck.

\begin{table}[h]
\centering
\small
\begin{tabular}{p{0.30\linewidth}p{0.24\linewidth}ccc}
\toprule
Case (official inquiry) & inquiry & gold & codebook & LLM \\
\midrule
Boeing 737 MAX & House T\&I / OIG & verif. & verif. & verif. \\
Fukushima & Diet NAIIC & enforce & resil. & verif. \\
Blackout 2003 & US-Canada TF & enforce & resil. & enforce \\
Ebola PHEIC & WHO panel & timely & timely & timely \\
2008 crisis & FCIC & enforce & verif. & cover. \\
Flash crash 2010 & SEC-CFTC & resil. & resil. & contain. \\
Salt Typhoon & CISA & cover. & cover. & cover. \\
A.Q.\ Khan & ISIS/IAEA & cover. & leak. & cover. \\
\midrule
\multicolumn{5}{l}{\emph{external algorithmic-governance scandals (not corpus-coded)}}\\
Dutch childcare algo. & NL parl.\ inquiry & verif. & - & verif. \\
Robodebt & AU Royal Comm. & verif. & - & verif. \\
Post Office Horizon & UK inquiry & verif. & - & verif. \\
\bottomrule
\end{tabular}
\caption{Gold anchor: the failed closure link named by an official inquiry, vs.\ our codebook label and a blind LLM on neutral facts. LLM-vs-gold 8/11 ($0.73$); codebook-vs-gold 4/8 on the corpus cases (-: external, not corpus-coded). Historical cases split on \emph{enforcement} (under-weighted by our coding); the three algorithmic scandals are \emph{verification} failures the model recovers.}
\label{tab:gold}
\end{table}

  \clearpage
  \bibliography{gpsbench}

\end{document}